%% file: main.tex
\documentclass[10pt,logo,copyright]{iwwhitepaper}
\newif\ifreport
\reportfalse

\input{styles/whitepaper/whitepaper_packages}
\input{metadata}
\input{tex/model_names}

\title{\whitepapertitle}
\hypersetup{pdftitle={\whitepapertitle}}
\renewcommand{\sectionmark}[1]{\markboth{\ifnum\value{secnumdepth}>0 \thesection.\ \fi#1}{}}
\newcommand{\docauthor}[2]{\author[#1]{#2}}
\newcommand{\docaffil}[2]{\affil[#1]{#2}}
\docauthors
\docaffiliations

\begin{abstract}
\input{tex/sections/abstract}
\end{abstract}

\begin{document}
\maketitle
\abscontent

\input{tex/sections/introduction}
\input{tex/sections/extended_compass_benchmark}
\input{tex/sections/real_world_benchmark}
\input{tex/sections/conclusion}

\clearpage
\bibliographystyle{plainnat}
\bibliography{references}

\clearpage
\appendix
\titleformat{\section}{\large\bfseries\headingfont}{Appendix \thesection:}{0.5em}{#1}
\renewcommand{\sectionmark}[1]{\markboth{Appendix \thesection: #1}{}}
\input{tex/appendix/appendix_a_controller_type_study}
\clearpage
\input{tex/appendix/appendix_b_omnidirectional_controller}
\clearpage
\input{tex/appendix/appendix_c_global_planner_integration}
\clearpage
\input{tex/appendix/appendix_d_training_datasets}
\clearpage
\input{tex/appendix/appendix_e_training_pipeline}
\clearpage
\input{tex/appendix/appendix_f_cosmos_samples}
\clearpage
\input{tex/appendix/appendix_g_jetson_deployment}
\clearpage
\input{tex/appendix/appendix_h_digital_twin_replica}
\end{document}

%% file: styles/whitepaper/whitepaper_packages.tex
\usepackage[authoryear,sort&compress,round]{natbib}

\usepackage[utf8]{inputenc} 
\usepackage[T1]{fontenc}    
\usepackage{xcolor}
\usepackage{parskip}        
\usepackage{url}            
\usepackage{booktabs}       
\usepackage{amsfonts}       
\usepackage{nicefrac}       
\usepackage{microtype}      
\usepackage{xcolor}         
\usepackage[dvipsnames]{xcolor} 
\usepackage{graphicx}
\usepackage{animate}        
\usepackage{subcaption}
\usepackage{tabularx}
\usepackage{makecell}
\usepackage{adjustbox}
\usepackage{setspace}
\newcolumntype{M}[1]{>{\centering\arraybackslash}m{#1}}
\usepackage{float}
\usepackage{tikz}
\usetikzlibrary{positioning,shapes,arrows}
\usepackage{amsmath,amsfonts,bm, bbm,leftindex}
\usepackage{multirow}
\usepackage{comment}
\usepackage{gensymb}
\usepackage{lipsum}
\usetikzlibrary{arrows.meta, positioning, fit, backgrounds, calc}
\usepackage[para]{threeparttable}
\usepackage{tikz}
\usetikzlibrary{tikzmark}
\usepackage{caption}
\usepackage{placeins}

%% file: metadata.tex
\def\whitepapertitle{Physical AI for Intralogistics: Benchmarking the NVIDIA Sim-to-Real Ecosystem for Cosmos-Enhanced COMPASS Navigation on Industrial Mobile Robots}
\newcommand{\docauthors}{%
    \docauthor{1}{Charbel Abi Hana}%
    \docauthor{1}{Rayan Zaitouni}%
    \docauthor{1}{Kameel Amareen}%
    \docauthor{1}{Mohamad Mostafa}%
    \docauthor{1}{Hasan~Sayour}%
    \docauthor{2}{Wei~Liu}%
    \docauthor{2}{Yan~Chang}%
    \docauthor{2}{Mihir Acharya}%
    \docauthor{1}{Anthony Rizk}%
}

\newcommand{\docaffiliations}{%
    \docaffil{1}{Idealworks}%
    \docaffil{2}{NVIDIA}%
}

\def\copyrightholder{Idealworks \& NVIDIA}
\def\partnerlogo{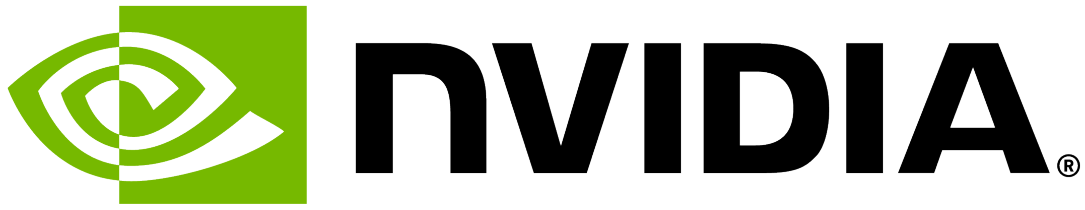}

%% file: tex/model_names.tex
\newcommand{\mdlBaseline}{IW Nav}             
\newcommand{\mdlNV}{NVIDIA COMPASS}           
\newcommand{\mdlSim}{IW.COMPASS}              
\newcommand{\mdlCE}{IW.CE-COMPASS}            
\newcommand{\wmSim}{IW.WM}       
\newcommand{\wmCE}{IW.CE-WM}     
\newcommand{\ilSim}{IW.IL}       
\newcommand{\ilCE}{IW.CE-IL}     

\newcommand{\na}{n/a}

\newcommand{\review}[1]{#1}
\newenvironment{reviewblock}{}{}

%% file: tex/sections/abstract.tex
Classical robot navigation stacks are reliable but often require re-engineering
for each new robot type and site. Learning-based navigation shifts this effort
from hand-tuned parameters to data generation and training, where a policy
adapts to each new embodiment or facility by fine-tuning in simulation. Whether policies
adapted this way stay reliable on real industrial robots has seen little
systematic testing. We study this for NVIDIA's
X-Mobility and COMPASS workflow for training learned navigation policies on
Idealworks robots. In simulation, we
extend the COMPASS benchmark with two embodiments, the differential-drive iw.hub
and an omnidirectional Humanoid Mobile Platform (HMP), and with a digital twin of the Idealworks R\&D
center. For the real world, we fine-tune X-Mobility on data from this digital
twin, with and without sim-to-real variants generated by NVIDIA Cosmos, train
COMPASS on top of it from scratch in the digital twin, deploy the resulting policies on the iw.hub's onboard Jetson AGX
Xavier, and benchmark them in the Idealworks R\&D center against COMPASS trained only on
NVIDIA's environments and the existing Idealworks navigation stack on task success, efficiency,
safety, smoothness and onboard compute. Running every static scenario both in a
digital-twin replica and in the real facility measures the sim-to-real gap and
how far Cosmos-augmented data closes it.
\review{Training in the digital twin with Cosmos sim-to-real augmentation raises
the real-world success rate of learned navigation from 65.3\% to 91.7\% across
four scenarios and nearly halves its weighted travel time compared with
simulation-only training, bringing it within 8.3 points of the production
navigation stack, fully onboard; with an integrated global planner, it also
reaches far, occluded goals in every trial.}

%% file: tex/sections/introduction.tex
\section{Introduction}
\label{sec:introduction}

Autonomous mobile robots (AMRs) have become a core component of modern
intralogistics, moving material between storage, production and shipping areas
without fixed infrastructure. Their navigation is typically built as a modular
pipeline: the robot localizes itself in a prior map, a global planner computes a
route to the goal, and a local planner follows that route while avoiding
obstacles, as in the widely used ROS~2 navigation stack
Nav2~\citep{macenski2020marathon}. The local planners of such stacks are
hand-designed, for example the dynamic window approach~\citep{fox1997dwa} or
timed elastic bands~\citep{rosmann2017teb}, and operate on costmaps built from
lidar and depth sensors.

These stacks are mature and dependable in the conditions they were engineered
for, but they are expensive to extend. Their behavior emerges from many
interacting parameters, such as cost weights, inflation radii, velocity and
acceleration limits and recovery behaviors, which must be re-tuned for every new
robot geometry, drive configuration and facility layout. Their perception is
largely geometric: an obstacle is an occupied cell, with no notion of what it
is or how it will move, which limits their behavior around people, other
vehicles and semi-structured clutter. As fleets grow to more robot types and
more sites, this per-robot and per-site engineering becomes a dominant cost of
deployment.

Physical AI, the application of learned models to systems that perceive and act
in the physical world, offers a different path: rather than being engineered, a
navigation policy is learned from data and improves as more data becomes
available. Several developments now make this practical. GPU-accelerated
simulation such as Isaac Lab~\citep{mittal2023orbit} trains policies across
thousands of parallel environments, which makes
interaction data cheap. World models learn compact, predictive representations
of the environment from which behavior can be
learned~\citep{ha2018worldmodels}. Navigation foundation
models trained on data from many robots, such as GNM, ViNT and
NoMaD~\citep{shah2023gnm,shah2023vint,sridhar2024nomad}, show that a single
policy can drive different embodiments, in line with the cross-embodiment
results of large-scale robot learning~\citep{oxe2024}. Training in simulation, however, raises its own
challenge: a policy that performs well in simulation can fail on a real robot,
because rendered images, physics and sensor noise never match reality exactly.
This sim-to-real gap is the main obstacle to deploying simulation-trained
policies~\citep{zhao2020sim2real}. A classical answer is domain randomization,
which varies textures, lighting and other simulation parameters during training
so that the real world appears to the policy as just another
variation~\citep{tobin2017domain}. Generative world foundation models such as
NVIDIA Cosmos~\citep{nvidia2025cosmos} take this a step further: they turn
simulated footage into photorealistic variations that preserve its geometry and
motion, bringing the training data closer to what a robot's cameras see in a
real facility.

Together, these advances make learned navigation policies the natural next step
for industrial navigation. A learned policy adapts to a new robot or a new site
through data and fine-tuning rather than manual re-engineering; it can use
camera images and the semantics they carry, not only geometry; and a single
training pipeline can serve a whole fleet of embodiments. Two questions stand
between this promise and industrial deployment: whether learned policies remain
reliable outside simulation, on real robots in real facilities, and whether they
can match the production navigation stacks they would replace on the metrics
that matter in operation, namely task success, efficiency, safety and onboard
compute.

NVIDIA's X-Mobility~\citep{liu2024xmobility} and
COMPASS~\citep{liu2025compass} are a concrete instance of this approach.
X-Mobility learns an end-to-end navigation policy: a world model, trained on
random-exploration data, encodes camera images, route information and robot
state into a latent representation, and a policy head, trained by imitation on
Nav2 expert demonstrations, decodes this representation into velocity commands.
COMPASS freezes the X-Mobility policy and trains a residual policy on top of it
with reinforcement learning in Isaac Lab, learning only the corrections a given
embodiment needs, either as a \emph{specialist} for one robot or as a
\emph{generalist} distilled across several.

Idealworks develops autonomous mobile robots for intralogistics together with
the software that operates them. Its iw.hub is a differential-drive AMR that
navigates with the Idealworks navigation stack, its production navigation system, and its Humanoid Mobile Platform (HMP) is a humanoid robot on an
omnidirectional mobile base.
New navigation capabilities at Idealworks are developed and validated first in
simulation and then in the Idealworks R\&D center, a facility used to test
robots under warehouse conditions before customer deployment. This makes the
Idealworks robots and the R\&D center a natural testbed for the two questions
above.

\paragraph{Scope of This Work}
\label{sec:scope}

This work evaluates whether the X-Mobility~\citep{liu2024xmobility} and
COMPASS~\citep{liu2025compass} navigation stack, combined with sim-to-real data
augmentation using NVIDIA Cosmos~\citep{nvidia2025cosmos}, can be adapted to
Idealworks robots and deployed in a real facility. We address this question in
two benchmarks, one in simulation and one on a real robot.

\textbf{Simulation benchmark.} We extend the simulation benchmark of COMPASS with two Idealworks embodiments,
the differential-drive iw.hub and the omnidirectional HMP,
and with a new evaluation environment, the digital twin of the Idealworks R\&D
center (Section~\ref{sec:sim_benchmark}). Because the released
stack moves the robot with a kinematic controller that bypasses actuation
dynamics, we adapt physics-based controllers to both robots: a differential
drive controller for the iw.hub and an omnidirectional controller for the
HMP. The HMP is evaluated in simulation only.

\textbf{Real-world benchmark.} On a physical iw.hub in the Idealworks R\&D center, we compare three COMPASS
policies with the Idealworks navigation stack, the production navigation system
of the iw.hub (Section~\ref{sec:real_world_benchmark}). The three add training on
the R\&D center step by step: \mdlNV{}, the iw.hub Specialist of the simulation
benchmark, is trained only on NVIDIA's environments; \mdlSim{} fine-tunes
X-Mobility on data generated in the digital twin of the R\&D center and trains
its COMPASS residual there; and \mdlCE{} does the same on a hybrid version of
that data extended with Cosmos-augmented sim-to-real variants. Each static
scenario is run both in a digital-twin replica and in the R\&D center. For
\mdlNV{}, this measures how it generalizes to a new facility, in simulation and
in the real world; for the policies trained in the digital twin, it exposes their
sim-to-real gap and how far Cosmos-enhanced data closes it. The systems are compared on task success,
efficiency, safety, motion smoothness and onboard compute. All learned policies
run on the iw.hub's onboard NVIDIA Jetson AGX Xavier. For far and occluded
goals, a global planner integrated with the COMPASS ROS~2 node supplies the route
that the COMPASS policy cannot plan on its own; it is used with \mdlCE{} on the
real robot.

\textbf{Organization.} Section~\ref{sec:sim_benchmark} introduces the two
embodiments and the digital twin of the R\&D center, and reports the simulation benchmark
across the original and new embodiments and environments.
Section~\ref{sec:real_world_benchmark} presents the real-world benchmark on the
iw.hub: the compared navigation systems, their training data, the scenarios, the
metrics, the onboard deployment and the results in the digital-twin replica and
in the R\&D center.
Section~\ref{sec:conclusion} concludes and outlines future work. The appendices
follow the order of the main text and detail the controller type study
(Appendix~\ref{app:controller_study}), the omnidirectional controller of the HMP
(Appendix~\ref{sec:omni_directional_controller}), the global planner integration
(Appendix~\ref{app:global_planner}), the training datasets
(Appendix~\ref{app:training_datasets}) and training pipeline
(Appendix~\ref{app:training_pipeline}) of the real-world models, samples of the
Cosmos sim-to-real augmentation (Appendix~\ref{app:cosmos_augmentation}), the
deployment on the Jetson AGX Xavier (Appendix~\ref{app:jetson_deployment}) and
the results of all metrics of the COMPASS policies in the digital-twin replica
(Appendix~\ref{app:twin_replica}).

%% file: tex/sections/extended_compass_benchmark.tex
\section{Extending the COMPASS Benchmark to New Embodiments and Environments}
\label{sec:sim_benchmark}

In this section, we build on the simulation benchmark of COMPASS~\citep{liu2025compass} and
extend it with two new embodiments, the iw.hub and the Humanoid Mobile Platform (HMP), and one new
evaluation environment, the digital twin of the Idealworks R\&D center.

\begin{figure}[htbp]
    \centering
    \begin{subfigure}[b]{0.40\linewidth}
        \centering
        \includegraphics[width=\linewidth]{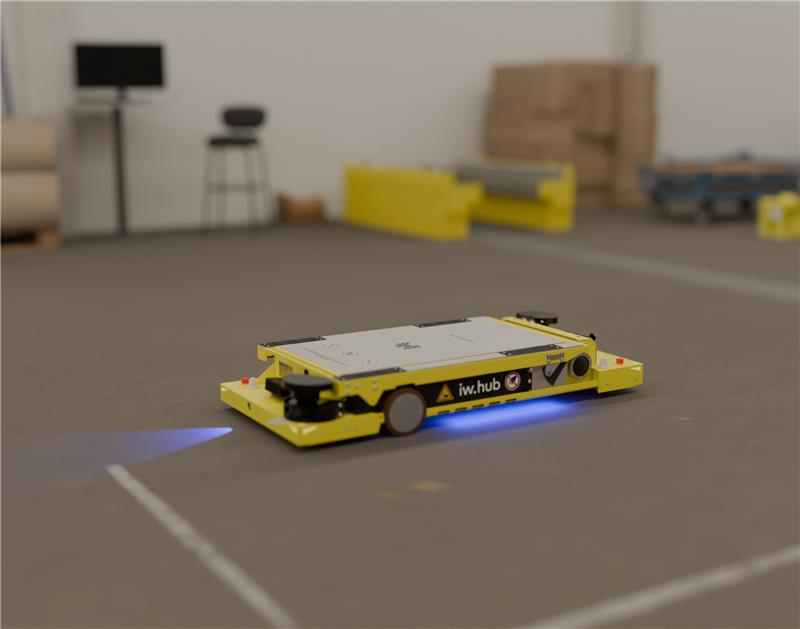}
        \caption{iw.hub}
        \label{fig:iwhub}
    \end{subfigure}
    \hspace{0.04\linewidth}%
    \begin{subfigure}[b]{0.40\linewidth}
        \centering
        \includegraphics[width=\linewidth]{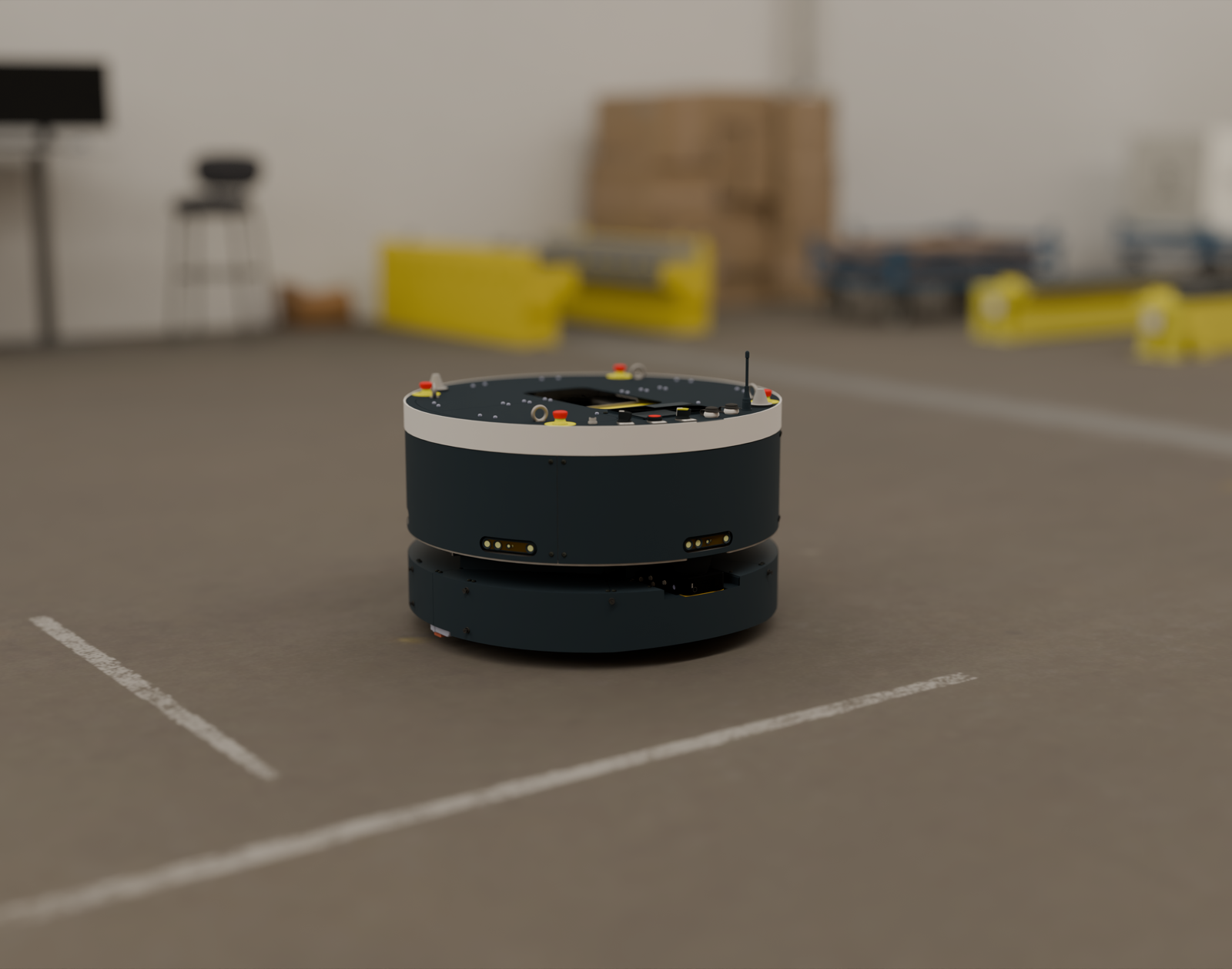}
        \caption{HMP}
        \label{fig:hmp}
    \end{subfigure}
    \caption{The two Idealworks embodiments added to the benchmark.}
    \label{fig:embodiments}
\end{figure}

\textbf{iw.hub.} The iw.hub is a differential-drive autonomous mobile robot (Figure~\ref{fig:iwhub}) designed for intralogistics tasks. It features a compact form factor suitable for navigating warehouse aisles, a payload capacity for transporting goods, and a sensor package that includes one LIPSEdge AE400 camera, two 2D LiDARs, and IMU sensors. The robot is equipped with an onboard Jetson AGX Xavier for on Edge computations.

\textbf{HMP.} The HMP is an omnidirectional Humanoid Mobile Platform
(Figure~\ref{fig:hmp}), a universal mobile base designed for autonomous mobile
manipulation robots (AMMRs). It can carry a humanoid upper body with multiple
degrees of freedom, an omnidirectional drive system for enhanced
maneuverability, and a comprehensive sensor suite including multiple cameras,
2D LiDARs and IMU sensors. For computation, the platform carries two onboard
NVIDIA modules, a Jetson AGX Orin and a Jetson AGX Thor.

\begin{figure}[htbp]
    \centering
    \begin{subfigure}[b]{0.49\linewidth}
        \centering
        \includegraphics[width=\linewidth]{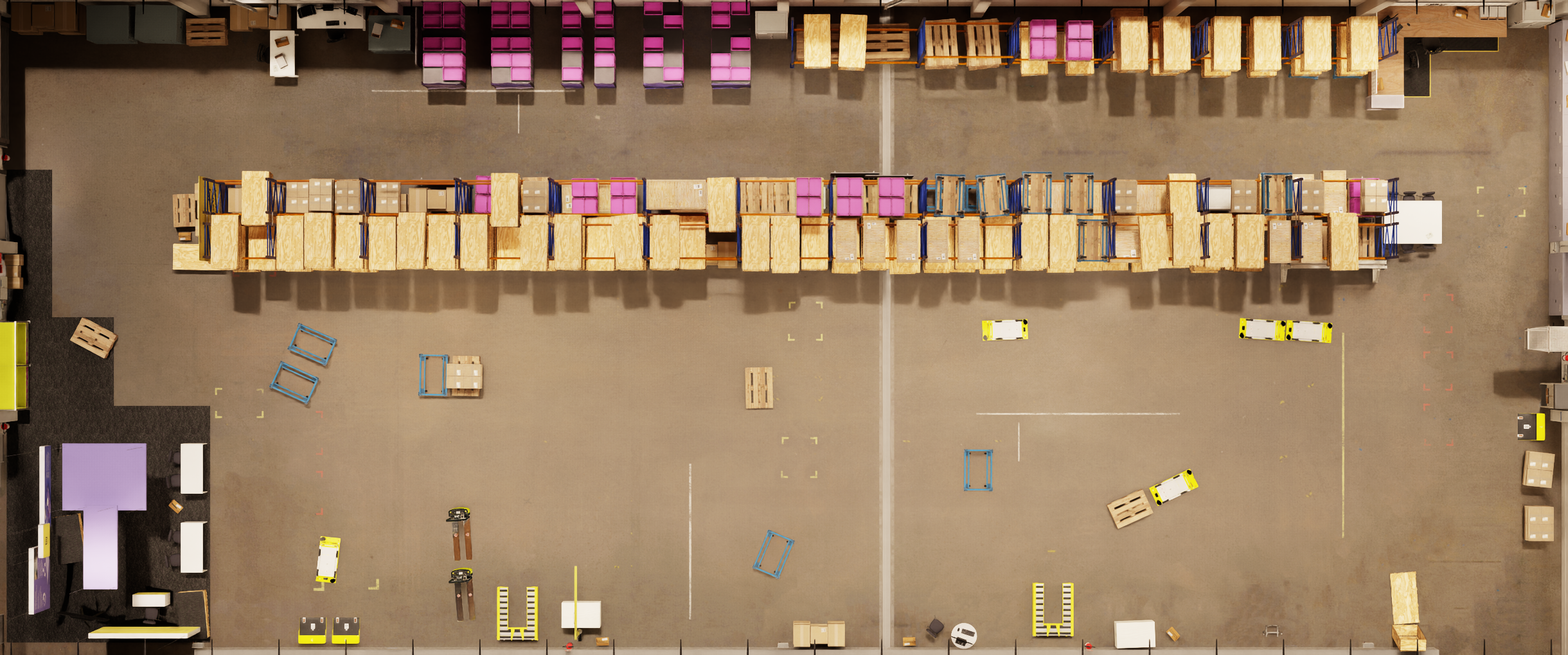}
        \caption{Randomized scenario, top view}
        \label{fig:iw_scenario_render}
    \end{subfigure}\hfill
    \begin{subfigure}[b]{0.49\linewidth}
        \centering
        \includegraphics[width=\linewidth]{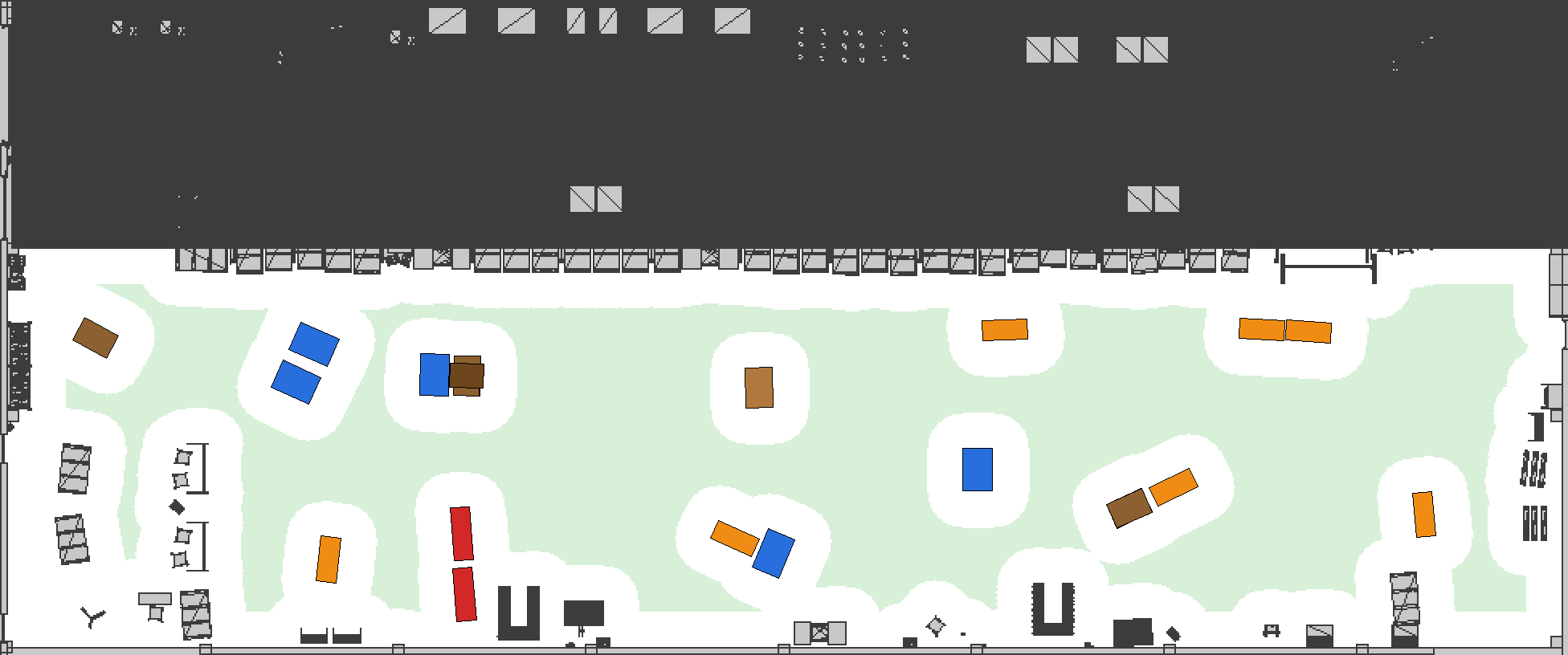}
        \caption{Occupancy grid of the same scenario}
        \label{fig:iw_scenario_occupancy}
    \end{subfigure}
    \\[0.6em]
    \begin{subfigure}[b]{0.75\linewidth}
        \centering
        \includegraphics[width=\linewidth]{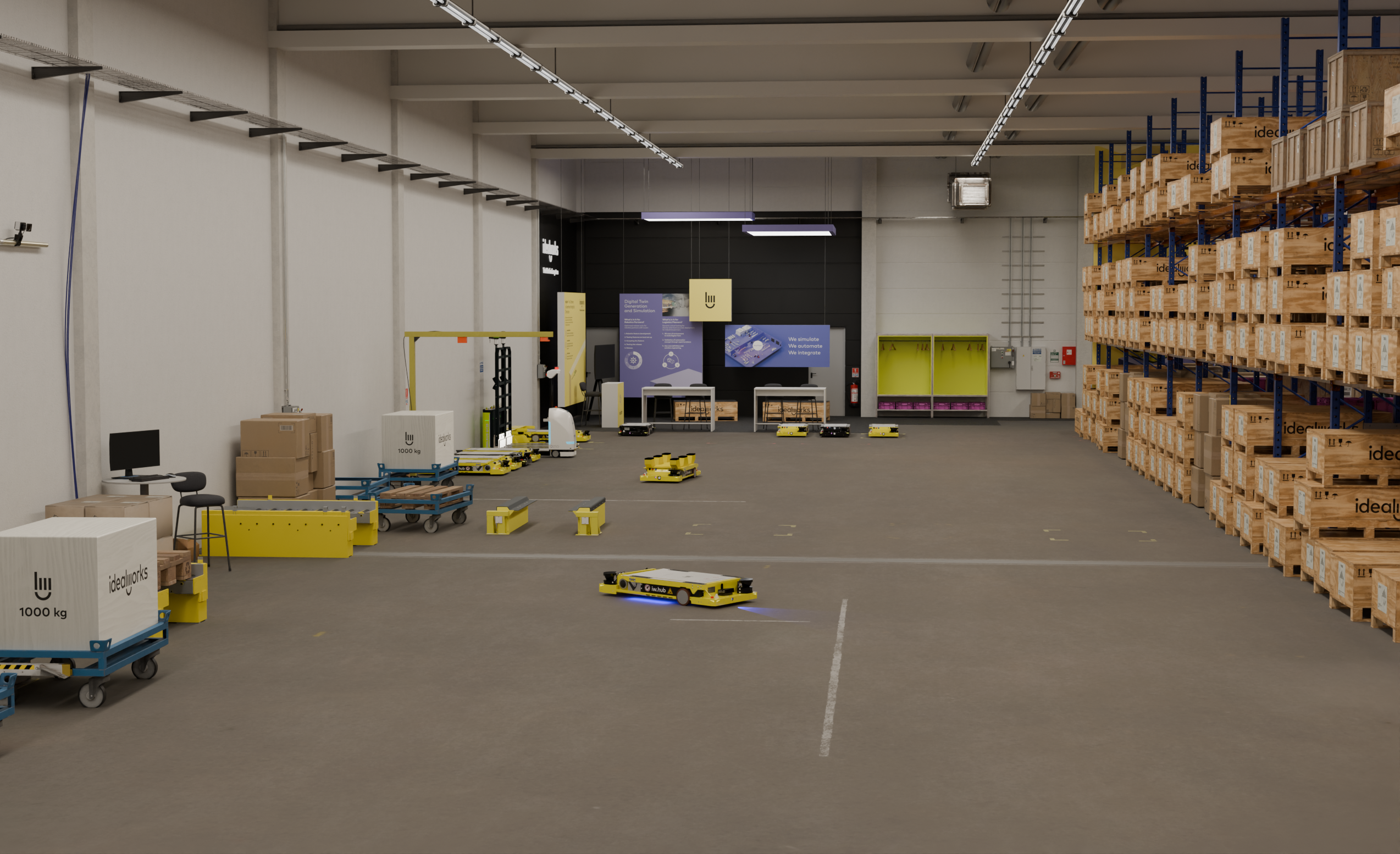}
        \caption{Inside view of the digital twin}
        \label{fig:iw_environment_inside}
    \end{subfigure}
    \caption{The Idealworks R\&D center in its digital twin. (a, b) The evaluation layout of this benchmark as a rendered top view and as its occupancy grid, in which dollies are blue, iw.hubs orange, pallets brown and forklifts red; the white band around each obstacle is the robot clearance, the light-green region is the drivable area, and the racking zone (dark) is not driven. (c) An eye-level view of the digital twin.}
    \label{fig:iw_environment}
\end{figure}

\textbf{Idealworks R\&D center.} The Idealworks R\&D center is a physical research and development facility. The simulation environment in Figure~\ref{fig:iw_environment} is its digital twin, replicating the layout, shelving, and other key features of the physical environment, providing a realistic simulation setting for evaluating navigation policies.
In this benchmark, the policies are evaluated in the digital twin on a single,
standalone evaluation layout of the R\&D center
(Figures~\ref{fig:iw_scenario_render} and~\ref{fig:iw_scenario_occupancy}).
Its floor is populated with 22 objects (7 pallets, one box stack, 5 dollies,
7 iw.hubs and 2 forklifts), placed in groups of one to three at random
positions and orientations in the open zone south of the racks. Groups keep
enough space between them for the robot to pass, so the layout stays fully
navigable: 241\,m$^2$ remain drivable for the iw.hub, about two thirds of the
drivable area of the empty R\&D center. The lighting is randomized as well, slightly brighter than nominal,
with 10 of the 46 ceiling lights switched off. Start and goal poses are sampled
on the occupancy grid of this layout.

\textbf{Integration into Isaac Lab and COMPASS.} The released COMPASS stack moves the robot with a kinematic controller that
updates its pose directly, bypassing actuation dynamics. We adapted
physics-based controllers to our embodiments: a differential drive controller
(DDC) for the iw.hub and an omnidirectional controller for the HMP
(Appendices~\ref{app:controller_study} and~\ref{sec:omni_directional_controller} detail their design and testing).

\textbf{Training.} All policies in this benchmark build on the pre-trained
X-Mobility policy released by NVIDIA, which is used as is, without fine-tuning,
as the frozen base policy of COMPASS (Figure~\ref{fig:sim_pipeline}). Each new
embodiment is integrated with its own controller, and a COMPASS residual policy
is trained for it in NVIDIA's environments of the original benchmark, which
yields the iw.hub and HMP Specialists. The digital twin of the R\&D center is
used only for evaluation; no policy in this benchmark is trained on it. The Generalist is distilled from the two Specialists,
following COMPASS~\citep{liu2025compass}.

\input{tex/figures/sim_benchmark_pipeline}

We evaluate trained policies across five environments: the four environments
of the original benchmark (Warehouse Multi-Rack, Warehouse Single-Rack, Simple
Office and Combined Multi-Rack, abbreviated in the tables as Warehouse M-R,
Warehouse S-R, Office and Combined M-R) and the digital twin of the R\&D center. Each policy is
trained for 1,000 iterations using 32 parallel simulation environments.
Each policy is evaluated over 20 episodes in each of 32 parallel environments,
that is 640 trials per environment, as in COMPASS. An episode terminates upon successful goal completion, collision, or
timeout.

\label{sec:evaluation_metrics}
We report the \textbf{success rate (SR)}, the percentage of episodes in which
the agent reaches the goal without collision or timeout, and the
\textbf{weighted travel time (WTT)}, the average travel time to the goal
conditioned on success and normalized by the success rate. We compute both
metrics exactly as the COMPASS benchmark does, so that the results are directly
comparable with those reported for the original embodiments; as in COMPASS, the
WTT values are of the order of thousands, and they are not comparable in absolute
terms with the per-trial travel times of the real-world benchmark of
Section~\ref{sec:real_world_benchmark}. We report the findings in Table~\ref{tab:navigation_results}, which summarizes
the performance of the COMPASS Specialist and Generalist policies across all
evaluation environments.

\begin{table}[H]
    \centering
    \caption{Cross-environment navigation performance in simulation: success rate and weighted travel time of the COMPASS Specialist and Generalist of each embodiment, over 640 trials per environment, computed as in COMPASS~\citep{liu2025compass}.}
    \label{tab:navigation_results}
    
    \renewcommand{\arraystretch}{1.2}
    \footnotesize
    
    \begin{adjustbox}{max width=\textwidth}
    \begin{tabular}{ll cc cc cc cc cc}
    \toprule

    \textbf{Embodiment} & \textbf{Model} &
    \multicolumn{2}{c}{\textbf{Warehouse M-R}} &
    \multicolumn{2}{c}{\textbf{Warehouse S-R}} &
    \multicolumn{2}{c}{\textbf{Office}} &
    \multicolumn{2}{c}{\textbf{Combined M-R}} &
    \multicolumn{2}{c}{\textbf{Idealworks R\&D center}} \\

    \cmidrule(lr){3-4}
    \cmidrule(lr){5-6}
    \cmidrule(lr){7-8}
    \cmidrule(lr){9-10}
    \cmidrule(lr){11-12}

    & &
    SR (\%) & WTT (s) &
    SR (\%) & WTT (s) &
    SR (\%) & WTT (s) &
    SR (\%) & WTT (s) &
    SR (\%) & WTT (s) \\

    \midrule

    \multirow{2}{*}{iw.hub}
    & COMPASS Specialist
    & 85 & 6,735
    & 76 & 5,495
    & 58 & 5,618
    & 76 & 6,802
    & 82 & 6,403 \\

    & Generalist
    & 66 & 6,174
    & 70 & 6,359
    & 36 & 5,604
    & 63 & 6,163
    & 77 & 5,938 \\

    \midrule

    \multirow{2}{*}{HMP}
    & COMPASS Specialist
    & 53 & 12,637
    & 65 & 11,866
    & 29 & 12,676
    & 53 & 11,850
    & 74 & 11,852 \\

    & Generalist
    & 50 & 12,117
    & 65 & 11,848
    & 31 & 11,855
    & 50 & 11,710
    & 72 & 11,596 \\

    \bottomrule
    \end{tabular}
    \end{adjustbox}
    
    \end{table}

\begin{reviewblock}
For the iw.hub, the Specialist reaches the goal in 58\% (Office) to 85\%
(Warehouse Multi-Rack) of the episodes and outperforms the Generalist in every
environment, by 5 to 22 points, the largest gap being in the Office. For the
HMP, the Generalist matches the Specialist: their success rates differ by at
most 3 points in every environment, so distilling across embodiments costs this
platform almost nothing. The HMP succeeds less often than the iw.hub overall
(29\% to 74\% for its Specialist) and takes about twice as long to reach the
goal. The Office is the hardest environment for every policy, with success rates
of 29\% to 58\%. The digital twin of the R\&D center, on which no policy is
trained gives the iw.hub Specialist its second-best
success rate (82\%) and every other policy its best one (72\% to 77\%).

Overall, both Idealworks embodiments navigate across all five environments with
COMPASS on top of NVIDIA's pre-trained X-Mobility policy, and the policies
trained in NVIDIA's environments transfer to the digital twin of the R\&D center
without having seen it. This extends the COMPASS benchmark with a
differential-drive and an omnidirectional platform, each of which required a
dedicated physics-based controller, and with a new environment modeled on a real
intralogistics facility.
\end{reviewblock}

%% file: tex/figures/sim_benchmark_pipeline.tex
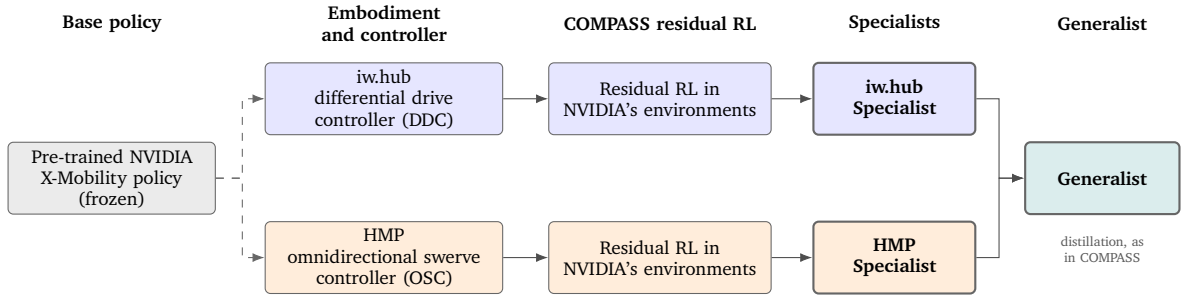
\begin{figure}[htbp]
    \centering
    \colorlet{nvcol}{black!8}
    \colorlet{hubcol}{blue!10}
    \colorlet{hmpcol}{orange!14}
    \colorlet{gencol}{teal!14}
    \begin{tikzpicture}[
        font=\scriptsize,
        box/.style={draw=black!60, rounded corners=2pt, align=center,
                    minimum height=0.95cm, inner sep=2pt},
        base/.style={box, text width=2.6cm, fill=nvcol},
        emb/.style={box, text width=3.0cm},
        rl/.style={box, text width=2.8cm},
        model/.style={box, text width=2.0cm, line width=0.8pt, font=\scriptsize\bfseries},
        flow/.style={-{Latex[length=1.6mm]}, black!70},
        frozen/.style={-{Latex[length=1.6mm]}, dashed, black!60},
        head/.style={font=\scriptsize\bfseries, align=center, text width=2.8cm},
        note/.style={font=\tiny, black!65, align=center},
    ]
    \def\xb{1.5}\def\xe{5.1}\def\xr{8.75}\def\xs{11.85}\def\xg{14.6}
    \def\yh{-0.1}\def\ym{-1.15}\def\yl{-2.2}

    \node[head] at (\xb,0.9) {Base policy};
    \node[head] at (\xe,0.9) {Embodiment and controller};
    \node[head] at (\xr,0.9) {COMPASS residual RL};
    \node[head] at (\xs,0.9) {Specialists};
    \node[head] at (\xg,0.9) {Generalist};

    \node[base] (xm) at (\xb,\ym) {Pre-trained NVIDIA\\X-Mobility policy\\(frozen)};

    \node[emb, fill=hubcol] (he) at (\xe,\yh) {iw.hub\\differential drive controller (DDC)};
    \node[rl, fill=hubcol]  (hr) at (\xr,\yh) {Residual RL in\\NVIDIA's environments};
    \node[model, fill=hubcol] (hs) at (\xs,\yh) {iw.hub\\Specialist};

    \node[emb, fill=hmpcol] (me) at (\xe,\yl) {HMP\\omnidirectional swerve controller (OSC)};
    \node[rl, fill=hmpcol]  (mr) at (\xr,\yl) {Residual RL in\\NVIDIA's environments};
    \node[model, fill=hmpcol] (ms) at (\xs,\yl) {HMP\\Specialist};

    \node[model, fill=gencol, text width=1.9cm] (g) at (\xg,\ym) {Generalist};

    \draw[frozen] (xm.east) -- ++(0.35,0) |- (he.west);
    \draw[frozen] (xm.east) -- ++(0.35,0) |- (me.west);
    \draw[flow] (he) -- (hr);
    \draw[flow] (hr) -- (hs);
    \draw[flow] (me) -- (mr);
    \draw[flow] (mr) -- (ms);
    \draw[flow] (hs.east) -- ++(0.3,0) |- node[pos=0.25, right, note] {} (g.west);
    \draw[flow] (ms.east) -- ++(0.3,0) |- (g.west);
    \node[note, text width=1.4cm] at ($(g.south)+(0,-0.45)$) {distillation, as in COMPASS};
    \end{tikzpicture}
    \caption{How the COMPASS policies of the simulation benchmark are obtained. Both embodiments share the pre-trained NVIDIA X-Mobility policy as frozen base. Each embodiment is integrated with its own physics-based controller, and a COMPASS residual policy is trained for it in NVIDIA's environments of the original benchmark, which yields the iw.hub and HMP Specialists; the iw.hub Specialist is the \mdlNV{} model deployed on the real robot in Section~\ref{sec:real_world_benchmark}. All three policies are then evaluated in NVIDIA's environments and in the digital twin of the R\&D center (Figure~\ref{fig:iw_environment}). The Generalist is distilled from the two Specialists, following COMPASS~\citep{liu2025compass}.}
    \label{fig:sim_pipeline}
\end{figure}

%% file: tex/sections/real_world_benchmark.tex
\section{Real-World Navigation Benchmark}
\label{sec:real_world_benchmark}

This section presents a real-world navigation benchmark that assesses the
deployability and performance of COMPASS navigation policies on a physical
iw.hub in the Idealworks R\&D center. We evaluate the policies along several
dimensions: generalization to out-of-distribution scenarios and environments, deployability on a
real robot, onboard compute efficiency, safety and robustness.

COMPASS policies are trained entirely in simulation, so running them in the
real scenarios of the R\&D center measures their intrinsic sim-to-real
transfer. To make this transfer as robust as production use demands, leaving as
little as possible to chance when moving from simulation to the real facility,
we also strengthen the X-Mobility backbone of COMPASS: we generate
photorealistic variants of the training episodes with NVIDIA Cosmos and train
the backbone on a hybrid dataset of original and Cosmos-augmented episodes.

The benchmark compares four navigation systems (described under \emph{Models}):
the Idealworks navigation stack as the robotics baseline and three COMPASS
policies that add training on the R\&D center step by step: the COMPASS policy
trained only on NVIDIA's environments (the iw.hub Specialist of
Section~\ref{sec:sim_benchmark}), the COMPASS policy trained on simulated data from
the digital twin of the R\&D center, and the Cosmos-enhanced COMPASS policy
trained on a hybrid of that data and its sim-to-real variants. Each system is evaluated across the scenarios and with the metrics described
under \emph{Scenarios} and \emph{Metrics},
and we report the findings in Table~\ref{tab:rw_real}
for the R\&D center, and in Table~\ref{tab:rw_compute} for onboard compute.
The COMPASS policies are also evaluated on a \emph{digital-twin replica}: a
one-to-one copy of the static scenarios in the digital twin of the R\&D center
(Table~\ref{tab:rw_sim2real}). For \mdlNV{}, which has never seen the R\&D
center, the digital-twin replica measures how it generalizes to a new facility in
simulation, and the real runs how it transfers to the real world; for \mdlSim{}
and \mdlCE{}, comparing the two gives their sim-to-real gap.

\paragraph{Models.}
\label{sec:rw_models}
Three COMPASS models are compared; Figure~\ref{fig:rw_model_pipeline} shows how
each is obtained. They form a lineage that adds training on the target facility
step by step: \mdlNV{} has seen no data from the R\&D center, \mdlSim{} is
trained on simulated data from its digital twin, and \mdlCE{} additionally makes
that data photorealistic with Cosmos. \review{Comparing them shows what fine-tuning in
simulation on the target facility brings, and what sim-to-real data adds on top
of it.}

\input{tex/figures/model_pipeline}

\textbf{\mdlNV{}.} Exactly the iw.hub Specialist of
Section~\ref{sec:sim_benchmark}, evaluated there in simulation and deployed here
unchanged on the real robot: NVIDIA's pre-trained X-Mobility policy, kept frozen, with a
COMPASS residual trained for the iw.hub and its differential drive controller in
NVIDIA's environments (Figure~\ref{fig:sim_pipeline}). Neither its X-Mobility
base nor its residual has seen any data from the R\&D center.

\textbf{\mdlSim{}.} Starting from the NVIDIA checkpoints, we fine-tune the
X-Mobility world model on the random-exploration episodes of the digital twin
(described under \emph{Training data} below), which yields \wmSim{}, and then the policy on the
path-following episodes, which yields \ilSim{}
(Appendix~\ref{app:training_pipeline}). The COMPASS residual policy is then
trained from scratch on top of the frozen \ilSim{} in layout 04, a dense layout
of the digital twin described under \emph{Training data} below.

\textbf{\mdlCE{}.} The same fine-tuning on the hybrid dataset, in which every
simulated episode is followed by its Cosmos sim-to-real copy (50\% simulated,
50\% sim-to-real; Appendix~\ref{app:cosmos_augmentation}), yields \wmCE{} and
\ilCE{}, and the COMPASS residual policy is again trained from scratch on top
of the frozen \ilCE{}, in layout 04 and with the same setup as for \mdlSim{}.

\textbf{\mdlBaseline{}.} The robotics baseline is the Idealworks navigation
stack, the proprietary navigation system deployed on the iw.hub in production.
It is a LiDAR-based stack that plans routes with a global planner and follows
them with a local planner. It is evaluated in the real world on the same
scenarios, start poses and goals as the learned policies, but not in the
digital twin.

The prefix \textbf{IW.} marks a model whose X-Mobility base is fine-tuned from
the NVIDIA checkpoints on data from the digital twin of the Idealworks R\&D
center, and \textbf{IW.CE} one fine-tuned on the Cosmos-enhanced (CE) hybrid
version of that data; in both, the COMPASS residual is trained from scratch in
the digital twin. \mdlNV{}, without prefix, keeps NVIDIA's pre-trained
X-Mobility base and has its residual trained in NVIDIA's environments.
Appendix~\ref{app:training_pipeline} details the training stages,
hyperparameters and evaluations of all models mentioned.

\begin{figure}[!htbp]
    \centering
    \begin{subfigure}[b]{0.72\linewidth}
        \centering
        \includegraphics[width=\linewidth]{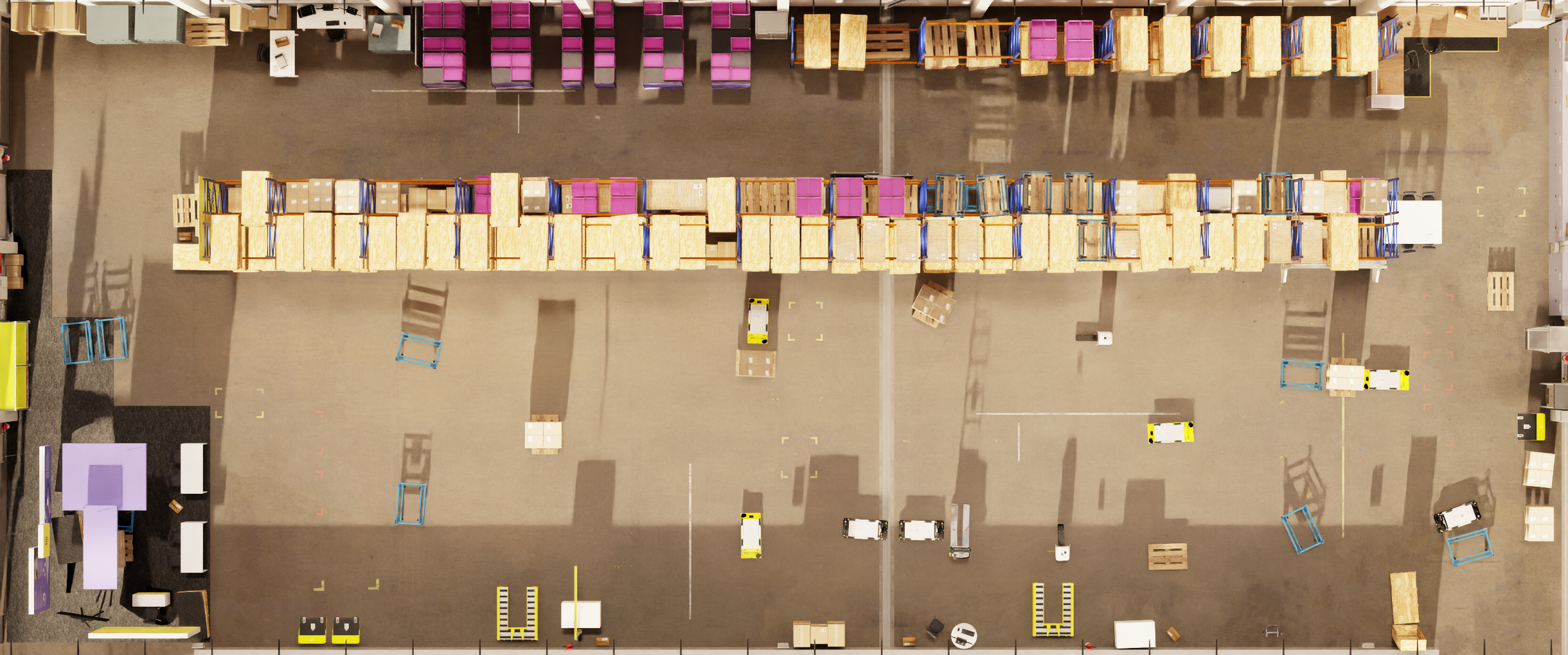}
        \caption{Layout 04, the COMPASS training layout of the IW models (top view)}
        \label{fig:layout04}
    \end{subfigure}
    \\[0.6em]
    \begin{subfigure}[b]{0.72\linewidth}
        \centering
        \includegraphics[width=\linewidth]{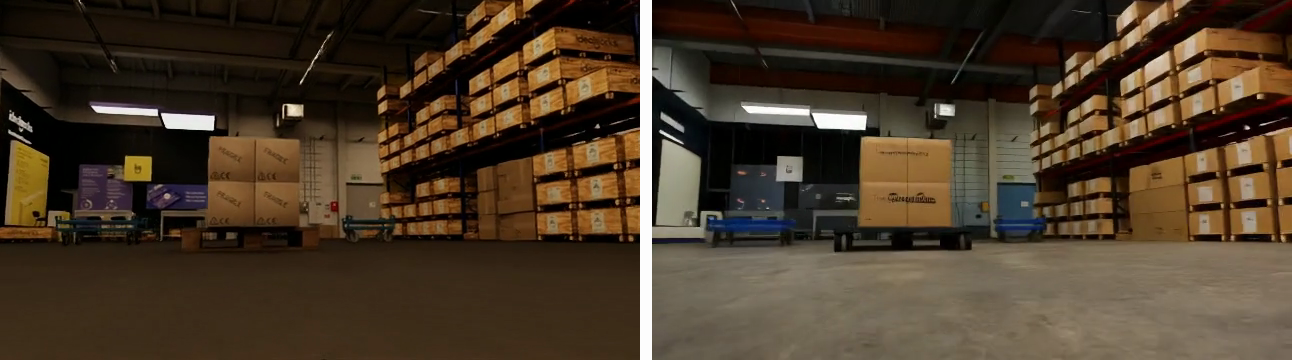}
        \caption{A simulated frame (left) and its Cosmos sim-to-real version (right)}
        \label{fig:layouts_stack_cosmos}
    \end{subfigure}
    \caption{Training data of the IW models. (a) Layout 04, a dense layout of the digital twin in which the COMPASS residual of the IW models is trained. (b) The Cosmos-enhanced dataset adds a photorealistic copy of every simulated episode. Four of the 16 layouts, one per density tier, are shown in Figure~\ref{fig:layouts_stack} (Appendix~\ref{app:training_datasets}).}
    \label{fig:rw_data}
\end{figure}

\paragraph{Training data.}
\label{sec:rw_data}
The IW models are trained on data generated in the digital twin of the R\&D
center. Rather than its single original arrangement, we build 16 layouts of the
R\&D center, each with its own floor objects (pallets, dollies, iw.hubs and
forklifts) at one of four density tiers, from near-empty to dense, and its own
lighting (Figure~\ref{fig:layouts_stack} in Appendix~\ref{app:training_datasets}). Layouts 14 and 15, one sparse and one
dense, are held out for testing. In these layouts we record with
MobilityGen~\citep{nvidia2025mobilitygen}, using the iw.hub and its own front
camera, 605 random-exploration episodes (76,284 frames) for the world model and
240 path-following episodes (56,535 frames) for the IL policy. Their semantic
labels include the iw.hub and the dolly, two classes that the released
X-Mobility model does not have and that we added to it
(Appendix~\ref{app:datasets_labels}). For the IW.CE
models, Cosmos Transfer~2.5 turns every simulated episode into a photorealistic,
frame-aligned copy, and the hybrid dataset appends these copies to the
simulated episodes (Figure~\ref{fig:layouts_stack_cosmos}).

The COMPASS residual of the IW models is trained online in one of these
layouts, layout 04 (Figure~\ref{fig:layout04}), chosen as a hard training
environment. It belongs to the
dense tier, with 36 objects (9 pallets, 5 box stacks, 8 dollies, 11 iw.hubs and
3 forklifts), so that only 185\,m$^2$, about half of the drivable area of the
empty R\&D center, remain drivable, and about 28\% of its training routes are
blocked by an obstacle across the width of the robot. Its lighting is dim, at
0.65 of the nominal brightness, with 9 of the 46 ceiling lights switched off.
Appendix~\ref{app:training_datasets} details the layouts and datasets,
Appendix~\ref{app:training_compass} the COMPASS training in layout 04, and
Appendix~\ref{app:cosmos_augmentation} the Cosmos augmentation.

\paragraph{Scenarios}
\label{sec:rw_scenarios}
Table~\ref{tab:rw_scenarios} lists the benchmark scenarios and
Figure~\ref{fig:rw_scenarios} shows their layouts. Each scenario is run over about 30 trials.
The static scenarios S1, S2, S3 and S5 are also recreated in the digital-twin
replica, with the same layout, obstacles, start poses and goals, and the COMPASS
policies are evaluated there as well. The dynamic scenario S4 is not replicated:
moving obstacles in the digital twin are not explored in this work. The
Idealworks navigation stack is benchmarked only in the real world.

\begin{table}[htbp]
    \centering
    \caption{Real-world benchmark scenarios. The static scenarios S1, S2, S3 and S5 are also run in the digital-twin replica.}
    \label{tab:rw_scenarios}
    \renewcommand{\arraystretch}{1.2}
    \footnotesize
    \begin{tabularx}{\textwidth}{l l X}
    \toprule
    \textbf{ID} & \textbf{Scenario} & \textbf{Description} \\
    \midrule
    S1 & Open-space traversal  & Point-to-point navigation across an empty floor. Layout: Figure~\ref{fig:scenario_s1}. \\
    S2 & Aisle navigation      & Traversal of narrow racking aisles. One side has racking, the other has a set of solid walls. Layout: Figure~\ref{fig:scenario_s2}. \\
    S3 & Static obstacles      & Route blocked by robots (iw.hubs) and dollies in a serpentine pattern. Layout: Figure~\ref{fig:scenario_s3}. \\
    S4 & Dynamic obstacles     & Open space navigation with moving robots (iw.hubs) crossing the path. Layout: Figure~\ref{fig:scenario_s4}. \\
    S5 & Far, occluded goal    & Goal beyond the range of the local policy and out of the robot's line of sight; requires the global planner integration (Appendix~\ref{app:global_planner}). Layout: Figure~\ref{fig:scenario_s5}. \\
    \bottomrule
    \end{tabularx}
\end{table}

\newlength{\scenariowidth}
\begin{figure}[p]
    \centering
    \captionsetup[subfigure]{skip=1pt}
    \ifreport\setlength{\scenariowidth}{0.43\textheight}\else\setlength{\scenariowidth}{0.448\textheight}\fi%
    \ifdim\scenariowidth>\linewidth\setlength{\scenariowidth}{\linewidth}\fi
    \begin{subfigure}{\linewidth}
        \centering
        \includegraphics[width=\scenariowidth]{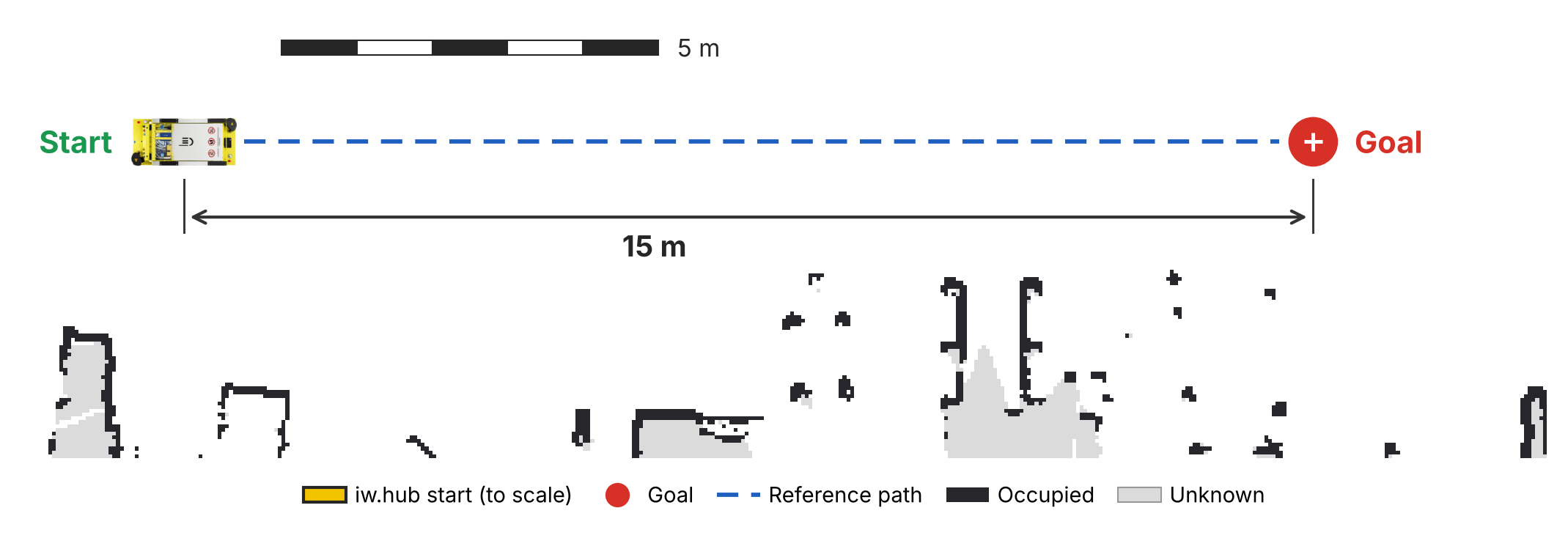}
        \caption{S1: open-space traversal}
        \label{fig:scenario_s1}
    \end{subfigure}
    \\[0.25em]
    \begin{subfigure}{\linewidth}
        \centering
        \includegraphics[width=\scenariowidth]{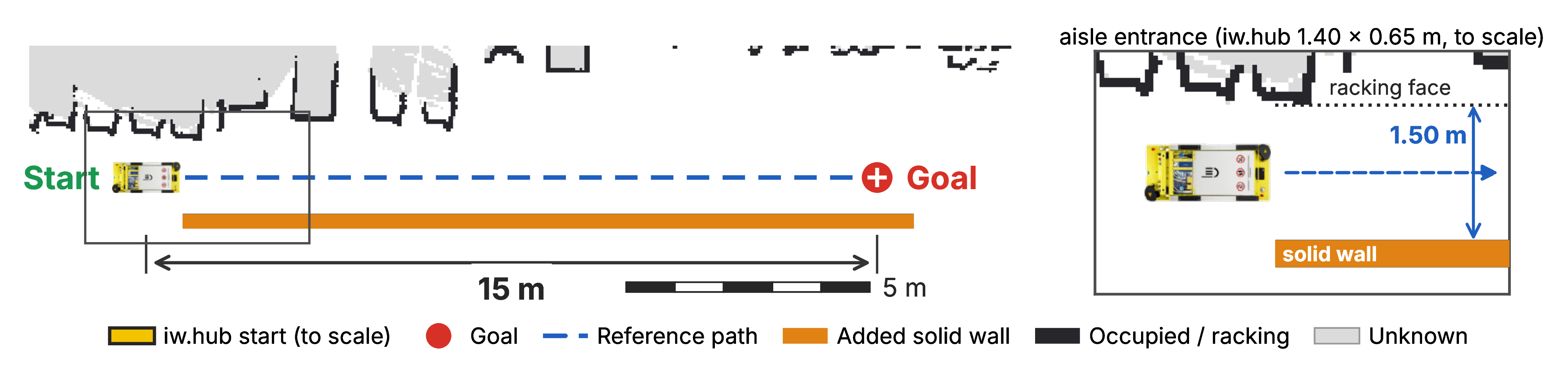}
        \caption{S2: aisle navigation}
        \label{fig:scenario_s2}
    \end{subfigure}
    \\[0.25em]
    \begin{subfigure}{\linewidth}
        \centering
        \includegraphics[width=\scenariowidth]{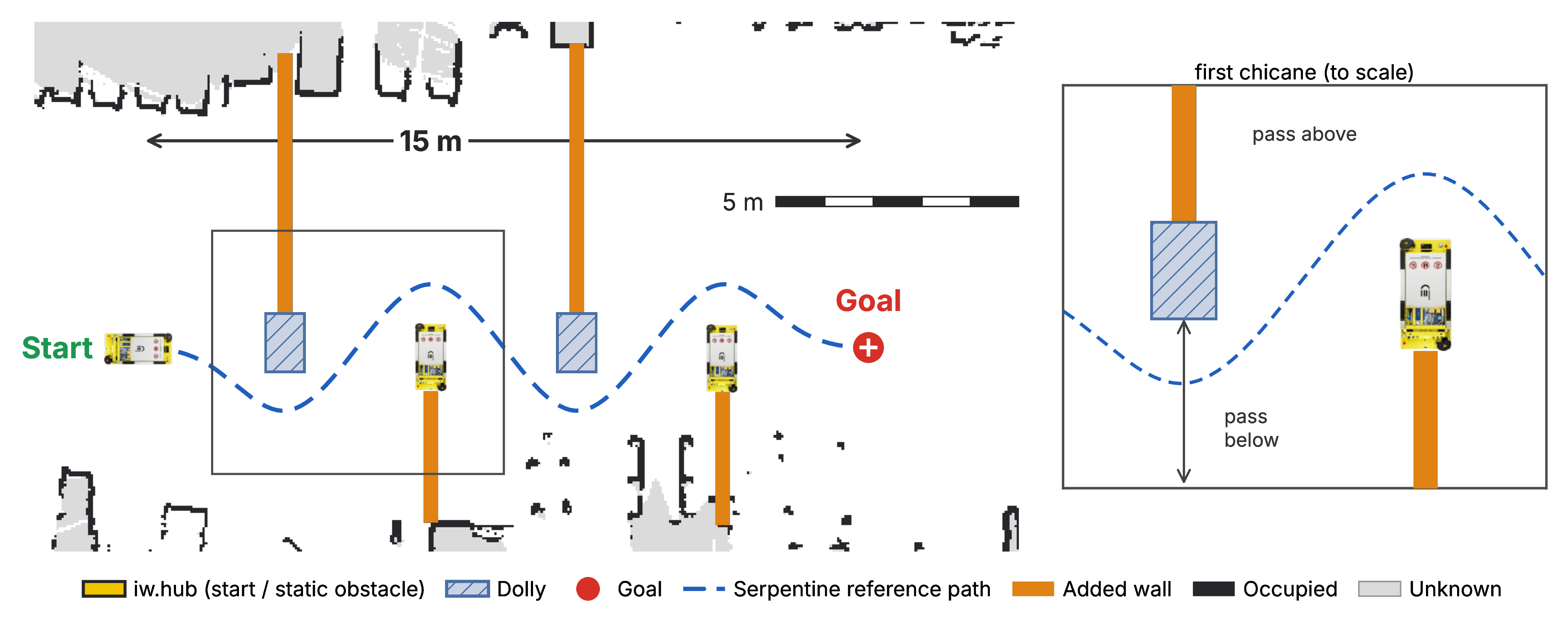}
        \caption{S3: static obstacles}
        \label{fig:scenario_s3}
    \end{subfigure}
    \\[0.25em]
    \begin{subfigure}{\linewidth}
        \centering
        \includegraphics[width=\scenariowidth]{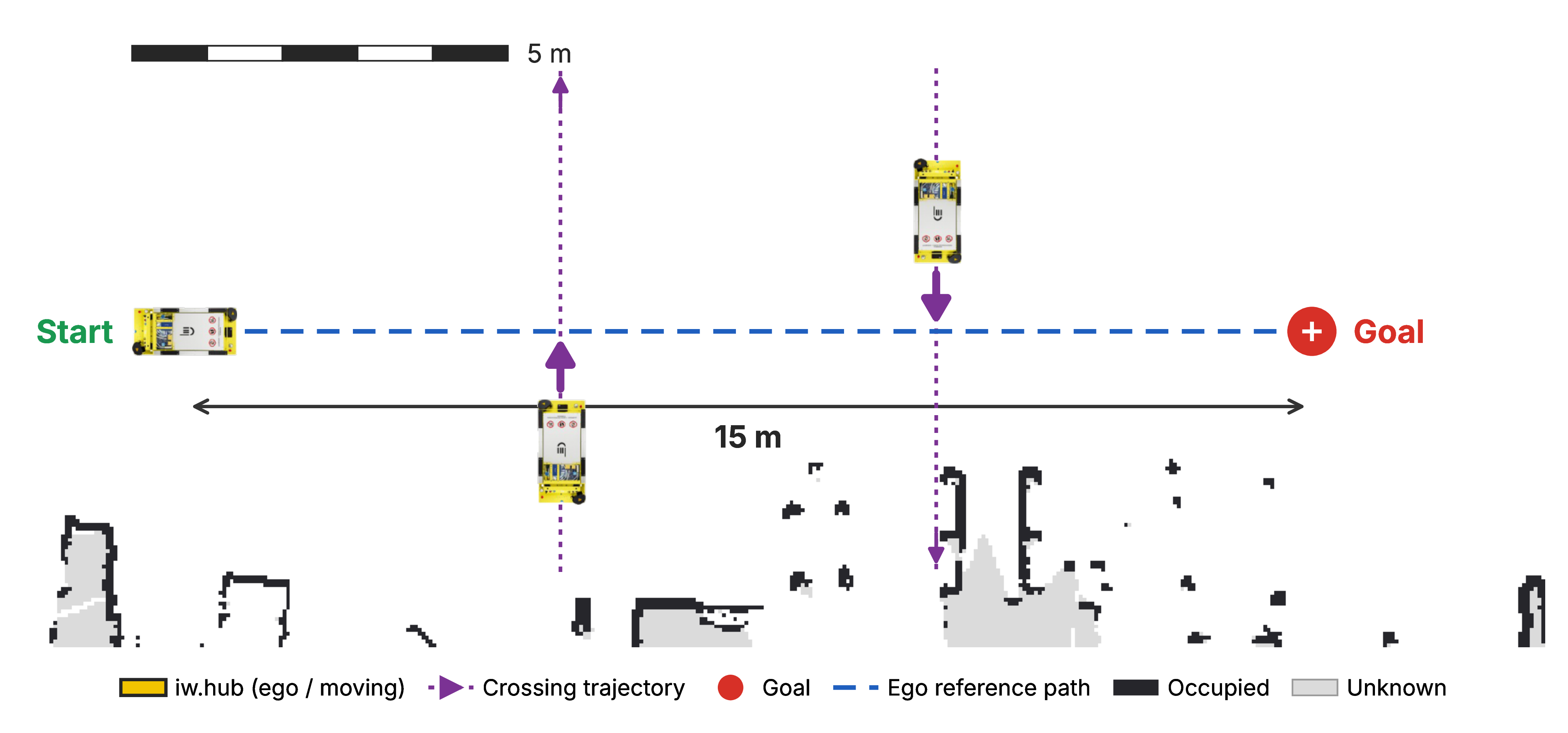}
        \caption{S4: dynamic obstacles}
        \label{fig:scenario_s4}
    \end{subfigure}
    \\[0.25em]
    \begin{subfigure}{\linewidth}
        \centering
        \includegraphics[width=\scenariowidth]{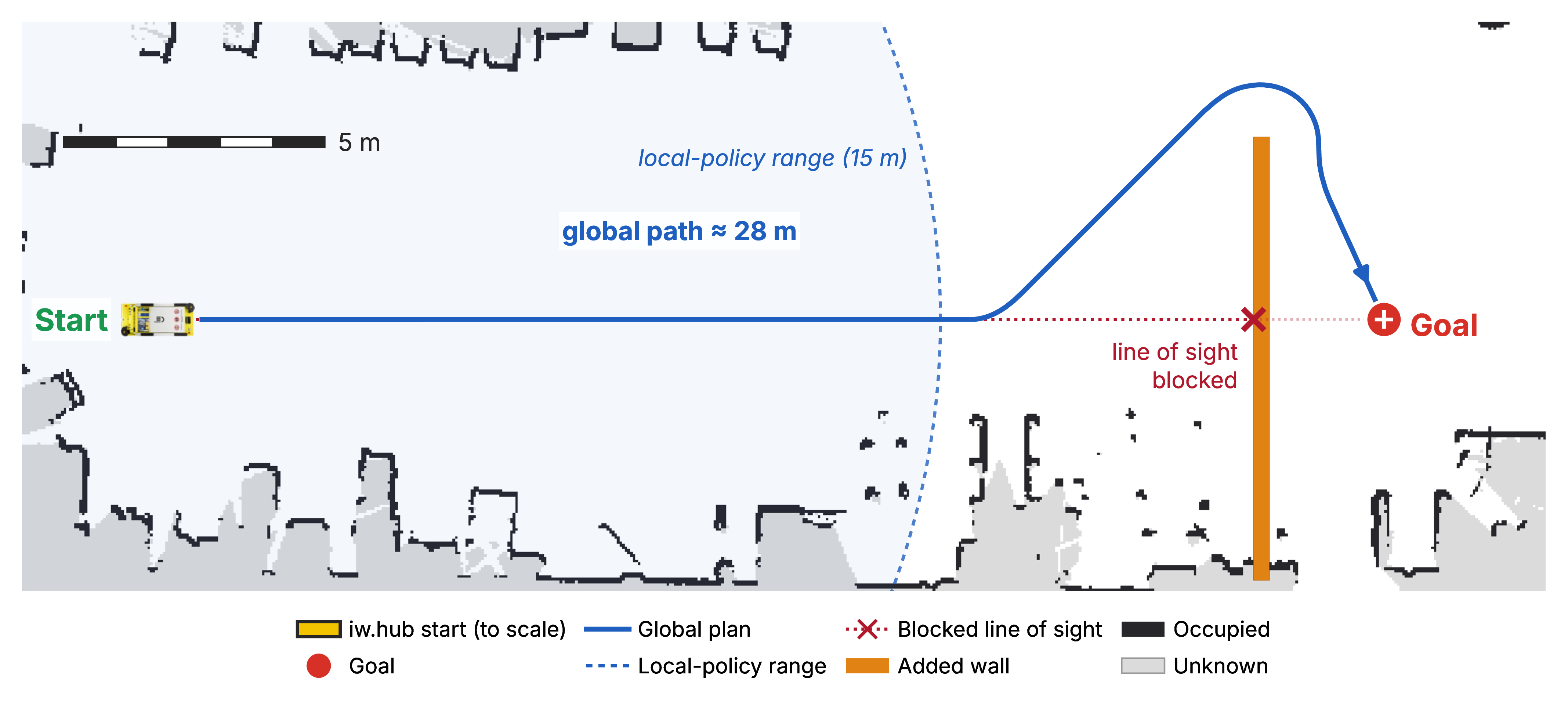}
        \caption{S5: far, occluded goal}
        \label{fig:scenario_s5}
    \end{subfigure}
    \caption{Layouts of the real-world benchmark scenarios in the R\&D center.}
    \label{fig:rw_scenarios}
\end{figure}

COMPASS is a local policy: it navigates towards a goal it can reach from its
current surroundings, but on its own it cannot plan a route to a goal that is
far away and hidden behind racks, walls or other occluders. For scenario S5 we
therefore integrated a global planner with the COMPASS stack and modified the
COMPASS ROS~2 node to accept its output.

Appendix~\ref{app:global_planner} describes the integration and the node
modifications. The global planner is implemented in the COMPASS ROS~2 node and
is therefore used only on the real robot, where S5 is run with \mdlCE{}, the
best-performing learned model on S1--S4, and with the Idealworks navigation
stack (Table~\ref{tab:rw_real}). \review{In the digital-twin replica, all three COMPASS
policies are run on S5 without the global planner. None of them reaches the
goal, although they cover most of the distance to it (Table~\ref{tab:rw_sim2real}
and Appendix~\ref{app:twin_replica}): the COMPASS policy alone, whatever its
training, cannot plan this far ahead around an occluding obstacle. This
observation from simulation is what motivated the global planner integration.}

\paragraph{Metrics}
\label{sec:rw_metrics}

Metrics are grouped into categories (Table~\ref{tab:rw_metrics}). Arrows give
the preferred direction. Unless stated otherwise, each metric is averaged over
all trials of a scenario. Interaction metrics are reported for scenarios with
dynamic obstacles only (S4); protective stops are reported for all scenarios.
A trial succeeds when the robot reaches the goal within 0.5\,m, with a linear
speed below 0.1\,m/s and an angular speed below 0.1\,rad/s. A collision ends
the trial as a failure, and so does a timeout when the robot is stuck: 2\,min in
general, longer in S3, where the robot may need more time to work its way
through the obstacles.

\begin{table}[htbp]
    \centering
    \caption{Real-world benchmark metrics by category.}
    \label{tab:rw_metrics}
    \renewcommand{\arraystretch}{1.2}
    \footnotesize
    \begin{tabularx}{\textwidth}{l l c c X}
    \toprule
    \textbf{Category} & \textbf{Metric} & \textbf{Unit} & & \textbf{Definition} \\
    \midrule
    \multirow{2}{*}{Task}
        & Success rate (SR)        & \%  & $\uparrow$   & Share of trials that reach the goal without collision or timeout. \\
        & Progress ratio (PR)      & \%  & $\uparrow$   & Fraction of the initial distance to the goal covered by the end of the trial. \\
    \midrule
    \multirow{3}{*}{Efficiency}
        & Weighted travel time (WTT) & s & $\downarrow$ & Travel time to the goal, conditioned on success and normalized by the success rate (Section~\ref{sec:sim_benchmark}). \\
        & Time to goal (TTG)       & s   & $\downarrow$ & Travel time of successful trials. \\
        & Path length ratio (PLR)  & --  & $\downarrow$ & Driven path length divided by the shortest collision-free path length. \\
    \midrule
    \multirow{2}{*}{Safety}
        & Collision count (CC)     & --  & $\downarrow$ & Number of collisions per trial. \\
        & Protective stops (PS)    & --  & $\downarrow$ & Number of protective stops triggered by the safety laser scanner per trial. \\
    \midrule
    \multirow{2}{*}{Smoothness}
        & Jerk (J)                 & m/s$^3$ & $\downarrow$ & Mean absolute jerk of the base along the trajectory. \\
        & Oscillation count (OC)   & --  & $\downarrow$ & Number of sign changes of the commanded angular velocity per trial. \\
    \midrule
    \multirow{2}{*}{\shortstack[l]{Interaction\\(dynamic\\scenarios)}}
        & Reaction time (RT)       & s   & $\downarrow$ & Time from a dynamic obstacle entering the robot's path to the robot slowing down or changing heading. \\
        & Recovery time (RcT)      & s   & $\downarrow$ & Time from the obstacle leaving the path to the robot regaining its nominal speed towards the goal. \\
    \midrule
    \multirow{4}{*}{Compute}
        & Inference latency        & ms  & $\downarrow$ & Time from sensor input to velocity command. \\
        & Control rate             & Hz  & $\uparrow$   & Frequency at which commands are issued. \\
        & CPU / GPU load           & \%  & $\downarrow$ & Mean utilization of the onboard compute while navigating. \\
        & Memory                   & GB  & $\downarrow$ & Peak memory used by the navigation system. \\
    \bottomrule
    \end{tabularx}
\end{table}

\paragraph{Onboard Deployment}
\label{sec:rw_deployment}

All learned policies run on the iw.hub's onboard computer, an NVIDIA Jetson AGX
Xavier, with no offboard compute. To run in real time on this hardware, the
COMPASS model is exported to ONNX and compiled into a TensorRT engine for the
Xavier, and inference is served by the COMPASS ROS~2 node on the robot. The
Xavier runs CUDA~11.4 and TensorRT~8.5.2, so the model is exported with ONNX
opset~16 for compatibility with this TensorRT version, and the engine is built in
FP16 precision. The resulting engine takes 177\,MB and runs at 5--8\,Hz on the
robot.
Appendix~\ref{app:jetson_deployment} details the target platform and the
conversion; Table~\ref{tab:rw_compute} reports the resulting onboard compute.

\paragraph{Results}
\label{sec:rw_results}

Table~\ref{tab:rw_real} reports all navigation systems on the iw.hub in the
R\&D center; scenario S5, which requires the global planner, is run only with
\mdlCE{} and the Idealworks navigation stack. Table~\ref{tab:rw_sim2real} compares the
success rate and weighted travel time of each COMPASS policy in the digital-twin
replica of S1, S2, S3 and S5 with its real-world runs; this table is a reduced
view, and Appendix~\ref{app:twin_replica} reports the results of all metrics in
the digital-twin replica. In S5, the replica runs are without the global planner
and the real runs with it.
Table~\ref{tab:rw_compute} reports onboard compute on the iw.hub.

\ifreport
\input{tex/tables/rw_real}
\else
\input{tex/tables/rw_real_whitepaper}
\fi

\begin{reviewblock}
\textbf{Task success and efficiency.} The scenarios separate the models more and
more as they get harder. In the open space of S1, all three COMPASS policies
reach the goal in every trial and differ only in how fast they get there:
\mdlCE{} needs 27.0\,s, against 38.2\,s for \mdlSim{} and 41.3\,s for \mdlNV{}.
In the aisles of S2, \mdlNV{} starts to fail (83.3\%), while \mdlSim{} (96.8\%)
and \mdlCE{} (100\%) remain reliable, and \mdlCE{} is the fastest system of the
benchmark there, \mdlBaseline{} included (13.5\,s against 16.2\,s). The static
obstacles of S3 are the decisive scenario. \mdlNV{} reaches the goal in only
6.7\% of the trials and covers less than half of the distance to it on average
(progress ratio, PR, of 46.2\%); \mdlSim{} reaches 54.8\%, and \mdlCE{} 83.3\%,
with a time to goal of its successful trials (TTG) of 35.0\,s, the same as
\mdlBaseline{} (34.9\,s). Because the weighted travel time (WTT) divides the
travel time by the success rate, it penalizes failed trials as well as slow
ones, which is why the WTT of \mdlNV{} in S3 reaches 4,998\,s. In the dynamic
scenario S4, \mdlCE{} again has the highest success rate (83.3\%) and the lowest
WTT (52.8\,s) of the learned policies. It recovers fastest once an obstacle has
left its path (recovery time, RcT, of 5.02\,s, close to the 4.19\,s of
\mdlBaseline{}), although it reacts later than the other systems (reaction
time, RT, of 6.21\,s against about 3\,s). With the global planner, \mdlCE{} also
reaches the far, occluded goal of S5 in every trial, within 6\% of the travel
time of \mdlBaseline{} (37.9\,s against 35.7\,s).

\textbf{Safety and smoothness.} No COMPASS policy collides in S1 and S2. In S3
and S4, collisions (collision count, CC) and protective stops of the safety laser
scanner (PS) appear for all of them. \mdlCE{} triggers far fewer protective stops
than \mdlNV{} (1.5 against 5.7 per trial on average over S1--S4) but more than
\mdlSim{} (0.7), which also collides least: \mdlSim{} is the most cautious
learned policy, while \mdlCE{} is the most successful and the fastest. \mdlCE{}
oscillates least: its commanded angular velocity changes sign 10.3 times per
trial on average (oscillation count, OC), against 15.1 for \mdlSim{} and 48.0 for
\mdlNV{}, whose path in S3 is almost three times longer than the shortest one
(path length ratio, PLR, of 2.83). The jerk of \mdlCE{} is slightly higher than
that of the other learned policies, but lower than that of \mdlBaseline{} in
every scenario except S2.

\textbf{Comparison with the Idealworks navigation stack.} \mdlBaseline{} is the
production navigation stack of the iw.hub, a mature system refined over many
deployments, and the reference the learned policies aim to approach. It reaches
the goal in every trial of every scenario, along almost straight paths (PLR of
1.00 to 1.04) and without collisions. It also starts from a stronger position:
it localizes in a prior map of the facility and perceives obstacles with LiDAR,
whereas COMPASS has no map and sees the scene through a single RGB camera. \mdlCE{} comes closest to it: on average
over S1--S4, it reaches 91.7\% of the goals against 100\%, and its WTT is 40\%
higher (33.9\,s against 24.2\,s), most of the gap coming from S3 and S4.
That a camera-only learned policy, trained in simulation, comes this close to a
LiDAR-based production stack is the strongest result of this benchmark.

\end{reviewblock}

\begin{table}[htbp]
    \centering
    \caption{Onboard compute of each navigation system on the iw.hub. \mdlNV{} and \mdlSim{} share the \mdlCE{} architecture and differ only in their training data, so their compute is the same as \mdlCE{}.}
    \label{tab:rw_compute}
    \renewcommand{\arraystretch}{1.2}
    \footnotesize
    \begin{tabular}{l ccccc}
    \toprule
    \textbf{Model}
      & \textbf{Latency (ms)} $\downarrow$
      & \textbf{Rate (Hz)} $\uparrow$
      & \textbf{CPU (\%)} $\downarrow$
      & \textbf{GPU (\%)} $\downarrow$
      & \textbf{Memory (GB)} $\downarrow$ \\
    \midrule
    \mdlCE      & 54 & 5--8 & 81.5 & 38.6 & 0.892 \\
    \mdlBaseline & 50 & 15--20 & 84.1 & 0.0 & 1.158 \\
    \bottomrule
    \end{tabular}
\end{table}

\begin{reviewblock}
\textbf{Onboard compute.} \mdlCE{} runs entirely on the Xavier with a latency
comparable to that of \mdlBaseline{} (54\,ms against 50\,ms), a lower peak memory
(0.89\,GB against 1.16\,GB) and a similar CPU load (Table~\ref{tab:rw_compute}).
Unlike \mdlBaseline{}, it uses the GPU (38.6\%), and it issues commands at
5--8\,Hz, against 15--20\,Hz.

\end{reviewblock}

\begin{table}[htbp]
    \centering
    \caption{Sim-to-real gap of the COMPASS policies: success rate and weighted travel time in the digital-twin replica of the static scenarios (Sim) and in the R\&D center (Real). In S5, the digital-twin replica is run without the global planner and the real world with it, only with \mdlCE{}. Best value per scenario and column in bold. All metrics of the digital-twin runs in Table~\ref{tab:rw_twin} (Appendix~\ref{app:twin_replica}).}
    \label{tab:rw_sim2real}
    \renewcommand{\arraystretch}{1.2}
    \footnotesize
    \begin{tabular}{ll cc cc}
    \toprule
    & & \multicolumn{2}{c}{\textbf{SR (\%)} $\uparrow$} & \multicolumn{2}{c}{\textbf{WTT (s)} $\downarrow$} \\
    \cmidrule(lr){3-4} \cmidrule(lr){5-6}
    \textbf{Scenario} & \textbf{Model} & Sim & Real & Sim & Real \\
    \midrule
    \multirow{3}{*}{S1}
        & \mdlNV & 90.3 & 100.0 & \textbf{9.4} & 41.3 \\
        & \mdlSim & \textbf{93.0} & 100.0 & 9.9 & 38.2 \\
        & \mdlCE & 92.0 & 100.0 & 10.9 & \textbf{27.0} \\
    \midrule
    \multirow{3}{*}{S2}
        & \mdlNV & 87.7 & 83.3 & 16.0 & 25.3 \\
        & \mdlSim & \textbf{99.7} & 96.8 & \textbf{14.4} & 17.1 \\
        & \mdlCE & 98.7 & \textbf{100.0} & 15.9 & \textbf{13.5} \\
    \midrule
    \multirow{3}{*}{S3}
        & \mdlNV & 51.3 & 6.7 & 28.0 & 4997.8 \\
        & \mdlSim & \textbf{72.0} & 54.8 & \textbf{21.6} & 110.9 \\
        & \mdlCE & 71.0 & \textbf{83.3} & 26.5 & \textbf{42.3} \\
    \midrule
    \multirow{3}{*}{S5}
        & \mdlNV & 0.0 & \na & \na & \na \\
        & \mdlSim & 0.0 & \na & \na & \na \\
        & \mdlCE & 0.0 & 100.0 & \na & 37.9 \\
    \bottomrule
    \end{tabular}
\end{table}

\begin{reviewblock}
\textbf{Sim-to-real gap.} In the digital-twin replica (Table~\ref{tab:rw_sim2real}),
\mdlSim{} and \mdlCE{} perform almost the same in every scenario, for example
72.0\% and 71.0\% in S3, as expected from two policies trained on the same
simulated layouts. On the real robot, they separate in S3, and \mdlNV{}, already
the weakest in simulation (51.3\%), falls to 6.7\%. Equal results in simulation and
different results in the real world isolate the effect of the Cosmos sim-to-real
data: it does not change how the policy behaves in simulation, but it makes it
robust to the gap between simulation and reality.
The digital-twin replica is a close, not exact, copy of the real scenarios:
object poses and start headings could only be reproduced approximately in the
real facility, while the replica was randomized more widely. Absolute simulated
and real success rates are therefore not directly comparable, and a higher real
than simulated rate (83.3\% against 71.0\% in S3) is not a negative sim-to-real
gap. What is comparable is the ordering of the models, since all of them ran on
the same scenarios with the same start and goal poses. Under these conditions,
the simulation-only model drops in the real facility (72.0\% to 54.8\%), while
the Cosmos-enhanced one does not.
In S5, no policy reaches the goal in the replica, without the global
planner, while \mdlCE{} reaches it in every real trial with it.

\textbf{Overall.} Training in the digital twin of the R\&D center, and making that
data photorealistic with Cosmos, turns COMPASS from a policy that rarely clears
obstacles in the real facility into one that approaches the production stack.
On average over S1--S4, \mdlCE{} raises the real-world success rate of learned
navigation from 65.3\% (\mdlNV{}) to 91.7\%, a gain of 26.3 points. The Cosmos
sim-to-real augmentation alone accounts for 8.8 of these points over \mdlSim{},
while it reduces the weighted travel time by 47\% (from 63.8\,s to 33.9\,s) and
the oscillation count by 32\%. In S3, the scenario that separates the policies
most, it raises the success rate from 54.8\% to 83.3\% (+28.5 points).
\end{reviewblock}


\FloatBarrier

%% file: tex/figures/model_pipeline.tex
\begin{figure}[htbp]
    \centering
    \colorlet{nvcol}{black!8}
    \colorlet{iwcol}{blue!10}
    \colorlet{cecol}{teal!14}
    \begin{tikzpicture}[
        font=\scriptsize,
        box/.style={draw=black!60, rounded corners=2pt, align=center,
                    minimum height=0.95cm, inner sep=2pt},
        data/.style={box, text width=3.4cm},
        stage/.style={box, text width=1.9cm},
        rl/.style={box, text width=2.5cm},
        model/.style={box, text width=2.1cm, line width=0.8pt, font=\scriptsize\bfseries},
        flow/.style={-{Latex[length=1.6mm]}, black!70},
        init/.style={-{Latex[length=1.6mm]}, dashed, black!60},
        head/.style={font=\scriptsize\bfseries, align=center, text width=2.4cm},
        note/.style={font=\tiny, black!65, align=center},
    ]
    \def\xd{1.8}\def\xw{5.25}\def\xi{7.85}\def\xr{11.05}\def\xm{14.05}
    \def\ynv{0}\def\yiw{-2.05}\def\yce{-3.35}

    \node[head] at (\xd,1.3) {Training data};
    \node[head] at (\xw,1.3) {Stage 1\\world model};
    \node[head] at (\xi,1.3) {Stage 2\\IL policy};
    \node[head] at (\xr,1.3) {Stage 3\\COMPASS residual RL};
    \node[head] at (\xm,1.3) {Model};

    \node[data, fill=nvcol]  (nvd) at (\xd,\ynv) {NVIDIA datasets and\\environments (no R\&D data)};
    \node[stage, fill=nvcol] (nvw) at (\xw,\ynv) {NVIDIA\\world model};
    \node[stage, fill=nvcol] (nvi) at (\xi,\ynv) {NVIDIA\\X-Mobility policy};
    \node[rl, fill=nvcol]    (nvr) at (\xr,\ynv) {Residual RL in NVIDIA\\environments};
    \node[model, fill=nvcol] (nvm) at (\xm,\ynv) {\mdlNV{}};

    \node[data, fill=iwcol]  (iwd) at (\xd,\yiw) {R\&D digital twin, 16 layouts\\random exploration: 605 ep.\\path following: 240 ep.};
    \node[stage, fill=iwcol] (iww) at (\xw,\yiw) {\wmSim{}};
    \node[stage, fill=iwcol] (iwi) at (\xi,\yiw) {\ilSim{}};
    \node[rl, fill=iwcol]    (iwr) at (\xr,\yiw) {Residual RL\\layout 04};
    \node[model, fill=iwcol] (iwm) at (\xm,\yiw) {\mdlSim{}};

    \node[data, fill=cecol]  (ced) at (\xd,\yce) {Same episodes + a Cosmos\\sim-to-real copy of each\\(50\,\% / 50\,\%)};
    \node[stage, fill=cecol] (cew) at (\xw,\yce) {\wmCE{}};
    \node[stage, fill=cecol] (cei) at (\xi,\yce) {\ilCE{}};
    \node[rl, fill=cecol]    (cer) at (\xr,\yce) {Residual RL\\layout 04};
    \node[model, fill=cecol] (cem) at (\xm,\yce) {\mdlCE{}};

    \foreach \r in {nv,iw,ce} {
        \draw[flow] (\r d) -- (\r w);
        \draw[flow] (\r w) -- (\r i);
        \draw[flow] (\r i) -- node[above, note] {frozen} (\r r);
        \draw[flow] (\r r) -- (\r m);
    }

    \begin{scope}[on background layer]
        \node[draw=black!40, dashed, rounded corners=3pt, inner sep=4pt, fit=(iww)(cew)] (fw) {};
        \node[draw=black!40, dashed, rounded corners=3pt, inner sep=4pt, fit=(iwi)(cei)] (fi) {};
        \node[draw=black!40, dashed, rounded corners=3pt, inner sep=4pt, fit=(iwr)(cer)] (fr) {};
    \end{scope}
    \draw[init] (nvw) -- node[right, note, text width=1.1cm, align=left] {initialize\\\& fine-tune} (fw);
    \draw[init] (nvi) -- node[right, note, text width=1.1cm, align=left] {initialize\\\& fine-tune} (fi);
    \node[note, below=1pt of fr] {trained from scratch};
    \node[note, below=1pt of fw] {fine-tuned};
    \node[note, below=1pt of fi] {fine-tuned};
    \end{tikzpicture}
    \caption{How the three COMPASS models are obtained, adding training on the R\&D center step by step. \mdlNV{}, the iw.hub Specialist of Section~\ref{sec:sim_benchmark}, is trained only on NVIDIA's data and environments. For \mdlSim{} and \mdlCE{}, the X-Mobility world model and IL policy are initialized from the NVIDIA checkpoints and fine-tuned on data recorded with the iw.hub in the digital twin of the R\&D center, without (\mdlSim{}) or with (\mdlCE{}) a Cosmos sim-to-real copy of every episode, and the COMPASS residual policy is trained from scratch in the digital twin on top of the frozen IL policy.}
    \label{fig:rw_model_pipeline}
\end{figure}

%% file: tex/tables/rw_real_whitepaper.tex
\begin{table}[htbp]
    \centering
    \caption{Real-world benchmark on the iw.hub in the R\&D center. Scenario S5 (far, occluded goal) uses the global planner integration and is run only with \mdlCE{} and \mdlBaseline{}. \mdlBaseline{}, the production navigation system, is shaded as the reference the learned policies aim to approach; the best COMPASS model per scenario and metric is in bold. $^{\dagger}$Dynamic scenarios only; n/a otherwise.}
    \label{tab:rw_real}
    \renewcommand{\arraystretch}{1.2}
    \footnotesize
    \begin{adjustbox}{max width=\textwidth}
    \begin{tabular}{ll cc ccc cc cc cc}
    \toprule
    & & \multicolumn{2}{c}{\textbf{Task}} & \multicolumn{3}{c}{\textbf{Efficiency}}
    & \multicolumn{2}{c}{\textbf{Safety}} & \multicolumn{2}{c}{\textbf{Smoothness}} & \multicolumn{2}{c}{\textbf{Interaction}$^{\dagger}$} \\
    \cmidrule(lr){3-4} \cmidrule(lr){5-7} \cmidrule(lr){8-9} \cmidrule(lr){10-11} \cmidrule(lr){12-13}
    \textbf{Scenario} & \textbf{Model}
      & SR (\%) $\uparrow$ & PR (\%) $\uparrow$
      & WTT (s) $\downarrow$ & TTG (s) $\downarrow$ & PLR $\downarrow$
      & CC $\downarrow$ & PS $\downarrow$
      & J (m/s$^3$) $\downarrow$ & OC $\downarrow$
      & RT (s) $\downarrow$ & RcT (s) $\downarrow$ \\
    \midrule
    \multirow{4}{*}{S1}
        & \cellcolor{black!7}\mdlBaseline & \cellcolor{black!7}100.0 & \cellcolor{black!7}100.0 & \cellcolor{black!7}17.9 & \cellcolor{black!7}17.9 & \cellcolor{black!7}1.00 & \cellcolor{black!7}0.00 & \cellcolor{black!7}0.00 & \cellcolor{black!7}1.06 & \cellcolor{black!7}1.4 & \cellcolor{black!7}\na & \cellcolor{black!7}\na \\
        & \mdlNV & 100.0 & 100.0 & 41.3 & 41.3 & 1.15 & 0.00 & \textbf{0.00} & \textbf{0.72} & 23.8 & \na & \na \\
        & \mdlSim & 100.0 & 100.0 & 38.2 & 38.2 & \textbf{1.02} & 0.00 & \textbf{0.00} & 1.08 & \textbf{6.3} & \na & \na \\
        & \mdlCE & 100.0 & 100.0 & \textbf{27.0} & \textbf{27.0} & 1.13 & 0.00 & 0.10 & 1.00 & 11.1 & \na & \na \\
    \midrule
    \multirow{4}{*}{S2}
        & \cellcolor{black!7}\mdlBaseline & \cellcolor{black!7}100.0 & \cellcolor{black!7}100.0 & \cellcolor{black!7}16.2 & \cellcolor{black!7}16.2 & \cellcolor{black!7}1.00 & \cellcolor{black!7}0.00 & \cellcolor{black!7}0.00 & \cellcolor{black!7}1.10 & \cellcolor{black!7}1.3 & \cellcolor{black!7}\na & \cellcolor{black!7}\na \\
        & \mdlNV & 83.3 & 91.1 & 25.3 & 21.1 & 1.10 & 0.00 & 0.34 & 0.78 & 24.8 & \na & \na \\
        & \mdlSim & 96.8 & 96.8 & 17.1 & 16.5 & \textbf{1.00} & 0.00 & 0.26 & \textbf{0.71} & 19.8 & \na & \na \\
        & \mdlCE & \textbf{100.0} & \textbf{100.0} & \textbf{13.5} & \textbf{13.5} & 1.11 & 0.00 & \textbf{0.06} & 1.50 & \textbf{7.1} & \na & \na \\
    \midrule
    \multirow{4}{*}{S3}
        & \cellcolor{black!7}\mdlBaseline & \cellcolor{black!7}100.0 & \cellcolor{black!7}100.0 & \cellcolor{black!7}34.9 & \cellcolor{black!7}34.9 & \cellcolor{black!7}1.02 & \cellcolor{black!7}0.00 & \cellcolor{black!7}0.00 & \cellcolor{black!7}1.89 & \cellcolor{black!7}4.5 & \cellcolor{black!7}\na & \cellcolor{black!7}\na \\
        & \mdlNV & 6.7 & 46.2 & 4997.8 & 277.7 & 2.83 & 0.28 & 20.60 & 1.20 & 91.1 & \na & \na \\
        & \mdlSim & 54.8 & 67.2 & 110.9 & 61.0 & 1.14 & \textbf{0.00} & \textbf{1.70} & \textbf{1.09} & 18.6 & \na & \na \\
        & \mdlCE & \textbf{83.3} & \textbf{87.3} & \textbf{42.3} & \textbf{35.0} & \textbf{1.07} & 0.24 & 4.41 & 1.61 & \textbf{12.3} & \na & \na \\
    \midrule
    \multirow{4}{*}{S4}
        & \cellcolor{black!7}\mdlBaseline & \cellcolor{black!7}100.0 & \cellcolor{black!7}100.0 & \cellcolor{black!7}27.6 & \cellcolor{black!7}27.6 & \cellcolor{black!7}1.03 & \cellcolor{black!7}0.00 & \cellcolor{black!7}0.94 & \cellcolor{black!7}2.42 & \cellcolor{black!7}3.4 & \cellcolor{black!7}3.14 & \cellcolor{black!7}4.19 \\
        & \mdlNV & 71.4 & 86.3 & 153.1 & 142.2 & 2.08 & \textbf{0.07} & 1.71 & \textbf{1.14} & 52.3 & 2.96 & 12.00 \\
        & \mdlSim & 80.0 & 92.9 & 89.0 & 71.2 & \textbf{1.18} & 0.10 & \textbf{0.75} & 1.27 & 15.5 & \textbf{2.72} & 8.67 \\
        & \mdlCE & \textbf{83.3} & \textbf{93.8} & \textbf{52.8} & \textbf{44.0} & 1.34 & 0.08 & 1.29 & 1.37 & \textbf{10.7} & 6.21 & \textbf{5.02} \\
    \midrule
    \multirow{2}{*}{S5}
        & \cellcolor{black!7}\mdlBaseline & \cellcolor{black!7}100.0 & \cellcolor{black!7}100.0 & \cellcolor{black!7}35.7 & \cellcolor{black!7}35.7 & \cellcolor{black!7}1.04 & \cellcolor{black!7}0.00 & \cellcolor{black!7}0.00 & \cellcolor{black!7}2.40 & \cellcolor{black!7}3.0 & \cellcolor{black!7}\na & \cellcolor{black!7}\na \\
        & \mdlCE & 100.0 & 100.0 & 37.9 & 37.9 & 1.09 & 0.00 & 0.00 & 1.03 & 8.9 & \na & \na \\
    \bottomrule
    \end{tabular}
    \end{adjustbox}
\end{table}

%% file: tex/sections/conclusion.tex
\section{Conclusion and Future Work}
\label{sec:conclusion}

This work extended the COMPASS benchmark for learned navigation to two Idealworks
embodiments and the digital twin of the Idealworks R\&D center, and evaluated
COMPASS policies on a physical iw.hub against the Idealworks navigation stack.
\review{In simulation, both embodiments navigated across all five environments,
and policies trained only in NVIDIA's environments transferred to the digital
twin of the R\&D center. On the real robot, training in the digital twin and
augmenting its data with Cosmos sim-to-real episodes raised the success rate of
learned navigation from 65.3\% to 91.7\% over S1--S4 and nearly halved the
weighted travel time compared with simulation-only training. The resulting
\mdlCE{} came within 8.3 points of the Idealworks navigation stack while running
entirely on the iw.hub's Jetson AGX Xavier, and with the global planner
integration it reached the far, occluded goal of S5 in every trial. The
digital-twin replica showed where this gain comes from: \mdlSim{} and \mdlCE{}
performed almost the same in simulation and separated only on the real robot, so
the Cosmos sim-to-real data made the policy robust to the gap between simulation
and reality rather than changing its behavior in simulation.}

Overall, this work shows that the NVIDIA sim-to-real ecosystem, combining
digital twins, X-Mobility, Cosmos and COMPASS, can take a learned navigation
policy from simulation to an industrial mobile robot and close most of the gap
to a production navigation stack engineered for that robot. With learned
navigation established as feasible, the next question is its ROI justification.
A tuned traditional stack such as the Idealworks navigation stack remains a
strong incumbent on the robots it was engineered for, so the clearest value of
learned navigation lies where no such stack exists yet, as when a fleet adopts a
new embodiment. A natural next step is therefore a cost and effort study on the
HMP, comparing the engineering effort of adapting the Idealworks navigation
stack with that of transferring a learned policy across embodiments, to quantify
the return on investment at fleet scale. Such a study would build on the HMP
Specialist and Generalist of Section~\ref{sec:sim_benchmark}, which first need
to be validated on the physical HMP to confirm that cross-embodiment transfer
holds in the real world.

Beyond this study, several steps remain before production use: extending the
benchmark to more facilities and robots; meeting industrial safety requirements
such as ISO~3691-4 for a camera-based local planner, together with the safety
laser scanner; and a hybrid integration alongside the production global planner,
started in Appendix~\ref{app:global_planner}, which would enable long-term
deployment and fine-tuning on logged data.

Such an integration would also benefit from a policy that makes its
understanding of the scene explicit. COMPASS currently maps camera images
directly to velocity commands, without exposing where the obstacles it reacts
to are. Extending it with an obstacle-sensing capability, which detects
obstacles and localizes them relative to the robot, would allow the global
planner to replan around obstacles the policy has perceived, and would make the
policy's decisions more explainable, which supports both debugging and the
safety assessment.

Finally, the pipeline of this work, from data generation to deployment, is well
suited to agentic workflows. COMPASS already provides agent
skills\footnote{\url{https://github.com/NVlabs/COMPASS/tree/main/.claude/skills}}
to train and evaluate residual policies, onboard a new embodiment and diagnose a
broken setup. The steps of this work could be contributed as further skills:
generating randomized layouts of a new facility, recording X-Mobility data with
the target robot and camera, building the Cosmos-enhanced dataset, fine-tuning
X-Mobility with new semantic classes, adding a physics-based controller, and
deploying the policy on a Jetson. Chained by an agent, they would bring a new
robot or site from its digital twin to an onboard policy with little manual
effort, which would also lower the engineering effort that the cost and effort
study weighs.

%% file: tex/appendix/appendix_a_controller_type_study.tex
\section{Controller Type Study}
\label{app:controller_study}

The released COMPASS stack uses a kinematic controller. To deploy on our
embodiments we adapted a physics-based differential drive controller for the
iw.hub. This appendix compares the two formulations:

\paragraph{Non-Holonomic Perfect Controller (NHPC).}
A kinematic, teleportation-based controller that directly updates the robot
pose using the commanded linear and angular velocities, bypassing physical
actuation and joint-level dynamics.

\paragraph{Differential Drive Controller (DDC).}
A physics-based controller that converts linear and angular velocity commands
into wheel-joint velocity targets. Motion is generated through simulated
actuator dynamics subject to acceleration and deceleration constraints.

\subsection{Evaluation Results}
\label{sec:controller_ablation}

We compare the Differential Drive Controller (DDC) against the
Non-Holonomic Perfect Controller (NHPC). Both variants use the same policy
architecture, observations, reward formulation, and general training
configuration, but differ in how linear and angular velocity commands are
executed.

The DDC clamps and ramps the commanded velocities according to acceleration
and deceleration limits, then converts them into left and right wheel-velocity
targets using differential-drive kinematics. Motion is generated through
physics-based wheel actuation. In contrast, the NHPC integrates the commanded
velocities and directly teleports the robot root state to the resulting pose,
bypassing wheel actuation and joint-level dynamics.

Table~\ref{tab:controller_ablation} compares the evaluation performance of
the two controller formulations across all environments. The corresponding
training curves and an analysis of their learning behavior are given in
Appendix~\ref{app:controller_training}.

\begin{table}[H]
\centering
\caption{Controller type ablation.}
\label{tab:controller_ablation}

\renewcommand{\arraystretch}{1.2}
\footnotesize

\begin{adjustbox}{max width=\textwidth}
\begin{tabular}{l cc cc cc cc cc}
\toprule

\textbf{Controller} &
\multicolumn{2}{c}{\textbf{Warehouse M-R}} &
\multicolumn{2}{c}{\textbf{Warehouse S-R}} &
\multicolumn{2}{c}{\textbf{Office}} &
\multicolumn{2}{c}{\textbf{Combined M-R}} &
\multicolumn{2}{c}{\textbf{R\&D center (twin)}} \\

\cmidrule(lr){2-3}
\cmidrule(lr){4-5}
\cmidrule(lr){6-7}
\cmidrule(lr){8-9}
\cmidrule(lr){10-11}

&
SR (\%) & WTT (s) &
SR (\%) & WTT (s) &
SR (\%) & WTT (s) &
SR (\%) & WTT (s) &
SR (\%) & WTT (s) \\

\midrule

DDC
& 85 & 6734.8
& 76 & 5495.6
& 58 & 5617.7
& 76 & 6801.8
& 82 & 6402.8 \\

NHPC
& 86 & 6654.5
& 78 & 5851.2
& 67 & 6935.9
& 78 & 7139.2
& 82 & 7008.0 \\

\bottomrule
\end{tabular}
\end{adjustbox}

\end{table}

The NHPC achieves slightly higher success rates in four environments and
matches the DDC in the digital twin of the R\&D center. The largest success-rate
difference is observed in the Office environment, where the NHPC achieves
67\% compared with 58\% for the DDC.

However, the DDC produces a lower WTT in four of the five environments,
indicating more efficient successful trajectories under the weighted travel
time metric. The only exception is the Warehouse Multi-Rack environment,
where the NHPC obtains a slightly lower WTT.

Despite the success-rate differences, the DDC is selected for the final
COMPASS Specialist because it models wheel actuation, acceleration limits,
and robot dynamics more faithfully. This provides a more realistic control
interface and reduces the discrepancy between simulated policy commands and
the behavior of a physically actuated differential-drive robot.

\FloatBarrier
\subsection{Training Results}
\label{app:controller_training}

This section compares the training behavior of the DDC and NHPC variants
over 1,000 iterations. Both use the same policy architecture, environment
configuration, observation space, and reward formulation. Dotted lines show
the signals smoothed with a moving average, and the shaded bands show
$\pm 1$ standard deviation of the raw signal over the same window.

\captionsetup{
    font=small,
    labelfont=bf,
    skip=4pt
}

\begin{figure}[H]
\centering
\includegraphics[width=0.85\textwidth]{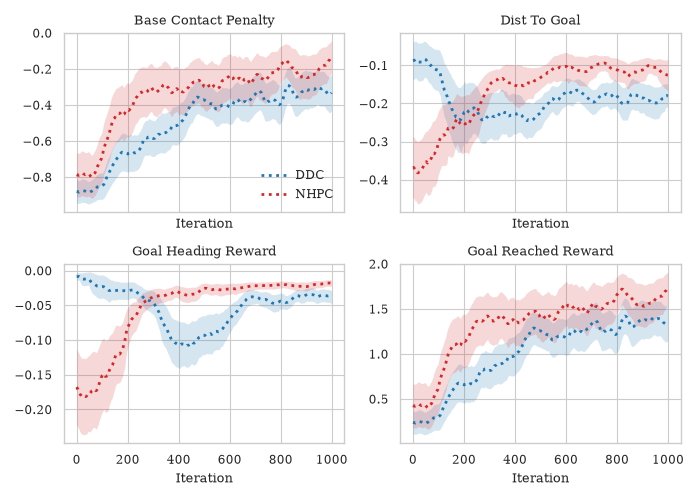}
\caption{Episode reward components for DDC and NHPC.}
\label{fig:controller_rewards}
\end{figure}

\begin{figure}[H]
\centering
\includegraphics[width=0.85\textwidth]{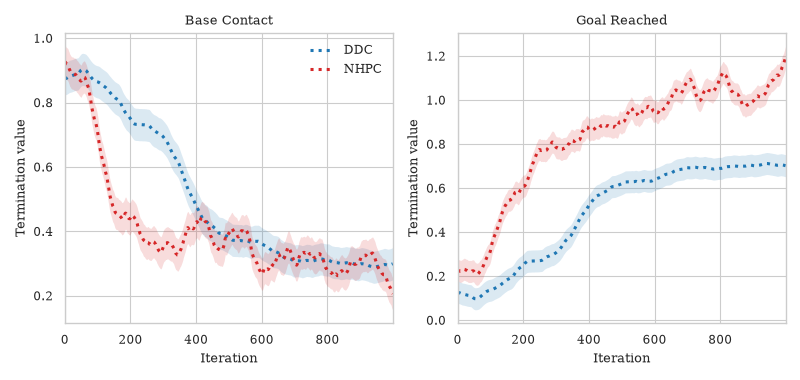}
\caption{Episode termination metrics for DDC and NHPC.}
\label{fig:controller_terminations}
\end{figure}

\begin{figure}[H]
\centering
\includegraphics[width=0.85\textwidth]{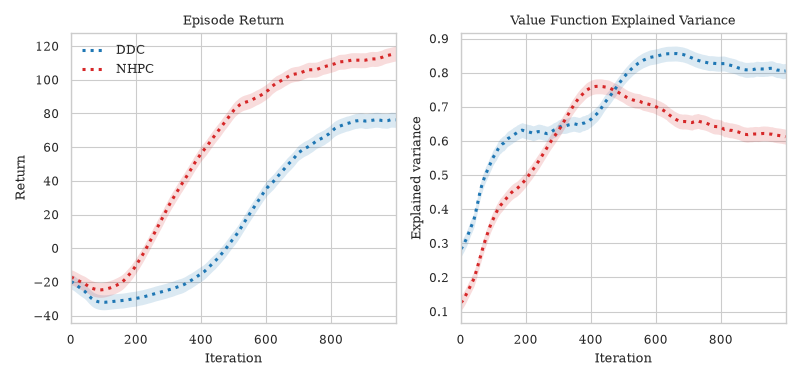}
\caption{Mean episode return and value-function explained variance for
DDC and NHPC.}
\label{fig:controller_ppo}
\end{figure}

\FloatBarrier

\subsubsection{Analysis}
\label{app:controller_training_analysis}

\paragraph{NHPC.}
The NHPC learns faster across all metrics. Its return starts rising after
approximately 150 iterations and reaches about 115, while base-contact
terminations drop sharply within the first 150 iterations. Heading and
distance rewards converge early, since direct pose updates make
corrections take effect immediately. However, its explained variance
peaks near 0.76 and then declines to approximately 0.6, suggesting the
critic struggles to track the rapidly shifting returns.

\paragraph{DDC.}
The DDC converges more gradually, as expected under physics-based
actuation and acceleration limits. Its return becomes positive around
iteration 500 and plateaus near 76, and its progress rewards dip
temporarily before recovering as the policy learns more assertive motion.
Despite the slower start, base-contact terminations match the NHPC after
iteration 400, and the explained variance stabilizes around 0.8,
indicating a more consistent value-function fit.

Overall, the NHPC is easier to optimize, while the DDC learns a stable
policy under a realistic control interface. Note that each DDC iteration
uses 350 rollout steps versus 256 for the NHPC, so the curves compare
iterations rather than equal numbers of environment steps.

\FloatBarrier

%% file: tex/appendix/appendix_b_omnidirectional_controller.tex
\section{Omnidirectional Controller Implementation}
\label{sec:omni_directional_controller}

COMPASS was originally demonstrated on differential-drive and legged
embodiments, all of which follow non-holonomic or gait-based motion models.
To extend its usability to a broader class of mobile platforms, we deploy
the framework on the HMP, an omnidirectional robot capable of independent
translation along both planar axes and simultaneous rotation. Supporting
this embodiment required a dedicated low-level controller, since none of
the existing COMPASS action interfaces can express holonomic motion or
actuate steerable drive modules. We therefore developed a physics-based
Omnidirectional Swerve Controller (OSC) from scratch. It converts the
policy's body-twist commands into steering-angle and wheel-velocity
targets while respecting the platform's kinematic and dynamic limits. This
appendix describes the platform model, the controller formulation, and the
design iterations that led to the final version.

\subsection{Platform Model}
\label{app:omni_platform}

Unlike the iw.hub, the HMP is holonomic. Its chassis has two independently
steerable drive modules mounted at diagonally opposite corners, at
$(0.205, -0.205)$\,m and $(-0.205, 0.205)$\,m in the base frame, together
with four fully passive casters. Each module has a drive joint under
velocity control and a steering joint under position control.
Table~\ref{tab:omni_actuators} lists the actuator models.

\begin{table}[h]
\centering
\caption{HMP actuator models.}
\label{tab:omni_actuators}
\renewcommand{\arraystretch}{1.2}
\footnotesize
\begin{tabular}{lcccc}
\toprule
\textbf{Joint group} & \textbf{Mode} & $\mathbf{K_p}$ & $\mathbf{K_d}$ &
\textbf{Vel.\ / effort limit} \\
\midrule
Drive ($\times 2$)    & Velocity & 0  & 500 & 32 rad/s / 40 N$\cdot$m \\
Steering ($\times 2$) & Position & 65 & 20  & 20 rad/s / 200 N$\cdot$m \\
Casters ($\times 4$)  & Passive  & 0  & 0   & -- / 0 N$\cdot$m \\
\bottomrule
\end{tabular}
\end{table}

\subsection{Command Interface}
\label{app:omni_interface}

The OSC operates on a planar body-frame twist
$\mathbf{u}^\star = (v_x^\star, v_y^\star, \omega_z^\star)$, comprising
longitudinal velocity, lateral velocity, and yaw rate, which together span
the full holonomic motion space of the platform. The policy action is
retained as a 6-dimensional twist to preserve compatibility with the
COMPASS policy head, of which only the three planar components are
actuated. To ensure dynamically feasible commands, each component is first
saturated to the platform velocity bounds and subsequently slew-rate
limited at every physics step $\Delta t$:
\begin{equation}
  \tilde{v}_k = \tilde{v}_{k-1} +
  \mathrm{clip}\!\left(v^\star - \tilde{v}_{k-1},\,
  -a_{\max}\Delta t,\, a_{\max}\Delta t\right).
\end{equation}

\subsection{Swerve Inverse Kinematics}
\label{app:omni_ik}

For module $i$ at position $(r_{x,i}, r_{y,i})$, the desired ground velocity
is
\begin{equation}
  v_{x,i} = \tilde{v}_x - \tilde{\omega}_z\, r_{y,i}, \qquad
  v_{y,i} = \tilde{v}_y + \tilde{\omega}_z\, r_{x,i},
\end{equation}
which gives a module speed $s_i = \lVert (v_{x,i}, v_{y,i}) \rVert$ and a
desired steering angle
$\theta_i^\star = \operatorname{atan2}(v_{y,i}, v_{x,i}) + \theta_0$, where
$\theta_0$ is a calibration offset. All angular differences below are
wrapped to $(-\pi, \pi]$ by $\mathrm{wrap}(\cdot)$. Let $\theta_i$ denote the
measured steering angle and $r_w$ the wheel radius.

\subsection{Design Iterations}
\label{app:omni_iterations}

The final OSC is the result of three design iterations. Each one addressed
failure modes observed in the previous version.

\paragraph{Naive Swerve Control.}
The first version applies the IK outputs directly, with
$\theta_i^{\mathrm{cmd}} = \theta_i^\star$ and
$\dot{q}_i^{\mathrm{cmd}} = s_i / r_w$. This exposed three problems. First,
the module heading is undefined as $s_i \to 0$, and $\operatorname{atan2}$
snaps the wheel back to zero at every stop. Second, velocity reversals
require up to $180^\circ$ of steering travel. Third, the wheel is driven at
full speed while still misaligned, which produces lateral scrubbing against
the ground.

\paragraph{Fully Constrained Control.}
The second version adds three safeguards. The last valid heading is held
whenever $s_i < s_{\min}$. Each module takes the shortest steering path: with
$e_i = \mathrm{wrap}(\theta_i^\star - \theta_i)$, if $|e_i| > \pi/2$ the
module targets $\theta_i^\star + \pi$ and reverses its drive sign. Finally,
the drive command is gated by steering alignment:
\begin{equation}
  \alpha_i = \max\!\left(0,\,
  \cos\,\mathrm{wrap}(\theta_i^{\mathrm{cmd}} - \theta_i)\right),
  \qquad
  \dot{q}_i^{\mathrm{cmd}} = \alpha_i\, \hat{s}_i / r_w,
\end{equation}
where $\hat{s}_i$ is the signed module speed. This removes scrubbing but
over-constrains the motion. During large steering swings the gated wheels
brake against the momentum of the body, and the shortest-path flip chatters
when $|e_i| \approx \pi/2$.

\paragraph{Smoothly Constrained Control (Final OSC).}
The final controller keeps the safeguards of the constrained version and
introduces two relaxations. The first is \emph{flip hysteresis}: each
module keeps a flip state $f_i$ that is updated as
\begin{equation}
  f_i \leftarrow
  \begin{cases}
    |e_i| > \pi/2 - h, & \text{if } f_i, \\
    |e_i| > \pi/2 + h, & \text{otherwise},
  \end{cases}
\end{equation}
with $h = 0.26$\,rad. This prevents chattering at the flip boundary. The
second is \emph{roll feed-forward}: instead of braking while the module
re-orients, each wheel rolls with the body. The measured body twist is
propagated through the IK to obtain the measured module velocity
$(\bar{v}_{x,i}, \bar{v}_{y,i})$, which is projected onto the current wheel
heading:
\begin{equation}
  u_i = \bar{v}_{x,i} \cos\theta_i + \bar{v}_{y,i} \sin\theta_i.
\end{equation}
The drive command blends the aligned command with this feed-forward term:
\begin{equation}
  \dot{q}_i^{\mathrm{cmd}} =
  \begin{cases}
    \hat{s}_i / r_w, & \text{if } s_i < s_{\min}, \\
    \left[\alpha_i\, \hat{s}_i + (1 - \alpha_i)\, u_i\right] / r_w,
    & \text{otherwise}.
  \end{cases}
\end{equation}
Commanded stops bypass the blend, so the robot always brakes fully when
asked to. In addition, the measured state is read at the start of each
control step, so alignment gating does not act on a stale estimate.

\begin{table}[h]
\centering
\caption{OSC parameters for the HMP.}
\label{tab:omni_params}
\renewcommand{\arraystretch}{1.2}
\footnotesize
\begin{tabular}{lc}
\toprule
\textbf{Parameter} & \textbf{Value} \\
\midrule
Max.\ linear speed $v_{\max}$              & 1.2 m/s \\
Max.\ yaw rate $\omega_{\max}$             & 2.97 rad/s \\
Linear acceleration limit                  & 0.3 m/s$^2$ \\
Angular acceleration limit                 & 0.3 rad/s$^2$ \\
Wheel radius $r_w$                         & 0.05 m \\
Min.\ module speed $s_{\min}$              & 0.03 m/s \\
Flip hysteresis $h$                        & 0.26 rad \\
\bottomrule
\end{tabular}
\end{table}

\subsection{Effect on Navigation Performance}
\label{app:omni_results}

With the OSC, the HMP COMPASS Specialist achieves robust navigation
performance across the warehouse environments, reaching success rates of
up to 74\% in the digital twin of the R\&D center. These results are obtained under
fully physics-based actuation, with all velocity and acceleration limits of
the platform enforced throughout training and evaluation. This demonstrates
that COMPASS can be effectively extended to holonomic, steer-actuated
embodiments without relying on idealized kinematic control.
\clearpage

%% file: tex/appendix/appendix_c_global_planner_integration.tex

\section{Global Planner Integration with COMPASS}
\label{app:global_planner}

This appendix describes how we extended the COMPASS stack with a global planner
so that it can reach far and occluded goals (scenario S5,
Section~\ref{sec:rw_scenarios}), and the modifications this required in the
COMPASS ROS~2 node.

\subsection{Motivation}
\label{app:global_planner_motivation}

COMPASS is a learned local policy. Its inputs are the camera image, a short route
window and the ego speed, and it has no map. It was trained to reach a goal along a
route that runs straight from the robot to the goal. The released ROS~2 node follows
the same logic: without an external route, its only route is a straight line from
the robot to the goal. When the goal is beyond the range of the policy, or hidden
behind racks, walls or other occluders, that straight line crosses the obstacle, and
the policy has no information about the way around it. Scenario S5
(Section~\ref{sec:rw_scenarios}) tests exactly this case. \review{In its digital-twin
replica, run without a global planner, none of the three COMPASS policies reaches
the goal (Appendix~\ref{app:twin_replica}), whatever its training.} The global
planner supplies the missing route, and the COMPASS ROS~2 node was modified to
accept it; since this integration lives in the ROS~2 node, it is used on the real
robot only.

\subsection{Architecture}
\label{app:global_planner_architecture}

\begin{figure}[H]
    \centering
    \begin{tikzpicture}[
        box/.style={draw, rounded corners, align=center, minimum width=3.4cm,
                    minimum height=0.9cm, font=\footnotesize},
        arr/.style={-latex, thick},
        lab/.style={font=\scriptsize, align=center, fill=white, inner sep=1pt}]
        \node[box] (bt) at (0,2.5) {Behavior tree};
        \node[box] (adp) at (0,0) {Navigation adaptor};
        \node[box] (gp) at (6,0) {Global planner\\(\mdlBaseline{})};
        \node[box] (cmp) at (6,-2.5) {COMPASS ROS~2 node};
        \node[box] (base) at (0,-2.5) {Velocity output};
        \draw[arr] (bt) -- node[lab, right=2pt] {goal pose} (adp);
        \draw[<->, thick] (adp) -- node[lab, above] {plan request;\\path} (gp);
        \draw[arr] (adp) to[bend left=15] node[lab, pos=0.55] {global route, goal pose, reset} (cmp);
        \draw[arr] (cmp) to[bend left=15] node[lab, pos=0.45] {velocity; ready, navigation type} (adp);
        \draw[arr] (adp) -- node[lab, right=2pt] {velocity} (base);
    \end{tikzpicture}
    \caption{Global planner integrated with the COMPASS ROS~2 node. The interfaces between the
    adaptor and the node are listed in Table~\ref{tab:global_planner_interfaces}.}
    \label{fig:global_planner_architecture}
\end{figure}
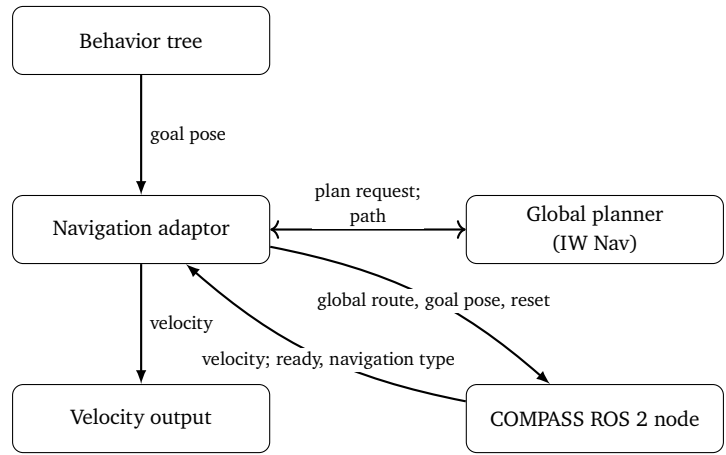

A goal pose is requested from the robot behavior tree (BT) through a navigation goal action.
The navigation adaptor, a ROS~2 component created for this integration, requests a plan from the
global planner, simplifies it, and passes the route and goal to COMPASS.
Figure~\ref{fig:global_planner_architecture} shows the data flow.

\paragraph{Navigation adaptor and global planner.} The adaptor implements the goal action that the BT
calls. For each goal it requests a plan from the global planner of the Idealworks navigation stack in the \texttt{map} frame. The plan is simplified
(Section~\ref{app:global_planner_simplification}) and published to COMPASS as the global route, followed by
the goal pose. COMPASS returns the velocity command, which the adaptor forwards to the base.
Table~\ref{tab:global_planner_config} lists the configuration of the integration.

\subsubsection{Path simplification}
\label{app:global_planner_simplification}

COMPASS was trained on straight routes from the robot to a goal. A curved polyline in the route window is
therefore out of distribution, and the policy does not hold such a path reliably, even in an empty field.
The adaptor reduces the plan to a few straight legs instead. The plan is first resampled to 1.0\,m spacing,
then simplified with the Ramer--Douglas--Peucker (RDP) algorithm at tolerance $\varepsilon = 0.5$\,m.

RDP starts with the segment joining the first and last pose and finds the interior pose farthest from it.
If that distance exceeds $\varepsilon$, the pose is kept and the procedure is applied to both halves. Otherwise
all interior poses are dropped and the segment becomes one leg. Every dropped pose lies within $\varepsilon$ of
the kept polyline, so the legs stay close to the original plan. A larger $\varepsilon$ gives fewer legs but
cuts further across the plan. Coincident waypoints are merged. The resulting waypoints are the ones the COMPASS
node tracks (Section~\ref{app:global_planner_node}).

\subsection{Integration Configuration}
\label{app:global_planner_config}

\begin{table}[H]
    \centering
    \caption{Configuration of the global planner integration.}
    \label{tab:global_planner_config}
    \renewcommand{\arraystretch}{1.2}
    \footnotesize
    \begin{tabular}{l c}
    \toprule
    \textbf{Parameter} & \textbf{Value} \\
    \midrule
    Global planner                          & A* lattice planner (\mdlBaseline{}) \\
    Planning frame                          & \texttt{map} \\
    Replanning rate (Hz)                    & 0 (one plan per goal) \\
    Route resampling spacing (m)            & 1.0 \\
    RDP tolerance $\varepsilon$ (m)         & 0.5 \\
    Intermediate goal switch radius (m)     & 1.5 \\
    Goal tolerance (m / $^\circ$)           & 0.5 / 20 \\
    \bottomrule
    \end{tabular}
\end{table}

\subsection{COMPASS ROS~2 Node Modifications}
\label{app:global_planner_node}

The released node builds its route only in mapless mode, as a straight line from the robot to the goal.
The modified node keeps that path and adds a sparse subgoal mode that follows the global route one waypoint
at a time. The global route is received on \texttt{/compass\_model/global\_route} with reliable, volatile QoS
of depth 1. Each message replaces the cached path and resets the subgoal index to 1.

\paragraph{Waypoint tracking.} The node keeps the latest waypoint as the current subgoal. The route window is
a straight line from the robot to that waypoint, sampled at 1.0\,m, which is the route shape the policy was
trained on. The goal heading is the orientation of the waypoint, which points to the next waypoint. The subgoal
advances when the robot is within 1.5\,m of the current waypoint, or when it has passed the leg end along the
leg direction. Advancing before the waypoint is reached keeps the robot from stopping at every waypoint, which
gives a smooth trajectory.

\paragraph{Interfaces.} Table~\ref{tab:global_planner_interfaces} lists the interfaces between the node and
the adaptor. The navigation type tells the adaptor whether the node drives in local mode (straight to the goal,
as in the released node) or in global mode (along the global route).

\begin{table}[H]
    \centering
    \caption{Interfaces between the modified COMPASS ROS~2 node and the adaptor. Changes with respect to the
    released node are marked as new or modified.}
    \label{tab:global_planner_interfaces}
    \renewcommand{\arraystretch}{1.2}
    \footnotesize
    \begin{tabularx}{\textwidth}{l l l l X}
    \toprule
    \textbf{Topic / service} & \textbf{Type} & \textbf{Direction} & \textbf{Change} & \textbf{Description} \\
    \midrule
    \texttt{/compass\_model/global\_route} & \texttt{nav\_msgs/Path} & In & New & Simplified global route (reliable, volatile, depth 1). \\
    \texttt{/compass/goal\_pose} & \texttt{geometry\_msgs/PoseStamped} & In & Modified & Goal pose; previously \texttt{/goal\_pose}. \\
    \texttt{/compass/reset} & \texttt{std\_msgs/Bool} & In & New & Clears goal, route and policy history. \\
    \texttt{/compass/cmd\_vel} & \texttt{geometry\_msgs/Twist} & Out & Modified & Velocity command to the adaptor; previously \texttt{/cmd\_vel}. \\
    \texttt{/compass/navigation\_type} & \texttt{std\_msgs/Bool} & Out & New & Latched; true for local navigation, false for global navigation. \\
    \texttt{/compass/ready} & \texttt{std\_srvs/Trigger} & Service & New & Ready once the model is loaded; the adaptor accepts goals only after. \\
    \bottomrule
    \end{tabularx}
\end{table}

\subsection{Logic Visualization}
\label{app:global_planner_logic}

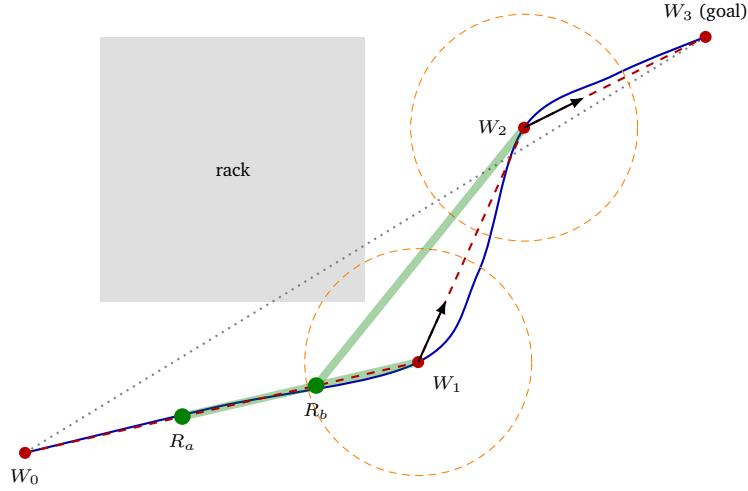
\begin{figure}[H]
    \centering
    \begin{tikzpicture}[scale=1.0,
        wp/.style={circle, fill=red!70!black, inner sep=1.6pt},
        rob/.style={circle, fill=green!50!black, inner sep=2.2pt},
        win/.style={line width=3pt, green!50!black, opacity=0.35},
        hdg/.style={-latex, thick},
        lab/.style={font=\scriptsize}]
        \fill[gray!25] (1.0,2.0) rectangle (4.5,5.5);
        \node[lab] at (2.75,3.75) {rack};
        \draw[gray, dotted, thick] (0,0) -- (9,5.5);
        \draw[blue!70!black, thick] plot[smooth, tension=0.6]
            coordinates {(0,0) (2.5,0.6) (5.2,1.2) (6.0,2.4) (6.6,4.3) (7.8,5.0) (9,5.5)};
        \draw[orange, densely dashed] (5.2,1.2) circle (1.5);
        \draw[orange, densely dashed] (6.6,4.3) circle (1.5);
        \draw[win] (2.08,0.48) -- (5.2,1.2);
        \draw[win] (3.85,0.89) -- (6.6,4.3);
        \draw[red!70!black, thick, dashed] (0,0) -- (5.2,1.2) -- (6.6,4.3) -- (9,5.5);
        \node[wp, label={[lab]below:$W_0$}] at (0,0) {};
        \node[wp, label={[lab]below right:$W_1$}] (w1) at (5.2,1.2) {};
        \node[wp, label={[lab]left:$W_2$}] (w2) at (6.6,4.3) {};
        \node[wp, label={[lab]above:$W_3$ (goal)}] at (9,5.5) {};
        \draw[hdg] (5.2,1.2) -- ++(0.37,0.82);
        \draw[hdg] (6.6,4.3) -- ++(0.80,0.40);
        \node[rob, label={[lab]below:$R_a$}] at (2.08,0.48) {};
        \node[rob, label={[lab]below:$R_b$}] at (3.85,0.89) {};
    \end{tikzpicture}
    \caption{Waypoint logic on an occluded goal. Gray dotted: straight route of the released node. Blue: global
    plan. Red dashed: RDP waypoints $W_i$ ($\varepsilon = 0.5$\,m). Orange: 1.5\,m switch radius. Green: route
    window given to COMPASS at $R_a$ (subgoal $W_1$) and at $R_b$ (inside the switch radius, subgoal $W_2$). Black
    arrows: goal heading, pointing to the next waypoint.}
    \label{fig:global_planner_logic}
\end{figure}

Figure~\ref{fig:global_planner_logic} illustrates the logic on a goal behind a rack. The straight route of the
released node crosses the rack. The global plan goes around it but is curved. RDP reduces it to three straight
legs. At $R_a$ the robot drives straight to $W_1$, with the heading pointing to $W_2$. At $R_b$ the robot has
entered the switch radius of $W_1$, so the subgoal advances and the route now runs straight to $W_2$ before
$W_1$ is reached.

\subsection{Other Modalities and Future Enhancements}
\label{app:global_planner_future}

\paragraph{Full-trajectory mode.} Before the sparse waypoint mode, the node was set up to follow the curved
global plan directly. The plan was resampled into 1.0\,m segments, and the global planner was invoked every
2\,s (0.5\,Hz) to compensate for the drift of the local planner. Following the curved plan naively gave
suboptimal paths. The policy did not hold the curved route, so the robot left the global plan and entered
restricted zones that the plan was routed around.

\paragraph{Restricted-zone-aware simplification.} The RDP-simplified route resolves this issue, and the results in Section~\ref{sec:real_world_benchmark}
confirm the effectiveness of this implementation. However, the simplification does not guarantee that the straight
legs respect restricted zones. Each leg may deviate from the plan by up to $\varepsilon$, so a leg can cut
through a restricted zone at a corner. A future implementation should check every leg against the restricted
zones and keep additional waypoints where a leg would cross one.

\paragraph{Replanning on drift.} When the local planner avoids an obstacle, the robot can drift significantly
from the originally planned path. In that case the remaining waypoints no longer give the best route from the
robot's current pose. A future implementation should invoke a replan when the robot deviates from the planned
path beyond a threshold, so that the route stays optimal and collision free.

\FloatBarrier

%% file: tex/appendix/appendix_d_training_datasets.tex

\section{Training Layouts and Datasets of the Real-World Benchmark Models}
\label{app:training_datasets}

\begin{figure}[H]
    \centering
    \begin{subfigure}[b]{0.42\linewidth}
        \centering
        \includegraphics[width=\linewidth]{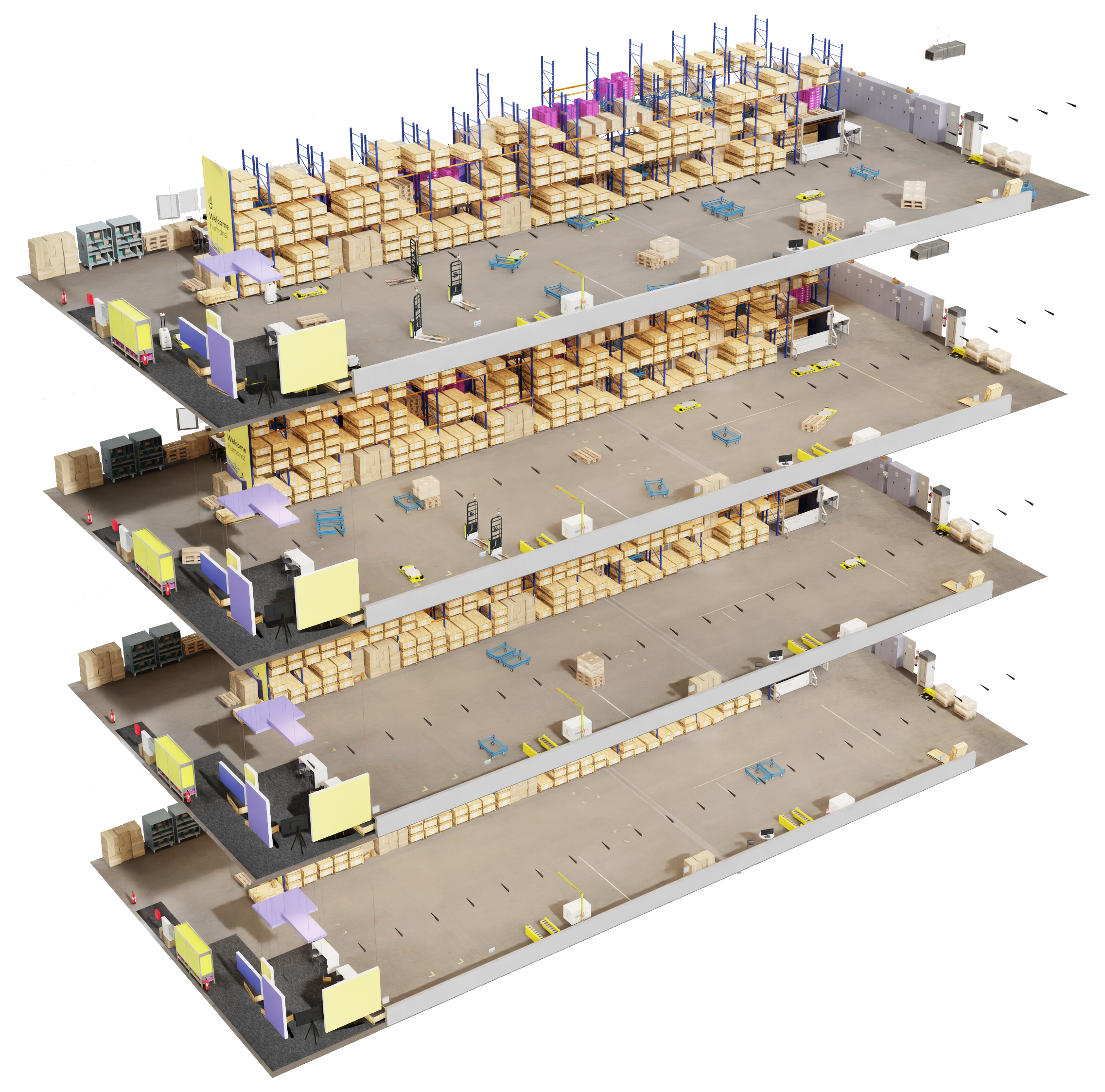}
        \caption{Rendered scenes}
        \label{fig:layouts_stack_render}
    \end{subfigure}\hspace{0.04\linewidth}%
    \begin{subfigure}[b]{0.42\linewidth}
        \centering
        \includegraphics[width=\linewidth]{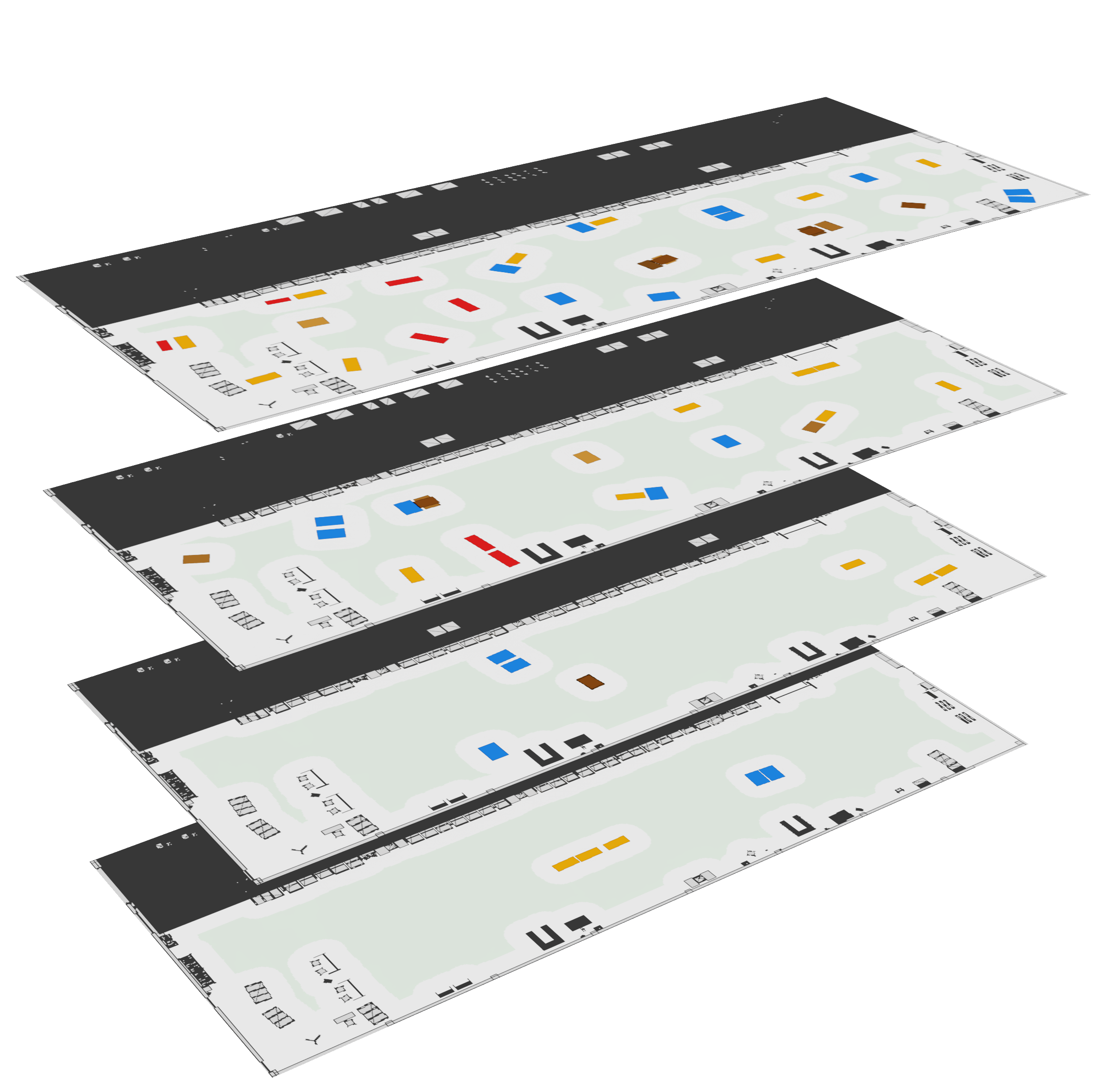}
        \caption{Occupancy maps}
        \label{fig:layouts_stack_occupancy}
    \end{subfigure}
    \caption{Four of the 16 layouts of the R\&D center, one per density tier, from top to bottom: layout 15 (dense), 06 (medium), 03 (sparse) and 09 (near-empty). In the occupancy maps, dollies are blue, iw.hubs orange, pallets brown and forklifts red, the light band around each obstacle is the robot clearance, and the remaining light-green region is the drivable area. }
    \label{fig:layouts_stack}
\end{figure}

This appendix describes the environments and data the models of the real-world
benchmark (Section~\ref{sec:rw_models}) are trained on;
Appendix~\ref{app:training_pipeline} describes how each training stage uses
them. Rather than the single arrangement of the original digital twin, we build
16 layouts of the R\&D center, each with its own set of floor objects at a
different density and its own lighting (Figure~\ref{fig:layouts_stack}). All
data and training use the iw.hub embodiment with its own camera setup, not the
NVIDIA Carter robot or its camera used for the released models. The same 16
layouts serve both training settings:
\begin{itemize}
    \item \textbf{X-Mobility data generation.} The random-exploration and
    path-following datasets used to fine-tune the world model (stage 1) and the IL policy
    (stage 2) are recorded in these layouts with
    MobilityGen~\citep{nvidia2025mobilitygen}
    (Appendix~\ref{app:datasets_dataset}). The hybrid versions, used for the
    IW.CE models, append a Cosmos-augmented sim-to-real copy of every episode
    (Appendix~\ref{app:datasets_sim2real}).
    \item \textbf{COMPASS training.} COMPASS (stage 3) learns online,
    directly in layout 04, one of the dense layouts. We implemented a custom
    start and goal sampling: each episode starts at a random collision-free
    pose, and the goal is sampled around it at a distance between 2 and 8\,m
    (Appendix~\ref{app:training_compass_scenario}).
\end{itemize}

\subsection{The Area}
\label{app:datasets_area}

All layouts share the building of the R\&D center (Table~\ref{tab:layouts_area}).
The north side of the building is a racking zone that the robot does not drive
in; the open zone south of the racks is where the floor objects are placed and
where the robot navigates.

The \emph{drivable area} is the free floor on which the robot center keeps a
clearance of 1.1\,m from every obstacle, taken as the largest connected region.
The 1.1\,m clearance is the planning radius of the iw.hub: the drive axle, which
is the robot origin, is 0.40\,m behind the front and 1.03\,m ahead of the rear,
so 1.1\,m keeps the rear clear when the robot turns in place. All drivable
areas in this appendix are measured in the same way, on a 5\,cm occupancy map of
the obstacles between 0.05 and 0.4\,m high.

\begin{table}[H]
    \centering
    \caption{Dimensions of the R\&D center in the digital twin.}
    \label{tab:layouts_area}
    \renewcommand{\arraystretch}{1.2}
    \footnotesize
    \begin{tabularx}{\textwidth}{l X}
    \toprule
    \textbf{Region} & \textbf{Size} \\
    \midrule
    Building interior (walls) & 48.8 $\times$ 20.4\,m = 996\,m$^2$ \\
    Racking zone (north, $y > {\sim}12$\,m) & Shelves, not driven \\
    Free floor of the empty layout & 647\,m$^2$ (553\,m$^2$ in the main connected part; the rest are offices and corners) \\
    \textbf{Drivable area of the empty layout} & \textbf{363\,m$^2$}, spanning $x$ 1.8--47.5\,m and $y$ 1.4--11.6\,m (the open zone south of the racks) \\
    \bottomrule
    \end{tabularx}
\end{table}

\subsection{Layout Generation}
\label{app:datasets_generation}

The layouts are generated procedurally from the digital twin in the following
steps:
\begin{enumerate}
    \item \textbf{Empty scene.} The original floor objects of the R\&D center are
    removed; walls, racks, fixtures, offices, signs and lights are kept. The
    occupancy map of this empty scene is the base on which every layout is
    planned.
    \item \textbf{Density tier.} Each layout is assigned a density factor equal
    to a random value in $[0.9, 1.1]$ times a tier scale. The tier scale cycles
    through dense (1.0), near-empty (0.1), medium (0.6) and sparse (0.3), so all
    densities appear across the 16 layouts and the two test layouts cover both
    ends (layout 14 sparse, layout 15 dense).
    \item \textbf{Object counts.} Base counts are drawn per layout and multiplied
    by the density factor: 8--12 pallets, 8--12 dollies, 8--12 iw.hubs and 2--4
    forklifts (stackers and other forklift models).
    \item \textbf{Placement in groups.} Objects are placed group by group at
    random free positions:
    \begin{itemize}
        \item a group holds 1, 2 or 3 objects side by side (with probabilities of
        45, 35 and 20\%), and 60\% of the time a neighbor is of another category
        (e.g.\ an iw.hub next to a pallet, or a dolly crosswise next to an
        iw.hub);
        \item 70\% of the groups are aligned with the building ($\pm 4^\circ$),
        like real staging areas, and the rest are placed at random angles;
        \item pallets are stacked 1--3 high (60, 30 and 10\%), and 30\% of the
        pallet stacks carry a box stack on top;
        \item groups keep at least 2.3\,m from each other, about twice the robot
        clearance, so the robot can pass between any two groups and obstacles
        form weaving paths rather than walls; they also keep at least 1.0\,m
        from walls and racks, so objects stand in the open floor;
        \item a group is rejected if it cuts more than 1\% of the drivable area
        off from the rest (no blocked aisles or closed corners), and the
        drivable area never drops below 30\% of that of the empty layout;
        \item groups are added until every category reaches its count, or until
        400 placements in a row fail (the layout is full).
    \end{itemize}
    \item \textbf{Lighting.} All lights are scaled by a brightness factor in
    $[0.5, 1.3]$, the color temperature is set between 3500 and 6500\,K (warm to
    cool white, $\pm 150$\,K per light), and 0--30\% of the 46 ceiling area
    lights are switched off, creating dark patches and uneven light.
    \item \textbf{Dolly blockers.} An invisible collider is placed over each
    dolly, so that planners treat a dolly as a solid obstacle and a robot that
    would fit under it collides instead of driving under it. The collider
    matches the dolly footprint in the data-generation scenes and is enlarged by
    0.3\,m in the COMPASS training scenes.
    \item \textbf{Labels.} Every placed object is a copy of an original scene
    object, with the same materials, colliders and semantic class, and each
    object is mapped to its class (iw.hub, dolly, pallet, forklift, \ldots) for
    the segmentation labels.
\end{enumerate}

\subsection{The 16 Layouts}
\label{app:datasets_layouts}

Table~\ref{tab:layouts_objects} lists the objects, the drivable area and the
lighting of each layout. Placed counts are the objects in the scene: each level of a pallet stack
counts as one pallet, and box stacks are counted separately. They can exceed the
requested counts because groups of up to three objects are added at once. The
drivable area is what remains of the 363\,m$^2$ of the empty layout once the
objects and the robot clearance around them are removed: the dense layouts keep
about half of it, the near-empty ones almost all of it.

\begin{table}[H]
    \centering
    \caption{Objects, drivable area and lighting of the 16 layouts. Layouts 14 and 15 are the test layouts. Light scale: below 0.7 dim, above 1.05 bright. Color temperature: below 4000\,K warm, above 6000\,K cool.}
    \label{tab:layouts_objects}
    \renewcommand{\arraystretch}{1.15}
    \footnotesize
    \begin{adjustbox}{max width=\textwidth}
    \begin{tabular}{l l l cc c ccccc c c}
    \toprule
    & & & \multicolumn{2}{c}{\textbf{Lighting}} & & \multicolumn{5}{c}{\textbf{Placed objects}} & & \\
    \cmidrule(lr){4-5} \cmidrule(lr){7-11}
    \textbf{Layout} & \textbf{Split} & \textbf{Tier}
      & Light scale & Color temp.\ (K)
      & \textbf{Density factor}
      & Pallets & Box stacks & Dollies & iw.hubs & Forklifts
      & \textbf{Total} & \textbf{Drivable (\% of empty)} \\
    \midrule
    00 & train & dense & 0.88 & 5706 & 1.027 & 10 & 2 & 9 & 9 & 3 & \textbf{33} & 55 \\
    01 & train & near-empty & 0.53 (dim) & 4504 & 0.108 & 1 & 0 & 1 & 0 & 0 & \textbf{2} & 97 \\
    02 & train & medium & 0.81 & 4760 & 0.550 & 4 & 2 & 5 & 4 & 1 & \textbf{16} & 71 \\
    03 & train & sparse & 0.64 (dim) & 4831 & 0.324 & 3 & 1 & 3 & 3 & 0 & \textbf{10} & 86 \\
    04 & train & dense & 0.65 (dim) & 4954 & 1.077 & 9 & 5 & 8 & 11 & 3 & \textbf{36} & 51 \\
    05 & train & near-empty & 0.84 & 6408 (cool) & 0.098 & 1 & 1 & 0 & 0 & 0 & \textbf{2} & 97 \\
    06 & train & medium & 1.07 (bright) & 5177 & 0.589 & 7 & 1 & 5 & 7 & 2 & \textbf{22} & 66 \\
    07 & train & sparse & 1.21 (bright) & 6319 (cool) & 0.272 & 2 & 1 & 3 & 2 & 1 & \textbf{9} & 86 \\
    08 & train & dense & 0.95 & 5167 & 1.052 & 14 & 2 & 11 & 8 & 2 & \textbf{37} & 51 \\
    09 & train & near-empty & 1.01 & 5379 & 0.101 & 0 & 0 & 2 & 3 & 0 & \textbf{5} & 91 \\
    10 & train & medium & 0.71 & 3712 (warm) & 0.600 & 8 & 2 & 5 & 6 & 2 & \textbf{23} & 69 \\
    11 & train & sparse & 1.06 (bright) & 6357 (cool) & 0.301 & 2 & 0 & 2 & 3 & 2 & \textbf{9} & 84 \\
    12 & train & dense & 1.00 & 3899 (warm) & 0.985 & 7 & 1 & 10 & 11 & 1 & \textbf{30} & 56 \\
    13 & train & near-empty & 0.52 (dim) & 5757 & 0.102 & 3 & 0 & 0 & 1 & 0 & \textbf{4} & 96 \\
    14 & \textbf{test} & sparse & 1.13 (bright) & 3806 (warm) & 0.303 & 4 & 1 & 2 & 3 & 2 & \textbf{12} & 80 \\
    15 & \textbf{test} & dense & 0.57 (dim) & 6418 (cool) & 1.024 & 12 & 3 & 9 & 9 & 5 & \textbf{38} & 52 \\
    \midrule
    \textbf{All} & 14 + 2 &  &  &  &  & \textbf{87} & \textbf{22} & \textbf{75} & \textbf{80} & \textbf{24} & \textbf{288} &  \\
    \bottomrule
    \end{tabular}
    \end{adjustbox}
\end{table}

\subsection{Roles of the Density Tiers}
\label{app:datasets_tiers}

Each density tier exposes the models to a different kind of navigation
(Table~\ref{tab:layouts_tiers}).

\begin{table}[H]
    \centering
    \caption{Density tiers of the layouts and their role in training.}
    \label{tab:layouts_tiers}
    \renewcommand{\arraystretch}{1.2}
    \footnotesize
    \begin{tabularx}{\textwidth}{l l c c X}
    \toprule
    \textbf{Tier} & \textbf{Layouts} & \textbf{Objects} & \textbf{Drivable} & \textbf{Role in training} \\
    \midrule
    Near-empty & 01, 05, 09, 13 & 2--5 & 91--97\% & Open-floor navigation: long, mostly straight drives to distant goals, turning and following the route; learning what free floor, walls and racks look like without clutter. The base case the policy must master before obstacle avoidance. \\
    Sparse & 03, 07, 11, 14 (test) & 9--12 & 80--86\% & Isolated obstacles: the robot meets one object or group at a time, so it learns to notice it in the camera image, steer around it and return to the route. Clean, close-up examples of each class (iw.hub, dolly, pallet, forklift). \\
    Medium & 02, 06, 10 & 16--23 & 66--71\% & Regular avoidance: several groups in view, choosing a side, passing between groups at least 2.3\,m apart, and more frequent re-planning. \\
    Dense & 00, 04, 08, 12, 15 (test) & 30--38 & 51--56\% & Cluttered staging areas: snaking through gaps between groups, close passes at the clearance limit, many classes in one frame, and objects appearing late from behind others. The hardest perception and avoidance cases. \\
    \bottomrule
    \end{tabularx}
\end{table}

The mix of tiers is chosen for the following reasons:
\begin{itemize}
    \item \textbf{A spread of difficulty} rather than one average density: the
    policy sees both easy open-floor driving, which makes up most of a real
    warehouse, and the hard cluttered cases, and the world model sees each
    class both alone and in clutter.
    \item \textbf{Test layouts at both ends.} Layouts 14 (sparse) and 15 (dense)
    are never trained on, so generalization is measured on new object
    arrangements and new lighting, in both an easy and a hard layout.
    \item \textbf{Mixed groups and random angles.} Objects stand next to other
    classes and at varied angles, so the model cannot learn that iw.hubs always
    come in rows or rely on fixed positions.
    \item \textbf{Gaps of at least 2.3\,m and no cut-off areas.} Every layout
    stays fully navigable, and obstacles force weaving paths rather than dead
    ends.
    \item \textbf{Lighting independent of density.} Dim, bright, warm, cool and
    partly switched-off lighting occurs in every tier, so appearance is not tied
    to clutter.
\end{itemize}

\subsection{The Dataset}
\label{app:datasets_dataset}

The X-Mobility datasets are recorded in the 16 layouts with MobilityGen and
consist of two sets (Table~\ref{tab:layouts_dataset}):
\begin{itemize}
    \item \textbf{Path following} (stage 2, IL policy): in each episode,
    MobilityGen's path-following expert drives the robot to one random goal
    after another until a time limit of 60\,s or a collision, with 15 episodes
    per layout recorded at 4\,Hz. The expert plans a shortest path to each goal
    on the occupancy map of the scene, with obstacles inflated by the 1.1\,m
    robot clearance, and follows it with a look-ahead heading controller at a
    constant 1.0\,m/s, turning in place for heading errors above $45^\circ$.
    This set is named \emph{nav2} after the X-Mobility dataset convention, but
    no ROS~2 Nav2 stack is used. Each goal segment
    is stored as a separate file, about 3.9 per episode.
    \item \textbf{Random exploration} (stage 1, world model): in each episode the
    robot drives randomly until a collision, and episodes are kept if they last
    at least 5\,s. They are recorded at 5\,Hz and stored in chunks of 50 frames.
\end{itemize}
The test split consists of layouts 14 and 15, which are never trained on; the
train and validation splits are a random split by episode over layouts 00--13.
All episodes are driven by the iw.hub embodiment, and all images are rendered
from the front camera of the real iw.hub, with its calibrated mounting pose and
intrinsics, at a resolution of $640 \times 360$; no data comes from the Carter
robot or the Carter camera of the NVIDIA datasets. In the
semantic labels, the floor under a dolly or within 0.3\,m of its footprint is
labeled as not drivable.
The random-exploration data of layouts 03 and 11 is smaller than that of the
other layouts because two of their recorders hung and were stopped; overall,
96\% of the planned random-exploration data was recorded.

\begin{table}[H]
    \centering
    \caption{Episodes, files and frames of the X-Mobility datasets recorded in the 16 layouts.}
    \label{tab:layouts_dataset}
    \renewcommand{\arraystretch}{1.15}
    \footnotesize
    \begin{tabular}{l l cccc}
    \toprule
    \textbf{Set} & \textbf{Split} & \textbf{Episodes} & \textbf{Files} & \textbf{Frames} & \textbf{Driving time} \\
    \midrule
    \multirow{4}{*}{\shortstack[l]{Path following\\(4\,Hz)}}
      & Train & 187 & 732 & 44,073 & \\
      & Val   & 23  & 95  & 5,501  & \\
      & Test  & 30  & 113 & 6,961  & \\
      & \textbf{Total} & \textbf{240} & \textbf{940} & \textbf{56,535} & \textbf{3.9\,h} \\
    \midrule
    \multirow{4}{*}{\shortstack[l]{Random\\(5\,Hz)}}
      & Train & 460 & 1,329 & 58,995 & \\
      & Val   & 57  & 159   & 7,092  & \\
      & Test  & 88  & 235   & 10,197 & \\
      & \textbf{Total} & \textbf{605} & \textbf{1,723} & \textbf{76,284} & \textbf{4.2\,h} \\
    \bottomrule
    \end{tabular}
\end{table}

\paragraph{Sim-to-real version.}
\label{app:datasets_sim2real}
For the IW.CE models we also create a sim-to-real version of the dataset.
Cosmos Transfer~2.5 augments every simulated episode into a photorealistic
episode, frame-aligned with the original so that its actions, states and labels
remain valid; the augmentation parameters and samples are given in
Appendix~\ref{app:cosmos_augmentation}. The hybrid dataset is built by
appending all the sim-to-real episodes to the simulated ones, which doubles the
size of the dataset and gives a 50\%/50\% split between simulated and
sim-to-real episodes. Each sim-to-real episode keeps the train, validation or test
split of its source episode, so layouts 14 and 15 remain unseen in both halves.

Table~\ref{tab:pipeline_datasets} summarizes the datasets used at each training
stage, including the hybrid versions used for the IW.CE models.

\begin{table}[H]
    \centering
    \caption{Fine-tuning datasets per training stage, all recorded in the 16 layouts of the R\&D center.}
    \label{tab:pipeline_datasets}
    \renewcommand{\arraystretch}{1.2}
    \footnotesize
    \begin{adjustbox}{max width=\textwidth}
    \begin{tabular}{l l l cccc}
    \toprule
    \textbf{Dataset} & \textbf{Used in} & \textbf{Content} & \textbf{Episodes} & \textbf{Frames} & \textbf{Cosmos-augmented (\%)} & \textbf{Split} \\
    \midrule
    R\&D random            & Stage 1, \wmSim{} & Random exploration   & 605  & 76,284 & 0    & Layouts 00--13 train/val, 14--15 test \\
    R\&D random (hybrid)   & Stage 1, \wmCE{}  & Random exploration   & 1,210 & 152,568 & 50 & Layouts 00--13 train/val, 14--15 test \\
    R\&D path following    & Stage 2, \ilSim{} & Path-following demonstrations  & 240  & 56,535 & 0    & Layouts 00--13 train/val, 14--15 test \\
    R\&D path following (hybrid) & Stage 2, \ilCE{}  & Path-following demonstrations  & 480   & 113,070 & 50 & Layouts 00--13 train/val, 14--15 test \\
    \bottomrule
    \end{tabular}
    \end{adjustbox}
\end{table}

\subsection{Semantic Labels for the iw.hub and Dolly Classes}
\label{app:datasets_labels}

The iw.hub and dolly classes are added without changing the architecture of
X-Mobility, by giving two of its seven existing semantic class slots a new
meaning. The NVIDIA model predicts background, navigable, forklift, pallet,
cone, sign and fence. Since the R\&D center contains no traffic cones or fences,
slot 4 (cone) is reassigned to the iw.hub and slot 6 (fence) to the dolly, both
in the label definitions of the model and in the training labels, where iw.hub
and dolly pixels are mapped to classes 4 and 6. The evaluations of
Appendix~\ref{app:training_world_model} therefore report iw.hub and dolly in
place of cone and fence.

Because the pretrained weights of these two slots had learned to recognize
cones and fences, only the parts of the network specific to them are
re-initialized. In each of the three semantic segmentation heads (full, half and
quarter resolution), the final $1 \times 1$ output convolution has one row of
weights and one bias per class; rows 4 and 6 are reset to the default PyTorch
initialization of a new convolution layer. Every other parameter keeps the
values of the NVIDIA checkpoint: the DINOv2 image encoder, the recurrent state
model of the world model, the shared decoder layers, the output rows of the
other five classes and the action policy. In total, 352 of the 358 weight
tensors of the model are identical to the NVIDIA checkpoint.

Starting from this checkpoint, the whole model is fine-tuned with a low learning
rate ($1 \times 10^{-5}$, one-cycle schedule) in two steps
(Appendix~\ref{app:training_pipeline}):
\begin{enumerate}
    \item the world model is fine-tuned on the random-exploration data with the
    new labels, so that the re-initialized outputs learn the iw.hub and dolly
    classes from features the encoder already provides;
    \item the NVIDIA action policy is merged back in, since world-model
    checkpoints contain no policy, and the world model and policy are fine-tuned
    together on the path-following data.
\end{enumerate}

\begin{reviewblock}
Starting from the NVIDIA weights, changing only the two reassigned outputs and
fine-tuning with a low learning rate preserves the existing classes and the
driving behavior on our data. Fine-tuning still specializes the model to the
R\&D center: on NVIDIA's own test set, the accuracy on the other classes, such
as pallet and forklift, drops noticeably. This is a limitation of the current
models, and mixing part of NVIDIA's data back into training is a way to
mitigate it.
\end{reviewblock}

\FloatBarrier

%% file: tex/appendix/appendix_e_training_pipeline.tex
%

\section{Training Pipeline of the Real-World Benchmark Models}
\label{app:training_pipeline}
\edef\appEfloatrestore{%
    \noexpand\renewcommand{\noexpand\floatpagefraction}{\floatpagefraction}%
    \noexpand\renewcommand{\noexpand\topfraction}{\topfraction}%
    \noexpand\renewcommand{\noexpand\textfraction}{\textfraction}}
\renewcommand{\floatpagefraction}{0.85}
\renewcommand{\topfraction}{0.85}
\renewcommand{\textfraction}{0.1}

This appendix documents how the \mdlSim{} and \mdlCE{} models evaluated in
Section~\ref{sec:real_world_benchmark} were obtained. A COMPASS model is the
result of three training stages, each producing the model the next stage builds
on:

\begin{enumerate}
    \item \textbf{World model (WM).} The X-Mobility world model is trained on
    random-exploration data.
    \item \textbf{Imitation-learning (IL) policy.} With the world model, the
    X-Mobility policy is trained by imitation learning on path-following expert
    demonstrations.
    \item \textbf{COMPASS.} The IL policy is frozen and used as the base policy
    for residual reinforcement learning on the iw.hub embodiment.
\end{enumerate}

For both variants, the world model and the IL policy start from the NVIDIA base
checkpoints and are fine-tuned exclusively on data generated in the digital twin
of the Idealworks R\&D center, while the COMPASS residual policy is always
trained from scratch in the digital twin on top of the resulting IL policy
(Appendix~\ref{app:training_compass}). Following the naming convention of
Section~\ref{sec:rw_models}, every model trained this way carries the prefix
IW., and IW.CE marks the Cosmos-enhanced ones (e.g.\ \wmSim{} and \wmCE{} for
the world models). The datasets used at stages 1 and 2 are described in
Appendix~\ref{app:training_datasets}. Each variant is a chain of its own models:
\begin{itemize}
    \item \mdlSim{}: \wmSim{} $\rightarrow$ \ilSim{} $\rightarrow$ residual RL;
    \item \mdlCE{}: \wmCE{} $\rightarrow$ \ilCE{} $\rightarrow$ residual RL.
\end{itemize}
\mdlCE{} differs from \mdlSim{} only in its fine-tuning data: its world model
and IL policy are fine-tuned on datasets that combine the original episodes with
Cosmos-augmented sim-to-real variants, and its residual policy is trained on top
of the Cosmos-enhanced IL policy. At the world model and IL stages we also evaluate the NVIDIA base model, without
fine-tuning, on the same evaluation data as a baseline.

\begin{table}[hbp]
    \centering
    \caption{Training stages of the real-world benchmark models.}
    \label{tab:pipeline_overview}
    \renewcommand{\arraystretch}{1.2}
    \footnotesize
    \begin{tabularx}{\textwidth}{l l X X}
    \toprule
    \textbf{Stage} & \textbf{Initialized from} & \textbf{\mdlSim{} chain} & \textbf{\mdlCE{} chain} \\
    \midrule
    1. World model  & NVIDIA base world model & \wmSim{}: R\&D random exploration & \wmCE{}: R\&D random exploration + Cosmos \\
    2. IL policy    & NVIDIA base X-Mobility & \ilSim{}: R\&D path-following demonstrations & \ilCE{}: R\&D path-following demonstrations + Cosmos \\
    3. COMPASS (RL) & Stage-2 IL policy (frozen) & \mdlSim{}: residual RL on \ilSim{} & \mdlCE{}: residual RL on \ilCE{} \\
    \bottomrule
    \end{tabularx}
\end{table}

\subsection{Stage 1: World Model}
\label{app:training_world_model}

The world model is fine-tuned by continuing the training of the released NVIDIA
world model on the random-exploration data of the digital twin
(Table~\ref{tab:pipeline_wm_params}). The training recipe is the released one;
the only change is the setup of the iw.hub and dolly labels described below. The
checkpoint used in the evaluations is the one with the lowest validation loss.

The released X-Mobility world model has no iw.hub or dolly class, two objects
common in the R\&D center; the evaluations below therefore report them
in place of two reassigned classes of the released model, as described in
Appendix~\ref{app:datasets_labels}.

\begin{table}[htbp]
    \centering
    \caption{World model fine-tuning hyperparameters.}
    \label{tab:pipeline_wm_params}
    \renewcommand{\arraystretch}{1.2}
    \footnotesize
    \begin{adjustbox}{max width=\textwidth}
    \begin{tabular}{l cc}
    \toprule
    \textbf{Parameter} & \textbf{\wmSim{}} & \textbf{\wmCE{}} \\
    \midrule
    Initial checkpoint   & \texttt{x\_mobility-nav2-semantic\_action\_path} & \texttt{x\_mobility-nav2-semantic\_action\_path} \\
    Training epochs      & 24   & 17 \\
    Batch size           & 8 per GPU & 8 per GPU \\
    Optimizer            & AdamW & AdamW \\
    Learning rate (schedule) & peak $1 \times 10^{-5}$ & peak $1 \times 10^{-5}$ \\
    Sequence length      & 4 frames (0.8 s at 5Hz) & 4 frames (0.8 s at 5Hz) \\
    Hardware (GPUs)      & 6x RTX PRO 6000 Blackwell & 6x RTX PRO 6000 Blackwell \\
    Training time        & 47 minutes & 45 minutes \\
    \bottomrule
    \end{tabular}
    \end{adjustbox}
\end{table}

\begin{table}[htbp]
    \centering
    \caption{World model evaluation on the validation split of the R\&D datasets. The iw.hub and dolly classes replace the fence and cone classes of the released model (Appendix~\ref{app:datasets_labels}); the NVIDIA base model was not trained on them.}
    \label{tab:pipeline_wm_eval}
    \renewcommand{\arraystretch}{1.15}
    \footnotesize
    \begin{adjustbox}{max width=\textwidth}
    \begin{tabular}{l ccc}
    \toprule
    \textbf{Metric} & \textbf{NVIDIA base} & \textbf{\wmSim{}} & \textbf{\wmCE{}} \\
    \midrule
    \multicolumn{4}{l}{\textit{World-model losses $\downarrow$}} \\
    \quad \textbf{Summed Loss} & 15.3256 & 1.1166 & 1.1972 \\
    \quad KL loss & 0.6024 & 0.1803 & 0.1653 \\
    \quad Semantic loss, scale 1 & 8.6618 & 0.5408 & 0.5948 \\
    \quad Semantic loss, scale 2 & 4.1827 & 0.2657 & 0.2941 \\
    \quad Semantic loss, scale 4 & 1.8787 & 0.1298 & 0.1430 \\
    \midrule
    \multicolumn{4}{l}{\textit{Semantic IoU (\%) $\uparrow$}} \\
    \quad Background & 69.82 & 94.09 & 94.21 \\
    \quad Navigable & 76.14 & 95.32 & 95.48 \\
    \quad Forklift & 1.05 & 13.36 & 12.56 \\
    \quad Pallet & 5.75 & 35.85 & 35.31 \\
    \quad Sign & 0.25 & 13.35 & 14.41 \\
    \quad \textbf{iw.hub} & 0.01 & 26.01 & 24.72 \\
    \quad \textbf{Dolly} & 0.14 & 21.75 & 22.21 \\
    \midrule
    \multicolumn{4}{l}{\textit{Semantic F1 (\%) $\uparrow$}} \\
    \quad Background & 81.48 & 96.86 & 96.92 \\
    \quad Navigable & 85.50 & 97.40 & 97.49 \\
    \quad Forklift & 1.89 & 17.95 & 16.85 \\
    \quad Pallet & 9.60 & 46.57 & 45.66 \\
    \quad Sign & 0.49 & 19.72 & 21.17 \\
    \quad \textbf{iw.hub} & 0.02 & 33.38 & 31.61 \\
    \quad \textbf{Dolly} & 0.28 & 29.93 & 30.43 \\
    \bottomrule
    \end{tabular}
    \end{adjustbox}
\end{table}

\paragraph{Imagination evaluation.}
The world model predicts future scenes from its latent state, without new
observations, and the predicted semantic segmentation is compared with the
ground truth at increasing time horizons (Table~\ref{tab:pipeline_wm_imagination}).
The world model first observes five frames and then imagines the following
frames from the recorded actions alone; the table reports the semantic IoU at
the end of each horizon.

\begin{table}[htbp]
    \centering
    \caption{World model imagination evaluation: semantic endpoint IoU (\%) of predicted future scenes on the validation split of the random-exploration dataset (57 episodes, 7,092 frames), after five observed frames and using the recorded actions, by prediction horizon (frames at 5\,Hz).}
    \label{tab:pipeline_wm_imagination}
    \renewcommand{\arraystretch}{1.15}
    \footnotesize
    \begin{adjustbox}{max width=\textwidth}
    \begin{tabular}{c ccc ccc ccc ccc}
    \toprule
    & \multicolumn{3}{c}{\textbf{Background IoU (\%)} $\uparrow$}
    & \multicolumn{3}{c}{\textbf{Navigable IoU (\%)} $\uparrow$}
    & \multicolumn{3}{c}{\textbf{iw.hub IoU (\%)} $\uparrow$}
    & \multicolumn{3}{c}{\textbf{Dolly IoU (\%)} $\uparrow$} \\
    \cmidrule(lr){2-4} \cmidrule(lr){5-7} \cmidrule(lr){8-10} \cmidrule(lr){11-13}
    \textbf{Horizon (s)}
      & NVIDIA base & \wmSim{} & \wmCE{}
      & NVIDIA base & \wmSim{} & \wmCE{}
      & NVIDIA base & \wmSim{} & \wmCE{}
      & NVIDIA base & \wmSim{} & \wmCE{} \\
    \midrule
    0.2 & 70.39 & 93.22 & 93.44 & 78.34 & 94.84 & 95.01 & 0.00 & 14.46 & 13.41 & 0.06 & 13.16 & 13.76 \\
    0.4 & 70.69 & 92.06 & 92.31 & 77.82 & 93.61 & 93.74 & 0.00 & 10.71 & 10.52 & 0.04 & 10.72 & 11.08 \\
    0.6 & 70.73 & 91.13 & 91.48 & 76.80 & 92.51 & 92.61 & 0.00 & 8.35 & 8.68 & 0.03 & 8.58 & 9.11 \\
    0.8 & 70.72 & 90.35 & 90.76 & 75.56 & 91.44 & 91.50 & 0.00 & 6.60 & 7.33 & 0.04 & 7.21 & 7.60 \\
    1.0 & 70.63 & 89.78 & 90.20 & 74.29 & 90.46 & 90.57 & 0.00 & 5.37 & 6.39 & 0.02 & 6.20 & 6.49 \\
    2.0 & 71.10 & 87.68 & 88.25 & 69.12 & 85.44 & 86.15 & 0.00 & 3.02 & 3.52 & 0.02 & 3.52 & 3.52 \\
    \bottomrule
    \end{tabular}
    \end{adjustbox}
\end{table}

\FloatBarrier
\subsection{Stage 2: Imitation-Learning Policy}
\label{app:training_il_policy}

The IL policy is fine-tuned from the NVIDIA checkpoint on the path-following
data, on top of its own stage-1 world model (\wmSim{} for \ilSim{}, \wmCE{}
for \ilCE{}). The world model is not frozen: it continues to train together with
the policy (Table~\ref{tab:pipeline_il_params}).

\begin{table}[htbp]
    \centering
    \caption{IL policy fine-tuning hyperparameters.}
    \label{tab:pipeline_il_params}
    \renewcommand{\arraystretch}{1.2}
    \footnotesize
    \begin{tabular}{l cc}
    \toprule
    \textbf{Parameter} & \textbf{\ilSim{}} & \textbf{\ilCE{}} \\
    \midrule
    Initial checkpoint   & \texttt{x\_mobility-nav2-semantic\_action\_path} & \texttt{x\_mobility-nav2-semantic\_action\_path} \\
    World model (stage 1) & \wmSim{} & \wmCE{} \\
    Training epochs & 31 & 29 \\
    Batch size           & 4 per GPU & 4 per GPU \\
    Optimizer            & AdamW & AdamW \\
    Learning rate (schedule) & peak $1 \times 10^{-5}$ & peak $1 \times 10^{-5}$ \\
    Hardware (GPUs)      & 6x RTX PRO 6000 Blackwell & 6x RTX PRO 6000 Blackwell \\
    Training time        & 44 minutes & 44 minutes \\
    \bottomrule
    \end{tabular}
\end{table}

\paragraph{Open-loop evaluation.}
The policy predicts actions and paths for the recorded episodes of the validation
split of the path-following dataset (23 episodes, 5,501 frames;
Appendix~\ref{app:datasets_dataset}), and its predictions are compared with the
expert's
(Table~\ref{tab:pipeline_il_openloop}). We report the mean absolute errors of
the linear speed, the angular speed and the predicted path (linear, angular and
path MAE), as defined in X-Mobility~\citep{liu2024xmobility}, together with the
weighted action and path losses of the X-Mobility training objective.

\begin{table}[htbp]
    \centering
    \caption{IL policy open-loop evaluation on the validation split of the R\&D path-following dataset.}
    \label{tab:pipeline_il_openloop}
    \renewcommand{\arraystretch}{1.15}
    \footnotesize
    \begin{tabular}{l ccc}
    \toprule
    \textbf{Metric} $\downarrow$ & \textbf{NVIDIA base X-Mobility} & \textbf{\ilSim{}} & \textbf{\ilCE{}} \\
    \midrule
    Linear MAE & 0.5131 & 0.0089 & 0.0080 \\
    Angular MAE & 0.1869 & 0.0451 & 0.0432 \\
    Path MAE & 0.1518 & 0.0805 & 0.0806 \\
    Weighted action loss & 7.0129 & 0.5458 & 0.5176 \\
    Weighted path loss & 3.3924 & 1.6578 & 1.6572 \\
    \bottomrule
    \end{tabular}
\end{table}

\paragraph{Closed-loop evaluation.}
The policy drives the iw.hub in the digital twin of the R\&D center
(Table~\ref{tab:pipeline_il_eval}). The policies are evaluated closed-loop in
layouts 14 and 15, the test layouts of the dataset
(Appendix~\ref{app:datasets_dataset}), which are never seen in training and
cover a sparse and a dense layout. An episode counts as a success when the robot
reaches the goal within 0.5\,m with a linear velocity below 1.0\,m/s and an
angular velocity below 1.0\,rad/s. These velocity thresholds are deliberately
loose: the path-following expert drives at a constant 1.0\,m/s and never brakes
at the goal (Appendix~\ref{app:datasets_dataset}), so the IL policy, which
imitates it, also reaches the goal at full speed. A stopping criterion such as
the one used for COMPASS (Appendix~\ref{app:training_compass_scenario}) would
count these arrivals as failures; stopping at the goal is learned only by the
COMPASS residual policy.
Each policy is evaluated over 300 rollouts per layout, with start and goal
poses sampled uniformly across the drivable area of the layout.

\begin{reviewblock}
Both fine-tuned policies far outperform the NVIDIA base policy, which reaches
the goal in only 2.3\% of the rollouts, against 69.3\% for \ilSim{} and
66.7\% for \ilCE{} over both layouts (Table~\ref{tab:pipeline_il_eval}).
\ilSim{} and \ilCE{} perform almost the same on every metric, and this is
expected: the test layouts are rendered in simulation, where only the simulated
half of the \ilCE{} training data matches the test images. That the two stay
this close is a good sign: adding the Cosmos sim-to-real episodes does not cost
any performance in simulation, and their benefit is expected on the real robot
(Section~\ref{sec:rw_results}).
\end{reviewblock}

\begin{table}[htbp]
    \centering
    \caption{IL policy closed-loop evaluation in the digital twin of the R\&D center, on the test layouts 14 and 15, over 300 rollouts per layout and model. Both layouts: SR, PR, CC, J and OC are averaged over all 600 rollouts; TTG and PLR over the successful rollouts, and WTT is TTG divided by the success rate.}
    \label{tab:pipeline_il_eval}
    \renewcommand{\arraystretch}{1.2}
    \footnotesize
    \begin{adjustbox}{max width=\textwidth}
    \begin{tabular}{ll cc ccc c cc}
    \toprule
    & & \multicolumn{2}{c}{\textbf{Task}} & \multicolumn{3}{c}{\textbf{Efficiency}}
    & \textbf{Safety} & \multicolumn{2}{c}{\textbf{Smoothness}} \\
    \cmidrule(lr){3-4} \cmidrule(lr){5-7} \cmidrule(lr){8-8} \cmidrule(lr){9-10}
    \textbf{Layout} & \textbf{Model}
      & SR (\%) $\uparrow$ & PR (\%) $\uparrow$
      & WTT (s) $\downarrow$ & TTG (s) $\downarrow$ & PLR $\downarrow$
      & CC $\downarrow$
      & J (m/s$^3$) $\downarrow$ & OC $\downarrow$ \\
    \midrule
    \multirow{3}{*}{\shortstack[l]{Layout 14\\(sparse)}}
       & NVIDIA base X-Mobility & 2.00 & 40.32 & 1225.00 & 24.50 & 1.194 & 0.180 & 1.28 & 7.72 \\
       & \ilSim{} & 80.33 & 85.79 & 23.25 & 18.68 & 1.178 & 0.203 & 1.41 & 0.75 \\
       & \ilCE{} & 78.33 & 83.93 & 24.26 & 19.00 & 1.183 & 0.223 & 1.40 & 1.06 \\
    \midrule
    \multirow{3}{*}{\shortstack[l]{Layout 15\\(dense)}}
       & NVIDIA base X-Mobility & 2.67 & 30.59 & 671.25 & 17.90 & 1.048 & 0.350 & 1.50 & 9.80 \\
       & \ilSim{} & 58.33 & 76.64 & 29.13 & 16.99 & 1.121 & 0.417 & 1.40 & 0.99 \\
       & \ilCE{} & 55.00 & 75.25 & 31.00 & 17.05 & 1.144 & 0.453 & 1.36 & 1.01 \\
    \midrule
    \multirow{3}{*}{\shortstack[l]{Both\\layouts}}
        & NVIDIA base X-Mobility & 2.33 & 35.45 & 888.37 & 20.73 & 1.111 & 0.265 & 1.39 & 8.76 \\
        & \ilSim{} & 69.33 & 81.22 & 25.91 & 17.97 & 1.154 & 0.310 & 1.41 & 0.87 \\
        & \ilCE{} & 66.67 & 79.59 & 27.30 & 18.20 & 1.167 & 0.338 & 1.38 & 1.04 \\
    \bottomrule
    \end{tabular}
    \end{adjustbox}
\end{table}

\FloatBarrier
\subsection{Stage 3: COMPASS Residual Reinforcement Learning}
\label{app:training_compass}

The residual policy is trained from scratch for each variant, on top of its
frozen IL policy (\ilSim{} for \mdlSim{}, \ilCE{} for \mdlCE{}). A COMPASS
residual learns corrections to the actions of the specific base policy it was
trained with, so a residual trained on one X-Mobility base, such as the \mdlNV{}
residual on the NVIDIA base policy, does not transfer to a fine-tuned base, and
fine-tuning it on top of \ilSim{} or \ilCE{} is not a viable starting point.

Compared with the released COMPASS training setup, we made the following
additions:
\begin{itemize}
    \item iw.hub embodiment asset with its sensor configuration, including its
    real front camera.
    \item Differential drive controller (Appendix~\ref{app:controller_study}).
    \item Layout 04 of the digital twin of the R\&D center
    (Appendix~\ref{app:training_datasets}) as training environment.
    \item Custom start and goal sampling (below).
\end{itemize}

\paragraph{Training scenario.}
\label{app:training_compass_scenario}
Both residual policies are trained online in a single layout, layout 04 of
Appendix~\ref{app:training_datasets}: a dense layout with 36 objects, in which
about 28\% of the training routes are blocked by an obstacle across the width of
the robot. We implemented a custom logic that sets the start and goal of each
training episode:
\begin{enumerate}
    \item \textbf{Start pose.} At each reset, the robot is placed at a uniform
    random pose inside the sampling box of the scene ($x$ 3--46\,m, $y$
    3.5--10.5\,m) with any yaw. The pose is accepted when the iw.hub body, grown
    by 0.25\,m, is free on the occupancy map, with up to 10 tries.
    \item \textbf{Goal pose.} The goal is then sampled around the start position
    in polar coordinates: a distance $d \sim \mathcal{U}(2\,\text{m}, 8\,\text{m})$
    and a direction $\theta \sim \mathcal{U}(-\pi, \pi)$, giving the goal
    position $\mathbf{p}_{\text{start}} + d\,(\cos\theta, \sin\theta)$, with a
    goal heading drawn uniformly from its range.
\end{enumerate}
Because the distance is uniform, short and long goals are equally frequent;
per unit of area, however, goals are denser near the robot than they would be
with uniform-area sampling. The base policy is given a straight-line route to
the goal. Each training iteration runs 450 steps (45\,s), after which every
robot is reset.

\paragraph{Rewards and success criterion.}
The rewards are listed in Table~\ref{tab:pipeline_compass_rewards}. A goal is
reached when the robot is within 0.5\,m of it, below 0.1\,m/s and 0.1\,rad/s,
with its heading within 0.1\,rad of the goal heading. PPO uses an exploration
noise of 0.1 on all residual outputs, an entropy bonus of 0.01 and an adaptive
learning rate. The X-Mobility base policy is frozen; only the residual policy
and the critic are trained.

\begin{table}[tbp]
    \centering
    \caption{Reward terms of the COMPASS residual RL training.}
    \label{tab:pipeline_compass_rewards}
    \renewcommand{\arraystretch}{1.2}
    \footnotesize
    \begin{tabular}{l c}
    \toprule
    \textbf{Term} & \textbf{Weight} \\
    \midrule
    Distance to the goal, per step & $-1$ \\
    Goal-heading alignment         & $2$ \\
    Reaching the goal              & $+2000$ \\
    Collision                      & $-1000$ \\
    \bottomrule
    \end{tabular}
\end{table}

\begin{table}[tbp]
    \centering
    \caption{COMPASS residual RL hyperparameters. Both runs use 32 robots per update.}
    \label{tab:pipeline_compass_params}
    \renewcommand{\arraystretch}{1.2}
    \footnotesize
    \begin{adjustbox}{max width=\textwidth}
    \begin{tabular}{l cc}
    \toprule
    \textbf{Parameter} & \textbf{\mdlSim{}} & \textbf{\mdlCE{}} \\
    \midrule
    Base policy (stage 2, frozen) & \ilSim{} & \ilCE{} \\
    Training environment  & Layout 04 & Layout 04 \\
    Training iterations & 440 & 440 \\
    Parallel environments (robots) & 32 & 32 \\
    GPUs $\times$ robots per GPU & 8 $\times$ 4 & 8 $\times$ 4 \\
    Rollout horizon (steps) & 450 (45\,s) & 450 (45\,s) \\
    Learning rate         & adaptive (KL target 0.01, init.\ $10^{-4}$) & adaptive (KL target 0.01, init.\ $10^{-4}$) \\
    Exploration noise (initial $\sigma$) & 0.1 & 0.1 \\
    Entropy coefficient   & 0.01 & 0.01 \\
    Discount $\gamma$ / GAE $\lambda$ & 0.99 / 0.95 & 0.99 / 0.95 \\
    PPO clip range        & 0.2 & 0.2 \\
    Goal reached          & 0.5\,m, 0.1\,m/s, 0.1\,rad/s, 0.1\,rad & 0.5\,m, 0.1\,m/s, 0.1\,rad/s, 0.1\,rad \\
    \bottomrule
    \end{tabular}
    \end{adjustbox}
\end{table}

\begin{figure}[tbp]
    \centering
    \includegraphics[width=\linewidth]{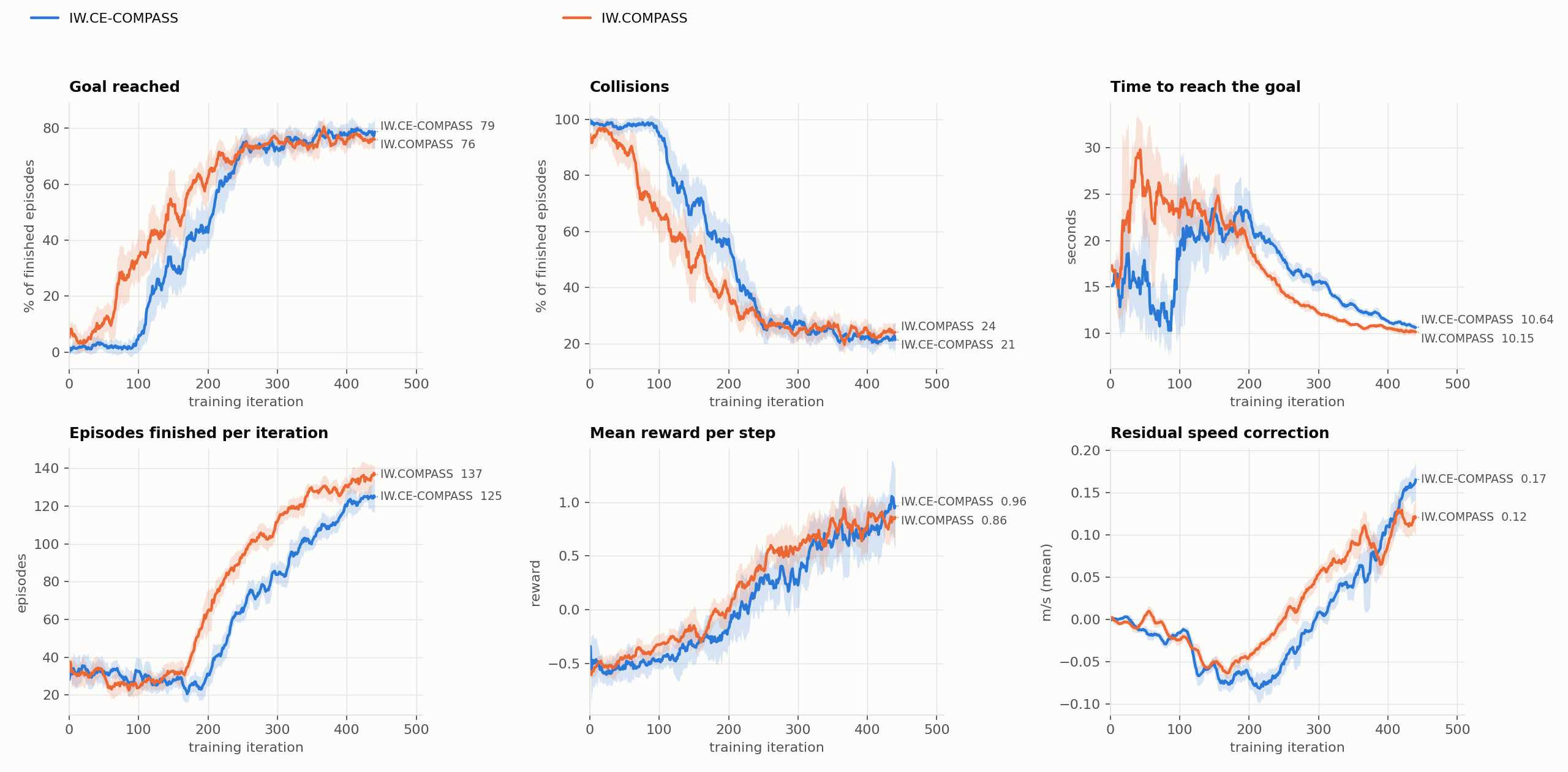}
    \caption{COMPASS residual RL training of \mdlCE{} (blue) and \mdlSim{} (orange) on layout 04, up to iteration 440. Training metrics, measured with exploration noise. Lines are a 10-iteration rolling mean; the shaded bands are $\pm 1$ standard deviation over the same 10 iterations.}
    \label{fig:pipeline_compass_curves}
\end{figure}

\paragraph{Training curves.}
\label{app:training_compass_curves}
The two runs use the same configuration, and the only difference between them
is the X-Mobility base policy the residual is trained on top of. Since both use
32 robots per update, their curves are directly comparable
(Figure~\ref{fig:pipeline_compass_curves}); both are shown up to iteration 440.
The figure reports the share of finished episodes that reach the goal and that
end in a collision, the mean duration of successful episodes, the number of
episodes finished per 45\,s iteration over all 32 robots (higher means faster
episodes), the mean training reward per step, and the mean value the residual
adds to the forward-speed command of the base (positive means faster than the
base). A finished episode ends in a goal or a collision; episodes cut off at the
end of an iteration count as neither.

Table~\ref{tab:pipeline_compass_checkpoints} compares the two runs at the same
checkpoints; each value is the mean of the 10 iterations before the checkpoint.

\begin{table}[tbp]
    \centering
    \caption{Training metrics of the two COMPASS runs at the same checkpoints: share of finished episodes reaching the goal and ending in a collision, and mean time to reach the goal. Mean of the 10 iterations before each checkpoint.}
    \label{tab:pipeline_compass_checkpoints}
    \renewcommand{\arraystretch}{1.2}
    \footnotesize
    \begin{tabular}{c ccc ccc}
    \toprule
    & \multicolumn{3}{c}{\textbf{\mdlCE{}}} & \multicolumn{3}{c}{\textbf{\mdlSim{}}} \\
    \cmidrule(lr){2-4} \cmidrule(lr){5-7}
    \textbf{Iteration} & Goal (\%) $\uparrow$ & Collision (\%) $\downarrow$ & Time (s) $\downarrow$
      & Goal (\%) $\uparrow$ & Collision (\%) $\downarrow$ & Time (s) $\downarrow$ \\
    \midrule
    100 & 6  & 94 & 18.4 & 34 & 66 & 22.7 \\
    200 & 43 & 57 & 22.4 & 62 & 38 & 19.2 \\
    300 & 73 & 27 & 15.6 & 76 & 24 & 12.2 \\
    400 & 78 & 22 & 11.3 & 75 & 25 & 10.5 \\
    440 & 79 & 21 & 10.6 & 76 & 24 & 10.2 \\
    \bottomrule
    \end{tabular}
\end{table}

\begin{reviewblock}
The two trainings differ in four ways:
\begin{enumerate}
    \item \textbf{\mdlCE{} starts much slower.} For the first ${\sim}100$
    iterations its residual reaches almost no goals (6\% against 34\% at
    iteration 100): the \ilCE{} base needs more correction on the iw.hub at
    first. It catches up between iterations 150 and 250.
    \item \textbf{\mdlCE{} ends slightly ahead on success and safety.} At
    iteration 440 it reaches the goal in 79\% of the finished episodes with 21\%
    collisions, against 76\% and 24\% for \mdlSim{}. The difference of 2--3
    points is about the size of the iteration-to-iteration noise, so it needs to
    be confirmed in an evaluation.
    \item \textbf{\mdlSim{} is faster.} At iteration 440 it reaches the goal in
    10.2\,s against 10.6\,s and finishes more episodes per iteration (137 against
    125).
    \item \textbf{Both have plateaued in success} since roughly iteration
    300--400. Late in training both residuals add more and more forward speed,
    from about 0 to $+0.15$\,m/s: the gains go into arriving sooner rather than
    into fewer collisions.
\end{enumerate}

These are training metrics, measured with exploration noise; the deployed,
noise-free policy usually scores somewhat higher. The fair comparison is a
noise-free evaluation on the same start and goal pairs, ideally on the test
layouts 14 and 15, which are not used in training. Moreover, both residuals are
trained and measured on a single layout, so these curves do not show how well
either generalizes to other scenes.
\end{reviewblock}

\FloatBarrier
\appEfloatrestore

%% file: tex/appendix/appendix_f_cosmos_samples.tex

\section{Sim-to-Real Augmentation Samples}
\label{app:cosmos_augmentation}

This appendix shows samples of the sim-to-real augmentation used for the
IW.CE models (Section~\ref{sec:rw_models}, Appendix~\ref{app:training_pipeline}).
The Cosmos-enhanced datasets are generated with Cosmos Transfer~2.5, part of the
NVIDIA Cosmos world foundation models~\citep{nvidia2025cosmos}, which
re-renders a simulated episode as a photorealistic video while preserving its
scene geometry and motion. Each episode's camera frames, recorded at 4 or 5\,Hz, are
encoded into a 10\,fps video with one video frame per recorded frame,
 passed through Cosmos Transfer~2.5,
and the generated frames are written back into the episode, frame-aligned with
the original, so the actions, states and labels of the episode remain valid.

\begin{figure}[H]
    \centering
    \includegraphics[width=\linewidth]{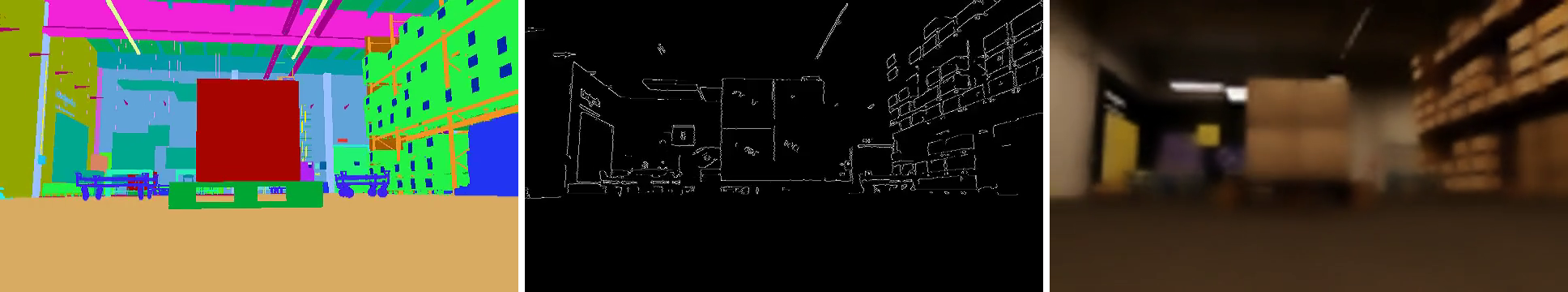}
    \caption{Control inputs of Cosmos Transfer 2.5 for one frame of a simulated episode: semantic segmentation rendered by the simulator (left, weight 1.0), edge map (center, weight 0.4) and blurred frame (right, weight 0.1), both computed from the input video.}
    \label{fig:cosmos_controls}
\end{figure}

All episodes are generated with the same configuration
(Table~\ref{tab:cosmos_settings}). The generation is conditioned on three
controls (Figure~\ref{fig:cosmos_controls}). The semantic segmentation rendered by the simulator, with weight 1.0,
fixes the layout and the objects of the scene. An edge map and a blurred copy of
the simulated frames, both computed from the input video, with weights 0.4 and
0.1, keep the geometry and the coarse appearance close to the original while
leaving Cosmos free to re-render textures and lighting. No depth control is
used. Each episode is output at its input resolution and
frame rate, $640 \times 360$ at 10\,fps, so that every generated frame replaces
exactly one simulated frame.

\begin{table}[H]
    \centering
    \caption{Cosmos Transfer 2.5 augmentation settings.}
    \label{tab:cosmos_settings}
    \renewcommand{\arraystretch}{1.2}
    \footnotesize
    \begin{tabular}{l c}
    \toprule
    \textbf{Setting} & \textbf{Value} \\
    \midrule
    Model                              & Cosmos Transfer 2.5, 2B multi-control model \\
    Control inputs (weight)            & Segmentation (1.0), edge (0.4), blur (0.1); no depth \\
    Edge threshold / blur strength     & Medium / medium \\
    Denoising steps / guidance scale   & 35 / 3 \\
    Conditional frames / frames per chunk & 1 / 93 \\
    Seed                               & 2025 \\
    Prompts                            & One per episode, describing its scene \\
    Output resolution / frame rate     & Input resolution kept: $640 \times 360$, 10\,fps \\
    Episodes augmented (random / path following) & 605 / 240 (all) \\
    Generation time per episode        & Median 10.8\,min (10.6--21.6\,min) \\
    \bottomrule
    \end{tabular}
\end{table}

\begin{figure}[p]
    \centering
    \includegraphics[width=\linewidth,height=0.85\textheight,keepaspectratio]{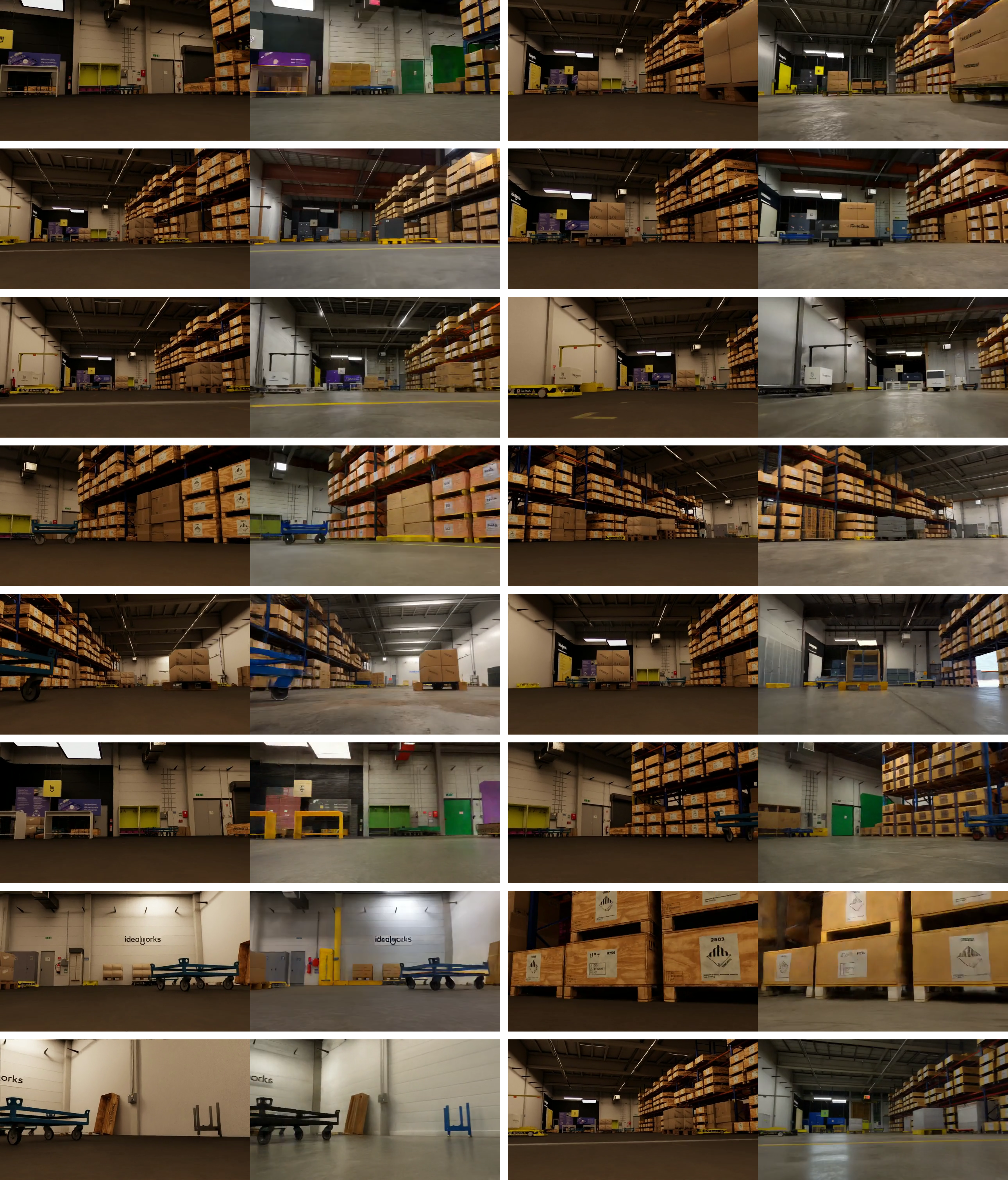}
    \caption{Sim-to-real augmentation samples. Each pair shows a frame of a simulated episode in the digital twin of the R\&D center (left) and the same frame generated by Cosmos Transfer 2.5 (right).}
    \label{fig:cosmos_samples}
\end{figure}

Figure~\ref{fig:cosmos_samples} shows 16 samples of simulated episodes and
their Cosmos-generated counterparts.

\clearpage

\FloatBarrier

%% file: tex/appendix/appendix_g_jetson_deployment.tex

\section{Onboard Deployment on the Jetson AGX Xavier}
\label{app:jetson_deployment}

This appendix describes how the COMPASS models of
Section~\ref{sec:real_world_benchmark} were deployed on the iw.hub's onboard
computer (Section~\ref{sec:rw_deployment}): the target hardware and the
conversion from the trained model to a TensorRT engine.

\subsection{Target Hardware and Software}
\label{app:jetson_platform}

\begin{table}[H]
    \centering
    \caption{Onboard compute platform of the iw.hub.}
    \label{tab:jetson_platform}
    \renewcommand{\arraystretch}{1.2}
    \footnotesize
    \begin{tabular}{l c}
    \toprule
    \textbf{Item} & \textbf{Value} \\
    \midrule
    Module                 & NVIDIA Jetson AGX Xavier \\
    Memory                 & 32\text{ GB} \\
    Power mode             & MAXN \\
    JetPack / L4T          & R35 \\
    CUDA / cuDNN           & 11.4 / 8.6.0 \\
    TensorRT               & 8.5.2 \\
    ROS~2 distribution     & iron \\
    \bottomrule
    \end{tabular}
\end{table}


\subsection{Model Conversion}
\label{app:jetson_conversion}


\begin{table}[H]
    \centering
    \caption{Export and TensorRT engine build settings.}
    \label{tab:jetson_conversion}
    \renewcommand{\arraystretch}{1.2}
    \footnotesize
    \begin{tabular}{l c}
    \toprule
    \textbf{Setting} & \textbf{Value} \\
    \midrule
    Source framework / checkpoint & PyTorch 2.10.0+cu128 \\
    ONNX opset                    & 16 \\
    TensorRT precision            & FP16 \\
    Input / output shapes         & Static (Table~\ref{tab:jetson_io}) \\
    \bottomrule
    \end{tabular}
\end{table}

\begin{table}[H]
    \centering
    \caption{Inputs and outputs of the TensorRT engine. All shapes are static (batch size 1). The recurrent state (\texttt{history}, \texttt{sample}) and the previous action are fed back from the outputs of the previous step.}
    \label{tab:jetson_io}
    \renewcommand{\arraystretch}{1.2}
    \footnotesize
    \begin{tabularx}{\textwidth}{l l l X}
    \toprule
    & \textbf{Tensor} & \textbf{Shape} & \textbf{Meaning} \\
    \midrule
    \multirow{7}{*}{In}
      & \texttt{image}         & $1 \times 1 \times 3 \times 320 \times 512$ & batch $\times$ time $\times$ RGB $\times$ height $\times$ width, values 0--1 \\
      & \texttt{route}         & $1 \times 1 \times 10 \times 4$ & 10 route segments, each $(x_1, y_1, x_2, y_2)$ in the robot frame \\
      & \texttt{goal\_heading} & $1 \times 1$ & goal yaw in the robot frame (rad) \\
      & \texttt{speed}         & $1 \times 1 \times 1$ & current forward speed (m/s) \\
      & \texttt{action\_input}  & $1 \times 6$ & previous action, fed back \\
      & \texttt{history\_input} & $1 \times 1024$ & recurrent state, fed back \\
      & \texttt{sample\_input}  & $1 \times 512$ & recurrent state, fed back \\
    \midrule
    \multirow{3}{*}{Out}
      & \texttt{action\_output}  & $1 \times 1 \times 6$ & action; the ROS~2 node uses element 0 as $v_x$ and element 5 as $\omega_z$ \\
      & \texttt{history\_output} & $1 \times 1024$ & next recurrent state \\
      & \texttt{sample\_output}  & $1 \times 512$ & next recurrent state \\
    \bottomrule
    \end{tabularx}
\end{table}

\begin{table}[H]
    \centering
    \caption{Model size at each conversion step.}
    \label{tab:jetson_model_size}
    \renewcommand{\arraystretch}{1.2}
    \footnotesize
    \begin{tabular}{l cccc}
    \toprule
    \textbf{Model / engine} & \textbf{Parameters} & \textbf{Checkpoint (MB)} & \textbf{ONNX (MB)} & \textbf{TensorRT engine (MB)} \\
    \midrule
    COMPASS & 135M (88M used in inference) & 180 & 358 & 177 \\
    \bottomrule
    \end{tabular}
\end{table}

\FloatBarrier

%% file: tex/appendix/appendix_h_digital_twin_replica.tex

\section{Digital-Twin Replica of the Real-World Scenarios}
\label{app:twin_replica}

This appendix reports the full results of the COMPASS policies in the
digital-twin replica of the static scenarios S1, S2, S3 and S5
(Section~\ref{sec:rw_scenarios}): a one-to-one copy in the digital twin of the
R\&D center, with the same layouts, obstacles, start poses and goals as in the
real world and the metrics of Section~\ref{sec:rw_metrics}. The dynamic scenario
S4 is not replicated, as moving obstacles in the digital twin are not explored in
this work, so no interaction metrics are reported, and protective stops are not
measured in simulation. The Idealworks navigation stack is evaluated only
in the real world. Unlike the simulation benchmark of Section~\ref{sec:sim_benchmark}, which evaluates
policies trained in NVIDIA's environments across those environments and the
digital twin, these runs reproduce the real-world benchmark of
Section~\ref{sec:real_world_benchmark}. Table~\ref{tab:rw_sim2real} compares the success rate and weighted travel time
of each policy with its real-world runs. S5 is run without the global planner,
which is implemented only in the COMPASS ROS~2 node of the real robot: \review{no policy
reaches the goal, which shows that the COMPASS policy alone cannot reach a far,
occluded goal (Section~\ref{sec:rw_scenarios}).}

\begin{table}[htbp]
    \centering
    \caption{COMPASS policies in the digital-twin replica of the static real-world scenarios S1, S2, S3 and S5 (S5 without the global planner), over 300 rollouts per scenario and model.}
    \label{tab:rw_twin}
    \renewcommand{\arraystretch}{1.2}
    \footnotesize
    \begin{adjustbox}{max width=\textwidth}
    \begin{tabular}{ll cc ccc c cc}
    \toprule
    & & \multicolumn{2}{c}{\textbf{Task}} & \multicolumn{3}{c}{\textbf{Efficiency}}
    & \textbf{Safety} & \multicolumn{2}{c}{\textbf{Smoothness}} \\
    \cmidrule(lr){3-4} \cmidrule(lr){5-7} \cmidrule(lr){8-8} \cmidrule(lr){9-10}
    \textbf{Scenario} & \textbf{Model}
      & SR (\%) $\uparrow$ & PR (\%) $\uparrow$
      & WTT (s) $\downarrow$ & TTG (s) $\downarrow$ & PLR $\downarrow$
      & CC $\downarrow$
      & J (m/s$^3$) $\downarrow$ & OC $\downarrow$ \\
    \midrule
    \multirow{3}{*}{S1}
       & \mdlNV & 90.33 & 89.20 & 9.42 & 8.51 & 1.180 & 0.090 & 2.22 & 4.26 \\
       & \mdlSim & 93.00 & 92.64 & 9.91 & 9.22 & 1.057 & 0.070 & 2.71 & 2.86 \\
       & \mdlCE & 92.00 & 92.97 & 10.85 & 9.98 & 1.077 & 0.080 & 2.32 & 2.69 \\
    \midrule
    \multirow{3}{*}{S2}
       & \mdlNV & 87.67 & 87.70 & 15.98 & 14.00 & 1.010 & 0.123 & 1.43 & 11.23 \\
       & \mdlSim & 99.67 & 99.57 & 14.35 & 14.31 & 1.003 & 0.003 & 2.82 & 10.74 \\
       & \mdlCE & 98.67 & 98.30 & 15.88 & 15.67 & 1.007 & 0.013 & 2.00 & 13.28 \\
    \midrule
    \multirow{3}{*}{S3}
       & \mdlNV & 51.33 & 71.35 & 27.96 & 14.35 & 1.018 & 0.487 & 2.21 & 8.78 \\
       & \mdlSim & 72.00 & 86.67 & 21.62 & 15.57 & 1.016 & 0.280 & 3.05 & 6.42 \\
       & \mdlCE & 71.00 & 85.0 & 26.49 & 17.22 & 1.04 & 0.290 & 2.32 & 5.42 \\
    \midrule
    \multirow{3}{*}{S5}
       & \mdlNV & 0.00 & 85.98 & \na & \na & \na & 1.000 & 1.06 & 28.91 \\
       & \mdlSim & 0.00 & 87.39 & \na & \na & \na & 1.000 & 3.17 & 8.55 \\
       & \mdlCE & 0.00 & 88.91 & \na & \na & \na & 1.000 & 1.75 & 7.50 \\
    \bottomrule
    \end{tabular}
    \end{adjustbox}
\end{table}

\FloatBarrier